\documentclass[10pt,letterpaper]{article}

\usepackage[top=0.85in,left=2.75in,footskip=0.75in]{geometry}

\usepackage{amsmath,amssymb}
\usepackage{tikz}
\usetikzlibrary{arrows.meta, positioning, shapes.geometric, fit, backgrounds, calc}

\usepackage{changepage}

\usepackage{textcomp,marvosym}

\usepackage{natbib}

\usepackage{nameref,hyperref}

\usepackage[right]{lineno}

\usepackage[nopatch=eqnum]{microtype}
\DisableLigatures[f]{encoding = *, family = * }

\usepackage[table]{xcolor}

\usepackage{array}
\usepackage{booktabs}   
\usepackage{longtable}  
\usepackage{hyphenat}
\newcolumntype{L}{>{\tablefont\ttfamily\small}p{11cm}} 

\usepackage[top=0.85in,left=2.75in,footskip=0.75in]{geometry}
\usepackage{changepage}

\newcolumntype{+}{!{\vrule width 2pt}}

\newlength\savedwidth

\usepackage[utf8]{inputenc}  
\usepackage[T1]{fontenc}
\usepackage{inconsolata}  

\newcommand{\tablefont}{\ttfamily}

\usepackage[aboveskip=1pt,labelfont=bf,labelsep=period,justification=raggedright,singlelinecheck=off]{caption}

\usepackage{lastpage,fancyhdr,graphicx}
\usepackage{epstopdf}
\fancyheadoffset[L]{2.25in}
\fancyfootoffset[L]{2.25in}
\usepackage{subcaption}
\usepackage{float}
\usepackage[normalem]{ulem} 

\usepackage{authblk}

\usepackage{tabularx}

\begin{document}

\vspace*{0.2in}

\begin{flushleft}
{\Large
\textbf{Sentiment sensing: comparative analysis of multilingual sentiment dynamics}
}
\newline
\\
Mara Sabina Bernardi\textsuperscript{1*},
Elisa Di Nuovo\textsuperscript{1},
Francesca Torti\textsuperscript{1}
Giuseppe Tirone\textsuperscript{1,2},
Nicolas Stefanovitch\textsuperscript{1},
Zsuzsanna Pasztor\textsuperscript{1},
Charles Macmillan\textsuperscript{1},
Domenico Perrotta\textsuperscript{1},
Bertrand De Longueville\textsuperscript{1}
\\
\bigskip
\textbf{1} European Commission, Joint Research Centre (JRC), Ispra, Italy\\
\textbf{2} FINCONS SPA, Vimercate, Italy
\\
\bigskip

%
%





* mara.bernardi@ec.europa.eu

\end{flushleft}

\section*{Abstract}

Cross-country sentiment analysis faces three fundamental challenges: language-specific biases distort comparisons, compositional data constraints invalidate standard statistics and temporal dynamics lack rigorous testing frameworks. We introduce sentiment sensing, a framework that addresses these gaps through integrated statistical methods. By combining compositional data analysis, functional data methods, and non-parametric inference, sentiment sensing enables valid multilingual comparisons and statistically defensible detection of sentiment shifts. Our framework has been tested in two case studies, analyzing news articles from the 27 countries of the European Union on the topics of energy and artificial intelligence. Sentiment sensing focuses on multilingual sentiment analysis enabling rigorous comparative science.


\section{Introduction} 
\label{sec:intro}

Sentiment analysis has grown remarkably  \citep{RODRIGUEZIBANEZ2023review}, driven by huge volumes of social media data and proven utility across diverse domains including marketing, politics, economics, health, and crisis management. Recent applications have demonstrated sentiment analysis's capacity to detect emotional responses during the COVID-19 pandemic across more than 100 countries \citep{wang2022global}, detect the impact of major events such as volcanic eruptions in real-time social media streams \citep{farquharson2025multilingual}, and reveal long-term trends in news media negativity spanning two decades \citep{rozado2022longitudinal}.

Despite these advances, significant methodological gaps remain in comparative sentiment analysis, particularly for multilingual corpora across countries, languages, and cultural contexts. Existing approaches are hindered by language-specific biases, lack standardized methods for temporal comparison, and provide limited statistical support for detecting significant sentiment shifts across heterogeneous corpora. These limitations become especially problematic when  comparing sentiment trends across countries with different languages, media ecosystems, and baseline sentiment distributions---precisely the scenario in European Union discourse analysis.

Most sentiment analysis research has focused on improving classification accuracy (improving the ability of models to label text as positive, neutral, or negative) achieving remarkable advances in model architectures, from lexica to transformers \citep{semary2024sentimentPlos,stefanovitch-etal-2022-resources,jabbari2025sentimentPlos}. Far less attention has been paid to the problem of how to analyze the resulting sentiment labels. As a consequence, different commercial and academic tools often produce substantially different outputs on the same data, with companies not providing detailed information on how their results and analyses are obtained \citep{RODRIGUEZIBANEZ2023review}.
As a matter of fact, research has focused on comparing architectures, reporting benchmark accuracies, and fine-tuning for specific domains \citep{RODRIGUEZIBANEZ2023review,wang2022global,rozado2022longitudinal,farquharson2025multilingual,jabbari2025sentimentPlos}. Once sentiment labels are assigned, analysis frequently consists of plotting proportions over time, computing descriptive statistics, visually inspecting trends, or using Granger causality \citep{granger1969causalrelations}. Our framework improves on these approaches, making it easy to perform statistically meaningful comparisons across languages. 

In multilingual settings, multilinguality is seen as important as it enables sentiment classification for low-resource languages \citep{10.1145/3645106,PRIBAN2024123247}.
Beyond performance differences across languages \citep{wang2022global}, sentiment classifiers exhibit systematic, language-dependent biases that distort cross-country comparisons. For example, if a model assigns 80\% neutral labels to German texts but only 40\% to French texts with identical sentiment distributions, then a direct comparison of sentiment proportions becomes misleading. Standard accuracy metrics applied to the classifier predictions are not designed to detect or correct for such distributional biases, and therefore provide limited guidance for comparative analysis.

A further challenge arises from the compositional nature of sentiment data: outputs are proportions summing to one, belonging to a constrained simplex rather than unconstrained Euclidean space. Many standard statistical methods cannot be used with compositional data. Nevertheless, sentiment is often aggregated using raw proportions, net sentiment score\footnote{An industry convention, e.g. used in Medallia \url{https://docs.medallia.com/en/medallia-agile-research/text-analysis/net-sentiment-score-nss}.}, or compound metrics such as those produced by the VADER tool \citep{youvan2024understanding}. 
These approaches ignore compositional constraints, violating standard statistical assumptions. 
Compositional data analysis addresses this limitation through logratio transformations that map proportions to unconstrained real-valued spaces \citep{pawlowsky2015modeling}. Recent studies have begun applying such transformations to sentiment analysis, enabling valid regression and hypothesis testing while preserving relative information \citep{10.1007/978-3-031-83705-0_13,10.3389/fpubh.2025.1597941}.

Finally, most sentiment studies lack a comparison with a baseline. Analyzing sentiment within a topic in isolation cannot distinguish topic-specific patterns from global media shifts. For example, a drop in positive sentiment about energy may reflect the topic itself, seasonal effects, or wide changes in journalistic tone. Comparing topic-specific sentiment against a reference baseline drawn from general discourse provides a way to control for such confounders. In addition, comparing per-country baselines and the topic-specific data addresses any potential language-dependent bias. 

These limitations are increasingly consequential as organizations and policymakers rely on sentiment analysis for critical decisions \citep{RODRIGUEZIBANEZ2023review}. While classifiers produce labels, it is the subsequent analysis that produces understanding, determining whether observed sentiment levels are unusual, change over time, or differ systematically across groups.
To address these gaps, we present \textit{sentiment sensing}, an integrated statistical framework which focuses on what to do with the sentiment labels, providing a statistically-sound framework for quantitative inter-country (inter-language) sentiment analysis via intra-country comparisons, answering the questions: 
(i) How different is the sentiment of the thematic corpus from the sentiment of the baseline corpus in the period considered?
(ii) Which time periods exhibit a significant difference between the sentiment of the thematic corpus from the sentiment of the baseline corpus?

\begin{figure}
    \centering
    \includegraphics[width=0.99\linewidth]{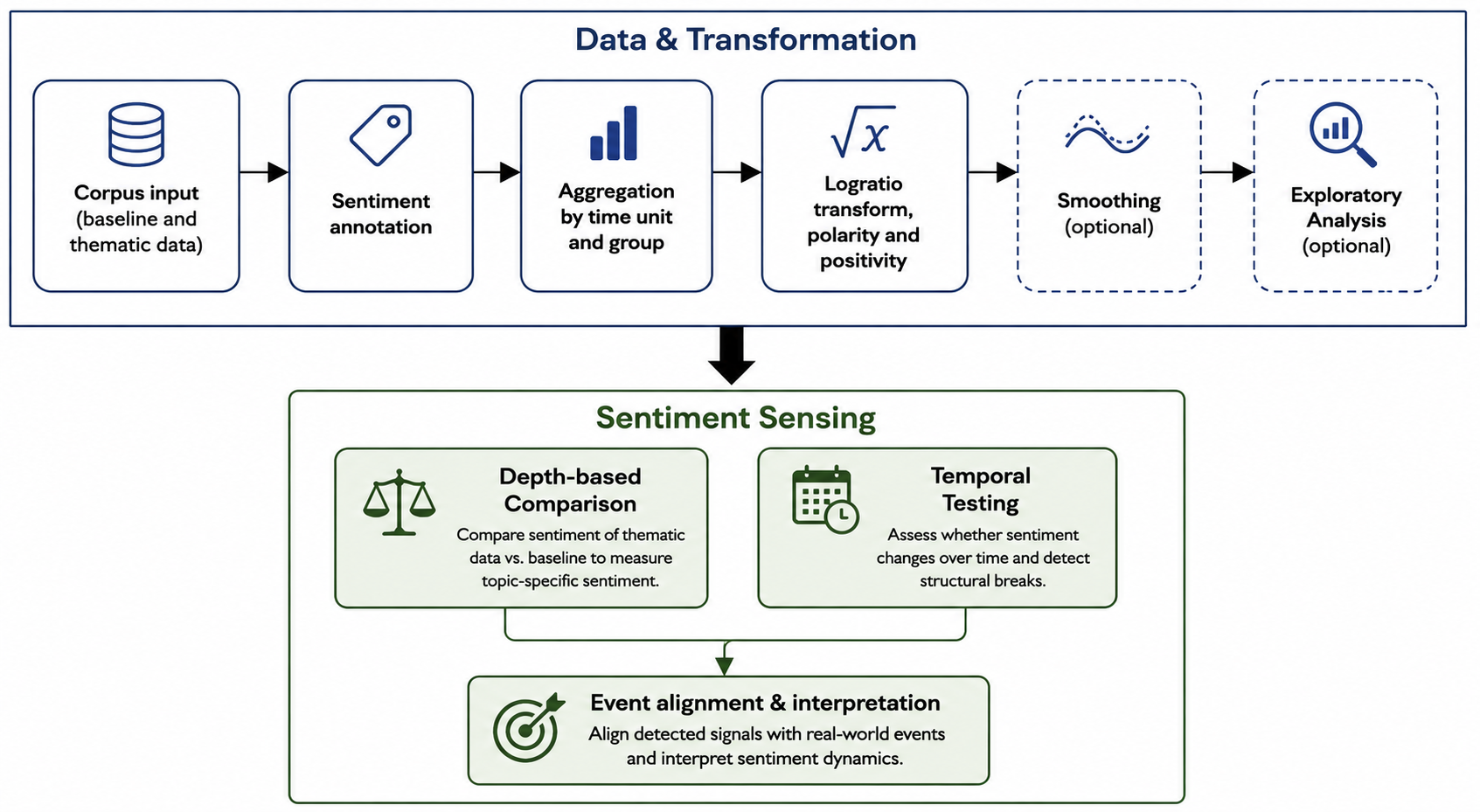}
    \caption{Sentiment sensing workflow. A general framework adaptable to any corpus with temporal structure and comparison groups (e.g., countries, platforms, time periods).}
    \label{fig:diagram_intro}
\end{figure}
The framework, exemplified in Fig~\ref{fig:diagram_intro}, combines \textit{compositional data analysis} to transform sentiment proportions into mathematically unconstrained representations, enabling sound statistical inference, 
\textit{comparative baseline design} to control language-dependent bias and global shifts by measuring sentiment on a specific corpus of interest against a corpus representing general discourse, providing the missing control group for sentiment analysis, and 
\textit{functional data methods} to model temporal patterns as continuous curves, facilitating qualitative comparison across countries to obtain objective criteria for detecting significant sentiment shifts thanks to non-parametric measures and hypothesis testing.

\textit{Sentiment sensing} is model-agnostic and domain-agnostic. It is model-agnostic because it does not depend on how sentiment labels are generated, whether by lexica, classical machine-learning methods, or large language models. It is domain-agnostic because it works on annotations rather than directly on text, and so can be applied to any domain (news media, social media, or other textual corpora) where an appropriate annotation tool is available, and to multiple domains if the annotation tool supports it.

We demonstrate \textit{sentiment sensing} by analyzing news coverage in the 27 EU countries on two topics, energy and Artificial Intelligence (AI), detecting statistically significant sentiment shifts aligned with major events: energy price spikes in mid-2021, and ChatGPT's release in late 2022. 
These case studies have been selected for their societal relevance and for the contrasting ways in which they are framed in public discourse. By examining each case study through the lens of \textit{sentiment sensing}, we aim to illustrate how the proposed framework enables consistent and meaningful comparison of sentiment patterns across different thematic and linguistic contexts.

\section{Related work}
\label{sec:relatedwork}

Sentiment analysis research has progressed from lexicon-based and traditional machine learning approaches toward transformer-based models, yet methodological advances have not fully resolved challenges related to evaluation rigor, domain adaptation, multilingual robustness, temporal inference, and statistically valid aggregation. This section situates the present work within these strands of prior research.

\textbf{Methods and Evaluation.}
\cite{RODRIGUEZIBANEZ2023review} provide a comprehensive overview of sentiment analysis methods, documenting the transition from dictionary-based techniques and classical classifiers to transformer architectures such as BERT, T5, and GPT models. Despite superior benchmark performance, transformers remain underutilized in applied settings, and preliminary GPT-3 evaluations achieved only 57–58\% accuracy, well below specialized fine-tuned models. The review emphasizes the lack of reproducibility and standardized evaluation protocols, noting that different tools can yield inconsistent results on identical data.
Human inter-annotator agreement is often used as a practical performance ceiling. \cite{rozado2022longitudinal} report 80\% agreement for news headline sentiment classification, and with the off-the-shelf RoBERTa model fine-tuned for sentiment, SiEBERT \citep{HARTMANN202375}, achieved 75\% accuracy, approaching non-expert human-level consistency \citep{mozetivc2016multilingual}.
These studies improve classification performance and evaluation practices, but they leave open the question of how sentiment outputs should be analyzed and compared in multilingual and longitudinal settings.

\textbf{Domain Adaptation and Applications.}
Domain adaptation remains a persistent challenge. Although sentiment analysis has been applied across marketing, politics, economics, and health, emergency management and crisis contexts remain underexplored \citep{RODRIGUEZIBANEZ2023review, kryvasheyeu2015performance}. \cite{rozado2022longitudinal} demonstrate that sentiment distributions vary systematically across U.S. news outlets over a 20-year period, while \cite{yan2024has} analyze Reddit college communities using an ensemble of RoBERTa and graph attention networks. Their mixed-effects modeling reveals that institutional characteristics such as size and research intensity are associated with higher odds of negative sentiment, underscoring the value of domain-specific covariates.
In crisis contexts, \cite{farquharson2025multilingual} examines multilingual social media discourse surrounding volcanic eruptions, identifying distinct cross-linguistic information flow patterns and highlighting structural imbalances in language representation. As \cite{RODRIGUEZIBANEZ2023review} note, different tools can produce divergent results on identical data. 
Together, these works highlight the importance of contextual and domain-specific factors, but they do not propose general statistical procedures for cross-domain or cross-language sentiment comparison.

\textbf{Multilingual Sentiment Analysis.}
Multilingual evaluation remains relatively uncommon in large-scale applied studies (e.g. \cite{rozado2022longitudinal} focus on English only), and performance often varies substantially across languages \citep{RODRIGUEZIBANEZ2023review}, as it is usually used as a method to carry out sentiment classification in under-resourced languages using cross-lingual transfer learning \citep{10.1145/3645106,PRIBAN2024123247}. \cite{wang2022global} analyze over 600 million social media posts across 65 languages during the COVID-19 pandemic, achieving 84\% accuracy for English and approximately 75\% for other languages, a significant performance drop that is not explicitly corrected for in their comparative analysis. \cite{farquharson2025multilingual} similarly shows that English dominates discourse even for non-English regionally localized disasters.
As a result, language variation is typically treated as a performance issue rather than as a structural source of bias for comparative inference.

\textbf{Temporal Analysis and Statistical Inference.}
Temporal sentiment analysis enables formal testing of event-driven and long-term trends. \cite{rozado2022longitudinal} use regression analysis to identify a statistically significant increase in negative news sentiment since 2000. \cite{wang2022global} employ regression discontinuity designs and synthetic control methods to estimate the causal impact of the COVID-19 outbreak and lockdown policies, demonstrating sentiment shifts far exceeding normal temporal variability. Change-point detection methods, such as the PELT algorithm used by \cite{farquharson2025multilingual}, provide complementary tools for identifying statistically significant regime shifts in sentiment time series.

\textbf{Aggregation and Compositional Methods.}
Sentiment is commonly aggregated using proportions, net sentiment scores, or compound metrics such as those produced by the VADER tool \citep{youvan2024understanding}. However, these approaches ignore the compositional nature of sentiment data, violating assumptions of standard statistical methods. Compositional data analysis addresses this limitation through log-ratio transformations that map proportions to unconstrained real-valued spaces \citep{pawlowsky2015modeling}. Recent studies have begun applying such transformations to sentiment analysis, enabling valid regression and hypothesis testing while preserving relative information \citep{10.1007/978-3-031-83705-0_13,10.3389/fpubh.2025.1597941}.
These approaches provide a principled statistical basis, but they have not yet been widely adopted in applied multilingual sentiment studies or combined with baseline and temporal comparison designs.

\textbf{Summary and Gaps.}
Prior work demonstrates sentiment analysis applications across media, public health, education, and crisis communication, but key components, i.e. multilingual bias control, compositional data treatment, baseline comparison, and formal temporal inference, have largely been developed in isolation. They have not yet been systematically integrated into a single framework for comparative, cross-linguistic, and longitudinal sentiment analysis.
In addition, prior work rarely adopts explicit baseline-referenced designs analogous to control groups, and functional data analysis remains largely unexplored in sentiment research.
These gaps motivate the need for statistically rigorous frameworks capable of supporting interpretable temporal and cross-regional sentiment comparisons, which the present study seeks to address.

\section{Materials: preparing for sentiment sensing} 
\label{sec:materials}

In this section, we describe the materials needed for \textit{sentiment sensing}. \textit{Sentiment sensing} is an interdisciplinary framework inspired by corpus linguistics and built on statistics and Natural Language Processing (NLP).
The contribution of corpus linguistics is the use of a well-established methodological approach, called keyness analysis or comparative corpus analysis, which involves characterizing a domain-specific corpus (thematic data) through comparison with a general corpus (baseline data). Therefore, in \textit{sentiment sensing} to characterize sentiment on a specific topic, we need two corpora: the baseline data representing general language use, and the thematic data, which collects specialized data on the topic(s) one wants to study. Then we employ NLP tools to label the sentiment of the data, with the constraint that the same tool must be used to annotate both baseline and thematic data. Statistical methods are then used to analyze the resulting datasets, giving valid cross-country comparisons by correcting for differences between countries that may be due to language-specific model performance or country-specific cultural differences.

To illustrate this framework, we apply it to news data from the 27 EU countries. However, it can be applied to any multilingual textual or multimodal data which has been annotated for sentiment. In this application, most of the considered countries are characterized by one predominant language.
We collect, for each EU country, the following corpora (more details on corpus collection are reported in Appendix~\ref{sec:appendix_data_collection}):
\begin{itemize}
    \item The \emph{energy data}, a corpus of thematic news items related to the topic of energy for the time period from 2020 to 2022 (total of 245,802 news articles).
    \item The \emph{AI data}, a corpus of thematic news items related to the topic of AI for the time period from 2020 to 2024 (total of 270,248 news articles).
    \item The \emph{baseline data}, a corpus of randomly selected news items covering the periods used for both the energy and AI data (total of 447,310 news articles).
\end{itemize}

To achieve the best results in automatic sentiment labeling, one needs to use the model that is the most suitable for the languages and domains one is targeting, as performance decreases when the data is more and more dissimilar. Since our illustrative examples collect multilingual news data, we use a sentiment model called XLM-RLnews-8 \citep{10.1007/978-3-031-70242-6_3}, an XLM-RoBERTa-Large model \citep{conneau-etal-2020-unsupervised} domain-adapted to multilingual news and subsequently fine-tuned on multilingual sentiment (more details on sentiment annotation are reported in Appendix~\ref{app:dataAnn}). 

\begin{figure}[t!]
\begin{minipage}[c]{0.03\textwidth}
\centering
\textbf{ } 
\end{minipage}
\hspace{2mm} \hfill
\begin{minipage}[c]{0.3\textwidth}
\centering\textbf{positive} 
\end{minipage}
\hfill
\begin{minipage}[c]{0.3\textwidth}
\centering\textbf{neutral} 
\end{minipage}
\hfill
\begin{minipage}[c]{0.3\textwidth}
\centering\textbf{negative} 
\end{minipage}
\\[1mm]
\begin{minipage}[c]{0.03\textwidth}
\centering
\rotatebox{90}{%
\shortstack{
\textbf{\footnotesize XLM-RLnews-8} \\
\textbf{\footnotesize } 
}}
\end{minipage}
\hspace{3mm} 
\hfill
\begin{minipage}{0.95\textwidth}
\includegraphics[width=\textwidth,trim=0 10mm 0 17mm,clip]{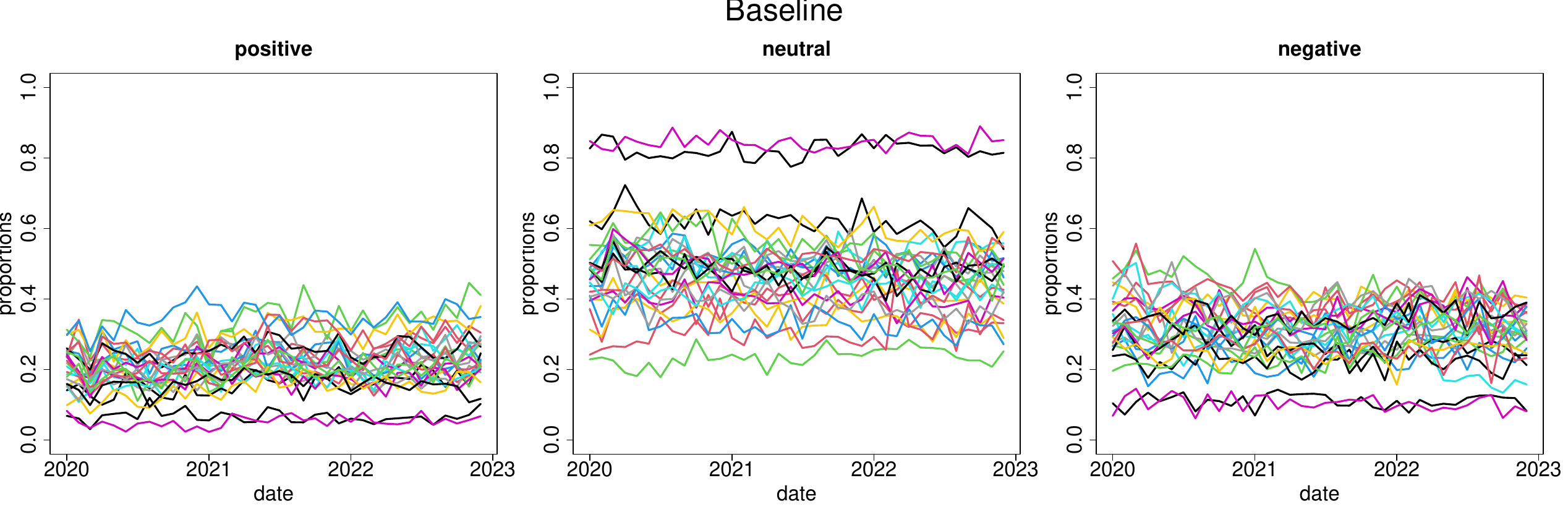}
\end{minipage}
\\[3mm]
\begin{minipage}[c]{0.03\textwidth}
\centering
\rotatebox{90}{%
\shortstack{
\textbf{\footnotesize XLM-RLnews-5} \\
\textbf{\footnotesize debiased}
}}
\end{minipage}
\hspace{3mm} 
\hfill
\begin{minipage}{0.95\textwidth}
\includegraphics[width=\textwidth,trim=0 10mm 0 17mm,clip]{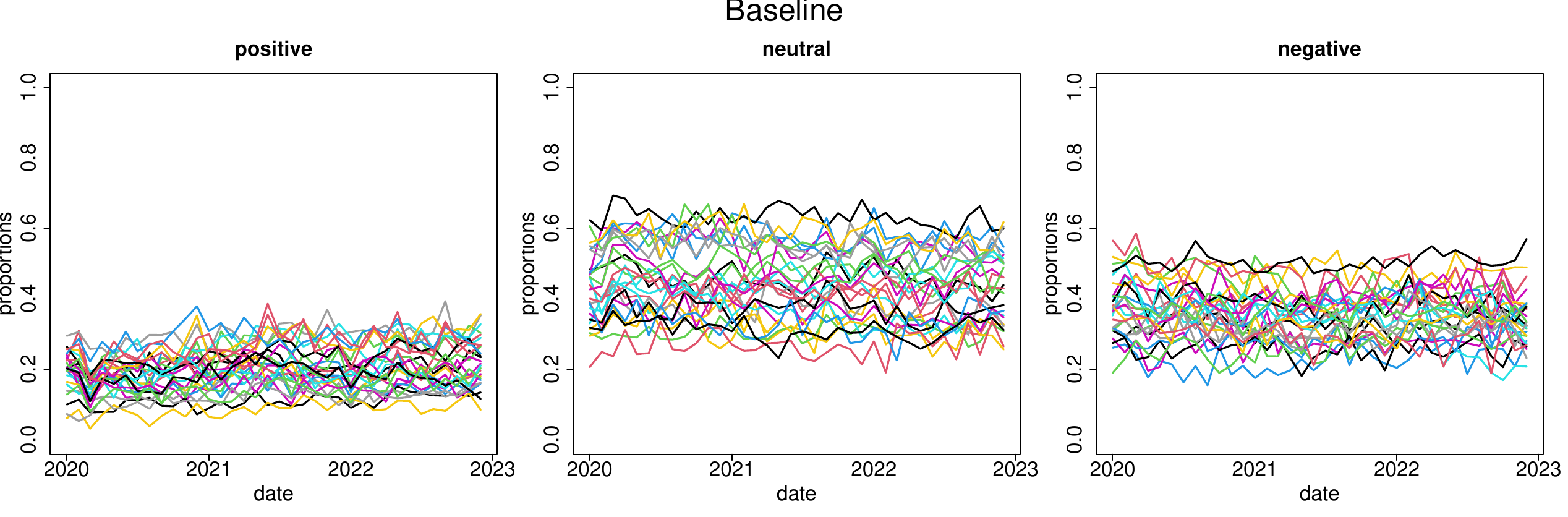}
\end{minipage}
\\[3mm]
\caption{Baseline data. Time series of monthly proportions of news articles with a specific sentiment label (one for each panel) annotated with XLM-RLnews-8 (top panels) and the debiased model XLM-RLnews-5 (bottom panels) for the 27 EU countries. Note the presence of two qualitatively-detected outliers in the top panels, absent in the bottom panels. These are the two German-speaking countries' headlines receiving more neutral labels when annotated with XLM-RLnews-8.
\label{fig:baseline_proportions_merged}}
\end{figure}

After sentiment annotation, the dataset is aggregated by month and country: this gives three time series, for each country, of monthly proportions of news articles assigned to each sentiment category (see top panels of Fig~\ref{fig:baseline_proportions_merged}). Visually inspecting the sentiment proportions of the countries in the figure (each line represents one country) enables the identification of two clusters, one with almost all the countries grouped together, and another with only two countries, standing as outliers. These outliers are German and Austrian sources, which consistently receive more neutral labels than the other countries' news (more than 80\% of the labeled news items). This might indicate a language-dependent bias of the model, as they are both German-speaking countries. As our framework characterizes sentiment by comparing thematic and baseline data per country, it inherently overcomes this classification issue. However, to improve annotation accuraccy for German, we retrained a new sentiment model, XLM-RLnews-5, which uses the same XLM-RLnews-8 architecture, with the removal of  non-EU languages and the German from the fine-tuning dataset (see bottom panels of Fig~\ref{fig:baseline_proportions_merged}), leaving a 5-language sentiment fine-tuning dataset (see Appendix~\ref{app:dataAnn}). We use the labels obtained with XLM-RLnews-8 for the energy data and those obtained with XLM-RLnews-5 for the AI data, to show that \textit{sentiment sensing} is model-agnostic and addresses language-induced model biases.

We apply a statistically motivated transformation to obtain a simplified, non-redundant representation of the sentiment data. Since the three proportions (positive, neutral, and negative) sum to one, they form a composition and do not vary independently. The transformation reduces the data from three to two dimensions while preserving interpretability and information, explicitly capturing the dependence structure between sentiments and enabling more meaningful visualization than analyses based on isolated or net sentiment measures.

We use the isometric logratio transformation with a sequential binary partition \citep{pawlowsky2015modeling}:
$$ p_1(t) = \log \frac{\sqrt{pos(t)*neg(t)}}{neu(t)}, $$ $$ p_2(t) = \log{ \frac{pos(t)}{neg(t)} }, $$ where $pos(t)$, $neu(t)$, $neg(t)$ are the values of the three time series, i.e. the proportions of positive, neutral and negative news articles, respectively.
The resulting coordinates can be interpreted as a polarity index ($p_1$), contrasting polarized versus neutral content, and a positivity index ($p_2$), contrasting positive versus negative sentiment. 
Similar transformations have recently been used in sentiment analysis (see Section~\ref{sec:relatedwork}).

\begin{figure}[t!]
\begin{minipage}{\textwidth}
\includegraphics[width=\textwidth,trim=0 12mm 0 15mm,clip]{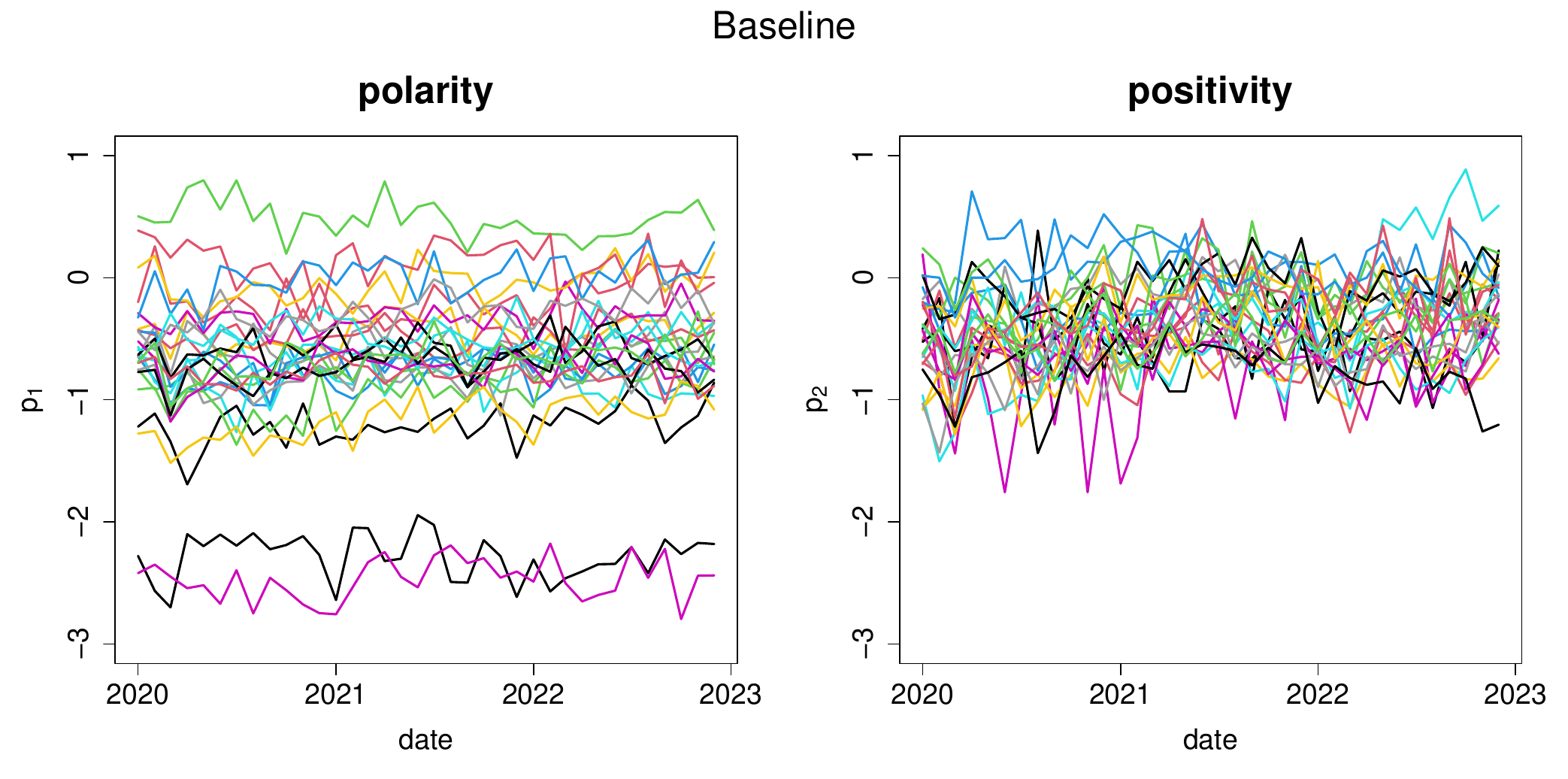}
\caption{XLM-RLnews-8 baseline data. Time series for the 27 EU countries obtained applying the polarity and positivity transformation to the data of the top panels of Fig~\ref{fig:baseline_proportions_merged}. 
\label{fig:baseline_sbp}\vspace{3mm}}
\end{minipage}
\begin{minipage}{\textwidth}
\includegraphics[width=\textwidth,trim=0 12mm 0 15mm,clip]{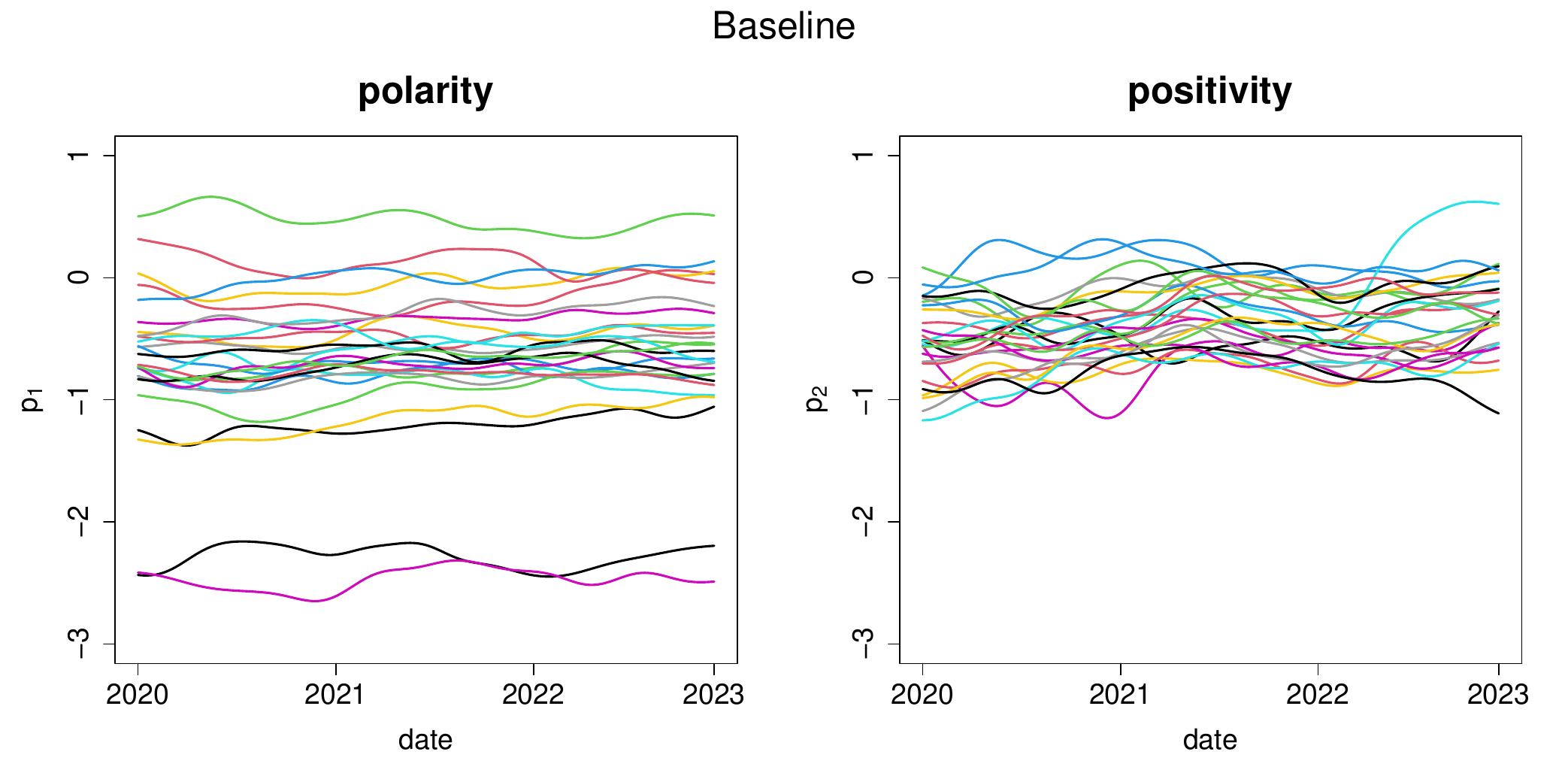}
\caption{XLM-RLnews-8 baseline data. Smooth curves of polarity and positivity for the 27 EU countries. Result of the smoothing step applied to the time series of Fig~\ref{fig:baseline_sbp}.
\label{fig:baseline_sbp_smooth}}
\end{minipage}
\end{figure}

To further corroborate the choice for the polarity and positivity indexes just described, we carry out correspondence analysis \citep{greenacre2017} 
pairing it to a recent robust variant that 
mitigates the effect of atypical data patterns~\citep{riani2022}. The results of the correspondence analysis, are reported in ~\cite{dinuovo2026sentimentbias}. Results confirm that an appropriate two-dimensional representation of the data should account first for the contrast between polarized sentiments and the neutral sentiment (the first direction identified by this dimensionality reduction technique) and then for the contrast between the two polarized sentiments (the second direction). This convergence between the conceptually guided transformation and the data-driven correspondence analysis underscores that polarity and positivity indices, $p_1$ and $p_2$ respectively, are intrinsic to the organization of sentiment patterns across countries. In addition, the robust correspondence analysis confirms with statistical confidence the outliers qualitatively observed in the top panels of Fig~\ref{fig:baseline_proportions_merged}.

Following the transformation of the data utilizing the polarity and positivity indexes, we derive the new data presented in Fig~\ref{fig:baseline_sbp} from the original, redundant input sentiment data depicted in the top panels of Fig~\ref{fig:baseline_proportions_merged}.

To facilitate a more accurate comparison between countries, we build smooth functions from the time series. This eliminates high-frequency fluctuations, thereby reducing noise and preserving the overall trend for each country.
Moreover, this smoothing step mitigates the impact of missing values in the time series of polarity and positivity, which could result from infinite or indeterminate forms that may arise from division by zero in the transformations considered.
We smooth the time series via local polynomial fitting with a Gaussian kernel weight. Fig~\ref{fig:baseline_sbp_smooth} shows the smoothed polarity and positivity curves for the 27 EU countries obtained from the time series of Fig~\ref{fig:baseline_sbp}.

As an exploratory step, we apply functional principal component analysis to the smoothed data to obtain a synthetic, low-dimensional representation of the curves, which facilitates qualitative exploration, interpretation, and visualization. Applying this approach to the baseline and thematic corpora in both case studies, we find that the leading components explain comparable proportions of variance across datasets. This suggests that the overall structure of sentiment variability is similar, but does not by itself capture substantive differences in their temporal dynamics. 
Moreover, because functional principal component analysis analyzes each corpus in isolation, it does not control for language- or corpus-specific biases and therefore cannot directly disentangle thematic effects from baseline linguistic variation. This motivates a more targeted and comparative analysis of the time evolution of sentiment in the thematic corpora, which we address through \textit{sentiment sensing}.
Full results of the functional principal component analysis are reported in Appendix~\ref{sec:appendix_PCA}.

\section{Methods: integrating the materials in the sentiment sensing framework}
\label{sec:dataAnalysis}

In this section, we present the proposed framework, \textit{sentiment sensing}, describing how to perform a meaningful comparison of sentiment patterns accounting for language-dependent performance differences and detecting trends in their temporal evolution.
\begin{figure}[!htbp]
\centering
\resizebox{\textwidth}{!}{%
\begin{tikzpicture}[
  phasetitle/.style={font=\bfseries\Large, align=center, text=white,
    rectangle, rounded corners=3pt, minimum width=4.2cm,
    minimum height=0.9cm, fill=#1},
  phasetitle/.default=black,
  stepbox/.style={rectangle, rounded corners=2pt, draw=black!50, line width=0.6pt,
    fill=white, minimum width=4cm, minimum height=0.85cm,
    font=\Large, align=center, text width=3.8cm},
  stepboxwide/.style={rectangle, rounded corners=2pt, draw=black!50, line width=0.6pt,
    fill=white, minimum width=5.2cm, minimum height=0.85cm,
    font=\Large, align=center, text width=6.5cm},
  optbox/.style={rectangle, rounded corners=2pt, draw=black!40, line width=0.6pt,
    fill=white, minimum width=4cm, minimum height=0.85cm,
    font=\Large\itshape, align=center, text width=3.8cm, dashed},
  interboxwide/.style={rectangle, rounded corners=3pt, draw=#1!70, line width=1.2pt,
    fill=#1!10, minimum width=5.2cm, minimum height=0.85cm,
    font=\Large\bfseries, align=center, text width=6.5cm},
  interboxwide/.default=black,
  qboxwide/.style={font=\Large, text=black!70, align=center, text width=7.5cm,
    inner sep=2pt},
  arr/.style={-{Stealth[length=2.5mm]}, line width=0.9pt, black!60},
  bigarr/.style={-{Stealth[length=3.5mm, width=3mm]}, line width=1.4pt, black!50},
  optarr/.style={-{Stealth[length=2.2mm]}, line width=0.8pt, black!40, dashed},
  phasebg/.style={rounded corners=6pt, fill=#1!5, draw=#1!25, line width=0.5pt},
]


\node[stepbox] (s3) at (0, -1.2) {Aggregate proportions\textsuperscript{2}\\by time unit \& group};
\node[stepbox, left=0.8cm of s3] (s2) {Sentiment annotation\textsuperscript{1}};
\node[stepbox, left=0.8cm of s2] (s1) {Baseline corpus +\\Thematic corpus};
\node[stepbox, right=0.8cm of s3] (s4) {Logratio transformation\textsuperscript{3}\\$p_1$ (polarity), $p_2$ (positivity)};
\node[optbox, right=0.8cm of s4] (s5) {Smooth curves\textsuperscript{4}};

\coordinate (phase1center) at ($(s1.north west)!0.5!(s5.north east)$);
\node[phasetitle=teal!80!black, above=0.8cm of phase1center] (t1) {Phase 0};
\node[font=\Large\bfseries, text=teal!60!black, right=0.15cm of t1] (t1sub) {Data \& Transformation};

\draw[arr] (s1) -- (s2);
\draw[arr] (s2) -- (s3);
\draw[arr] (s3) -- (s4);
\draw[arr] (s4) -- (s5);



\node[optbox, below=2.8cm of s3] (oB) {Correspondence\\analysis};
\node[optbox, left=0.8cm of oB] (oA) {Functional PCA};
\node[optbox, right=0.8cm of oB] (oC) {Cluster, bias \&\\outlier detection};

\coordinate (optcenter) at ($(oA.north west)!0.5!(oC.north east)$);
\node[phasetitle=black!40, above=0.6cm of optcenter] (topt) {Optional};
\node[font=\Large\bfseries, text=black!55, right=0.15cm of topt] (toptsub) {Exploratory Analysis};

\draw[optarr] (oA) -- (oB);
\draw[optarr] (oB) -- (oC);

\begin{scope}[on background layer]
\node[phasebg=teal, fit=(t1)(t1sub)(s1)(s2)(s3)(s4)(s5)(topt)(toptsub)(oA)(oB)(oC),
  inner sep=10pt] (bg1) {};
\end{scope}

\begin{scope}[on background layer]
\node[rounded corners=6pt, fill=white, draw=black!40, line width=0.7pt, dashed,
  fit=(topt)(toptsub)(oA)(oB)(oC), inner sep=6pt] (bgopt) {};
\end{scope}

\draw[optarr, line width=1pt] (s4.south) -- (bgopt.north);


\node[phasetitle=violet!65, minimum width=6cm] (tsa) at (0, -9.5) {Sentiment Sensing};

\node[phasetitle=orange!80!black, minimum width=10cm] (t2) at (-6.5, -11) {Phase 1};
\node[font=\Large\bfseries, text=orange!60!black, below=0.12cm of t2] (t2sub) {Depth-based Comparison};

\node[qboxwide, below=0.45cm of t2sub] (q1)
  {\textit{Q1: How different is thematic sentiment from the baseline?}};
\node[stepboxwide, below=0.5cm of q1] (s6)
  {Resample baseline\textsuperscript{5}\\($n$ iter.\ $\times$ $k$ items)};
\node[stepboxwide, below=0.6cm of s6] (s7)
  {Band depth computation\\(thematic vs.\ baseline)};
\node[stepboxwide, below=0.6cm of s7] (s8)
  {Standardized distance\\from baseline mean depth};
\node[interboxwide=orange, below=0.6cm of s8] (s9)
  {Inter-group comparison\textsuperscript{6}};

\draw[arr] (s6) -- (s7);
\draw[arr] (s7) -- (s8);
\draw[arr] (s8) -- (s9);

\begin{scope}[on background layer]
\node[phasebg=orange, fit=(t2)(t2sub)(q1)(s6)(s7)(s8)(s9),
  inner sep=8pt] (bg2) {};
\end{scope}

\node[phasetitle=red!70!black, minimum width=8cm] (t3) at (6, -11) {Phase 2};
\node[font=\Large\bfseries, text=red!50!black, below=0.12cm of t3] (t3sub) {Temporal Testing};

\node[qboxwide, below=0.45cm of t3sub] (q2)
  {\textit{Q2: Which time periods show significant differences?}};
\node[stepboxwide, below=0.5cm of q2] (s10)
  {First derivative\\of curves};
\node[stepboxwide, below=0.6cm of s10] (s11)
  {Pointwise bootstrap test\\(thematic vs.\ baseline)};
\node[interboxwide=red, below=0.6cm of s11] (s12)
  {Inter-group comparison\textsuperscript{6}};

\draw[arr] (s10) -- (s11);
\draw[arr] (s11) -- (s12);

\begin{scope}[on background layer]
\node[phasebg=red, fit=(t3)(t3sub)(q2)(s10)(s11)(s12),
  inner sep=8pt] (bg3) {};
\end{scope}

\draw[bigarr] (bg2.east) -- (bg3.west);

\coordinate (eventY) at (0, -22.5);
\node[stepboxwide, fill=gray!8, draw=black!60, line width=1.2pt,
  font=\Large\bfseries] (s13) at (eventY) {Event alignment \& interpretation};

\draw[arr, line width=1pt] (s9.south) |- (s13.west);
\draw[arr, line width=1pt] (s12.south) |- (s13.east);

\coordinate (sa-left) at (bg1.west |- tsa);
\coordinate (sa-right) at (bg1.east |- tsa);

\begin{scope}[on background layer]
\node[rounded corners=8pt, fill=violet!4, draw=violet!35, line width=1pt,
  fit=(tsa)(bg2)(bg3)(s13)(sa-left)(sa-right), inner sep=14pt] (bgsa) {};
\end{scope}

\begin{scope}[on background layer]
\node[phasebg=orange, fit=(t2)(t2sub)(q1)(s6)(s7)(s8)(s9), inner sep=8pt] {};
\node[phasebg=red, fit=(t3)(t3sub)(q2)(s10)(s11)(s12), inner sep=8pt] {};
\end{scope}

\draw[bigarr] (bg1.south) -- (bgsa.north);

\node[anchor=north west, text width=25cm, font=\Large, align=left]
  at ([yshift=-0.8cm]bgsa.south west) {%
  \textsuperscript{1} Same model must be used for both corpora.\\
  \textsuperscript{2} Time unit (day/week/month/quarter) and grouping variable (country/language/platform/source) are user-defined.\\
  \textsuperscript{3} Maps compositional data (proportions summing to 1) to unconstrained space for valid statistical inference.\\
  \textsuperscript{4} Optional: skip for early-warning or real-time applications.\\
  \textsuperscript{5} Resampling parameters ($n$, $k$) are case-specific, based on data volume per group.\\
  \textsuperscript{6} Bias-controlled via intra-group design: neutralizes group-specific biases (e.g.\ accuracy-related language-dependent model differences).%
};

\end{tikzpicture}%
}
\caption{A more detailed view of the sentiment sensing workflow. Dotted lines indicate optional steps. 
\label{fig:diagram}\vspace{3mm}}
\end{figure}
The diagram in Fig~\ref{fig:diagram} exemplifies it. 

The workflow presents a two-phase framework for analyzing temporal sentiment across multiple groups in a model-agnostic and statistically rigorous manner. Phase 0 integrates data collection, aggregation, and compositional transformation, combining baseline and thematic corpora, classification, temporal proportion aggregation, and logratio transformation, ensuring compositionally valid and comparable results. It optionally includes also curves smoothing and an exploratory analysis to carry out inter-group comparisons using functional principal component analysis to reveal sentiment structures, visualize low-dimensional patterns, and detect group clustering and biases. Phase 1 presents intra-group analyses, including baseline resampling, band depth computation, and standardized distance metrics, controlling for group-specific biases and isolating topic-specific patterns. Phase 2 covers temporal testing, highlighting derivatives, bootstrap hypothesis testing, changepoint detection, and cross-group aggregation to identify significant shifts in sentiment. The use of these methods does not preclude the possibility of expanding the framework with additional methods that may be more relevant for specific use cases. This approach is mathematically sound, controls confounding effects, and enables reproducible inference across diverse datasets.

We obtain inter-country analysis via intra-country exploration. First, we build a well-established, representative baseline distribution per country by resampling the baseline 100 times using random selections of 3,000 news items from the original baseline data. Then, we carry out intra-country qualitative and quantitative analyses by comparing the thematic curve with the baseline distribution. We perform two types of analysis answering the two research questions:
\begin{enumerate}
    \item How different is the sentiment of the thematic corpus from the sentiment of the baseline corpus in the period considered?
    \item Which time periods exhibit a significant difference between the sentiment of the thematic corpus from the sentiment of the baseline corpus?
\end{enumerate}
To reply to the first question, we compare, for each country, the position of the thematic curve with respect to the distribution of the baseline curves through the concept of depth, a mathematical tool that provides an ordering to a sample of curves from the most central to the most outlying with respect to the given data set. Many notions of depth have been introduced in the functional data analysis literature. Here, we consider the band depth notion introduced in \cite{lopez2009concept,lopez2007depth} and we compute the band depth for univariate functional data using the R package \verb+roahd+ \citep{roahd}. We then reduce the information obtained using the depth notion by computing the standardized distance of the thematic curves' depth from the mean of the depths of the resampled baseline, and visualizing it in a scatterplot. This synthetic view allows for inter-country comparison of the results of the intra-country analysis.

To reply to the second question, we consider the evolution of the sentiment of the thematic corpus over time and we quantitatively measure the differences in the thematic curves with respect to the resampled baseline data by performing bootstrap tests for significance testing. In the case studies considered, we apply this comparison to the first derivative of the positivity curves as the visual inspection of the curves suggests the presence of an abrupt and sharp decline in the curves. This observation is reflected in the analysis by looking for differences between baseline and thematic corpora in the slope of the positivity curves. However, the same approach can be applied to polarity curves as well as to the curve itself (to detect differences in the magnitude of the curves between baseline and thematic corpus) or to higher order derivatives of the curve (assessing different rates of change in the curvature of the curves). This analysis detects specific time periods when the thematic corpus sentiment deviates from the overall media landscape. This information can be used for monitoring purposes and helps to understand public opinion's reaction to known events.
The framework might also be expanded including methods to track the evaluation of sentiment over time, comparing the sentiment on a defined time period with the sentiment in a different time period.

Having outlined the framework, we now apply \textit{sentiment sensing} to the two case studies: energy (Section~\ref{sec:energy}) and AI (Section~\ref{sec:AI}).

\section{Energy case study through the lens of sentiment sensing}
\label{sec:energy}

The energy case study, including data from 1-1-2020 to 12-31-2022, collects news pieces from a period marked by major economic and geopolitical shocks that shaped energy discourse in Europe. In 2020, the COVID-19 pandemic caused a sharp decline in energy demand and prices. By 2021, however, the EU economy rebounded strongly as vaccination campaigns progressed and restrictions eased. Rising domestic demand, robust investment, and improving consumer confidence combined with persistent supply chain bottlenecks, labor shortages, and limited availability of raw materials (including electricity and natural gas) drove energy prices sharply upward. These supply-side pressures, alongside global commodity price increases, began to reshape public and policy debates around energy affordability and security \citep{EC2021report}.
The situation escalated in 2022 with the start of the Russian war of aggression against Ukraine, triggering spikes in gas and electricity prices, heightened market volatility, and renewed concerns over energy supply. The EU responded with the REPowerEU plan, aiming to reduce dependence on Russian fossil fuels and accelerate the transition to renewable energy. Rising energy prices also contributed to higher production costs for firms and increasing inflationary pressures on households \citep{EC2022report}.
Thanks to \textit{sentiment sensing} we can verify possible statistically significant changes in sentiment towards the topic of energy in the news media, ensuring that the results and the identified trends are statistically sound, accounting for language-dependent classification biases.

\subsection{Qualitative multi-country sentiment analysis}
\label{sec:intra_1_energy}

\textit{Sentiment sensing}, thanks to its thematic-baseline intra-country comparative design, allows to overcome possible sentiment model biases as the curves representing the positioning of the two German-speaking countries in the top panels of Fig~\ref{fig:baseline_proportions_merged} might suggest. 
\begin{figure}[tb!]
\begin{center}
\includegraphics[width=\textwidth,trim=0 12mm 0 15mm,clip]{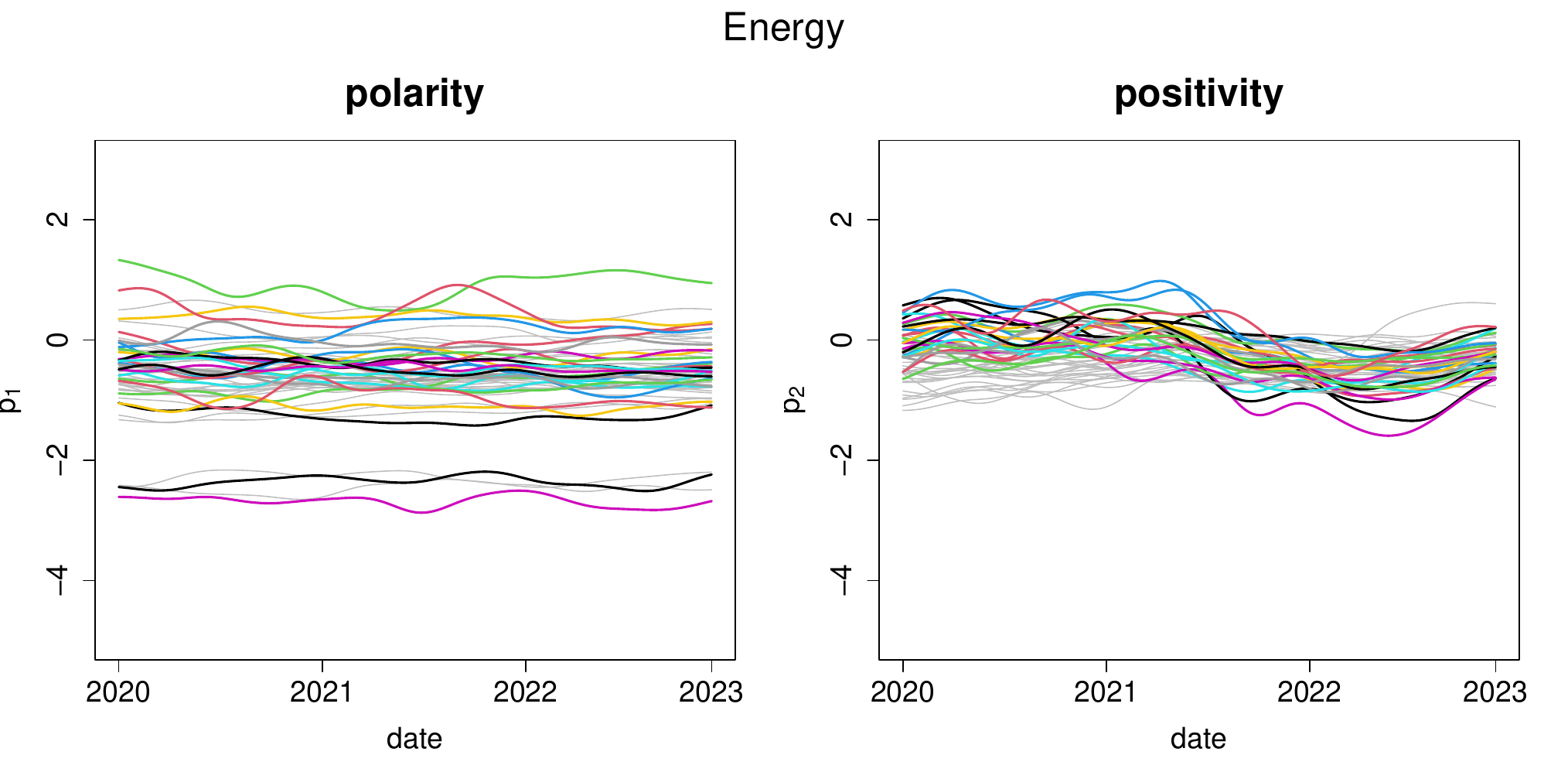}
\end{center}
\caption{Energy data. Curves of polarity and positivity for the 27 EU countries. The thematic corpus data are in colored curves, while the baseline corpus data are in gray.
\label{fig:energy_sbp_smooth}}
\end{figure}
In Fig~\ref{fig:energy_sbp_smooth} we report the curves of polarity and positivity for the energy (colored curves) and baseline (gray curves) data.

The qualitative exploration in Fig~\ref{fig:energy_sbp_smooth}, allows us to observe that, as far as polarity is concerned, we have rather similar values between the energy data and the baseline data, with a constant temporal evolution. As far as positivity is concerned, we can qualitatively see a difference between the two datasets in terms of values: for energy data, we observe a higher level of positivity with respect to the baseline data until mid-2021, when a sudden decrease brings the average level of energy curves slightly below the level of baseline data.

\textit{Sentiment sensing}, however, foresees further exploration aimed at quantitatively evaluate these qualitative observations obtained through the comparison of baseline and thematic data, as exemplified in the next subsections.

\begin{figure}[tb!]
\begin{center}
\includegraphics[width=\textwidth]{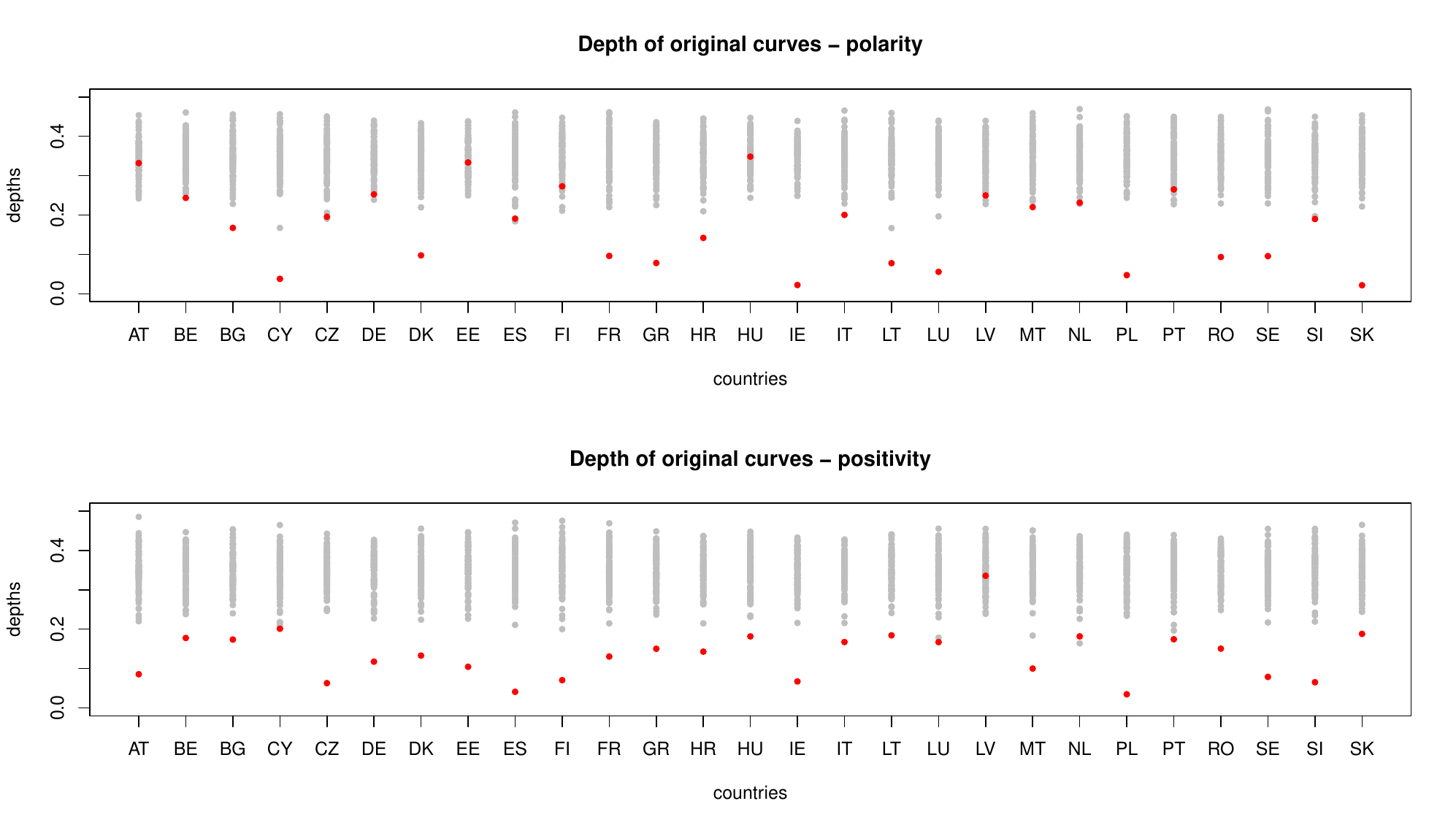}
\end{center}
\caption{Energy data. Depths of the 100 curves obtained from the re-sampling of the baseline data (gray) and of the energy curve (red) for the 27 EU countries (x axis) on polarity (top panel) and positivity (bottom panel). 
\label{fig:depth_sbp}}
\end{figure}

\subsection{Quantitative multi-country sentiment analysis via centrality and outlyingness in the notion of depth}

We perform the analysis separately on the polarity curves and the positivity curves. The computed depths are represented in Fig~\ref{fig:depth_sbp}, where the gray dots represent the depths of the curves of the resampled baseline data, while the red dot represents the depth of the energy curve. If the red dot falls in the span of the gray dots, it means that the sentiment around this topic is spoken similarly to the baseline sentiment, if it falls outside, it means that there is a quantitative difference in the sentiment of this topic than the general discourse.

In our case study, represented in Fig~\ref{fig:depth_sbp}, for several countries the depth of the energy curve is smaller than the depths of the baseline curves, either in terms of polarity or positivity, meaning that the energy curve is peripheral in the distribution of the curves considered, indicating that the energy topic is spoken of differently with respect to the overall media landscape. Among the countries with the most pronounced differences we have Poland, PL, for both polarity and positivity. The difference is less pronounced in cases where the depth of the energy curve is within the distribution of the depths of the baseline curves (such as polarity for Austria, AT, Estonia, EE, and Hungary, HU, and positivity for Latvia, LV). This quantitatively characterize the sentiment in the news representing these countries accounting for the language bias of the model (Austria, AT, and Germany, DE, received more neutral labels than other countries). Comparisons that do not systematically compare the sentiment on a thematic corpus against its baseline fail to gain these insights. 

\begin{figure}[tb!]
\centering\includegraphics[width=0.5\textwidth,trim=0 0 0 15mm,clip]{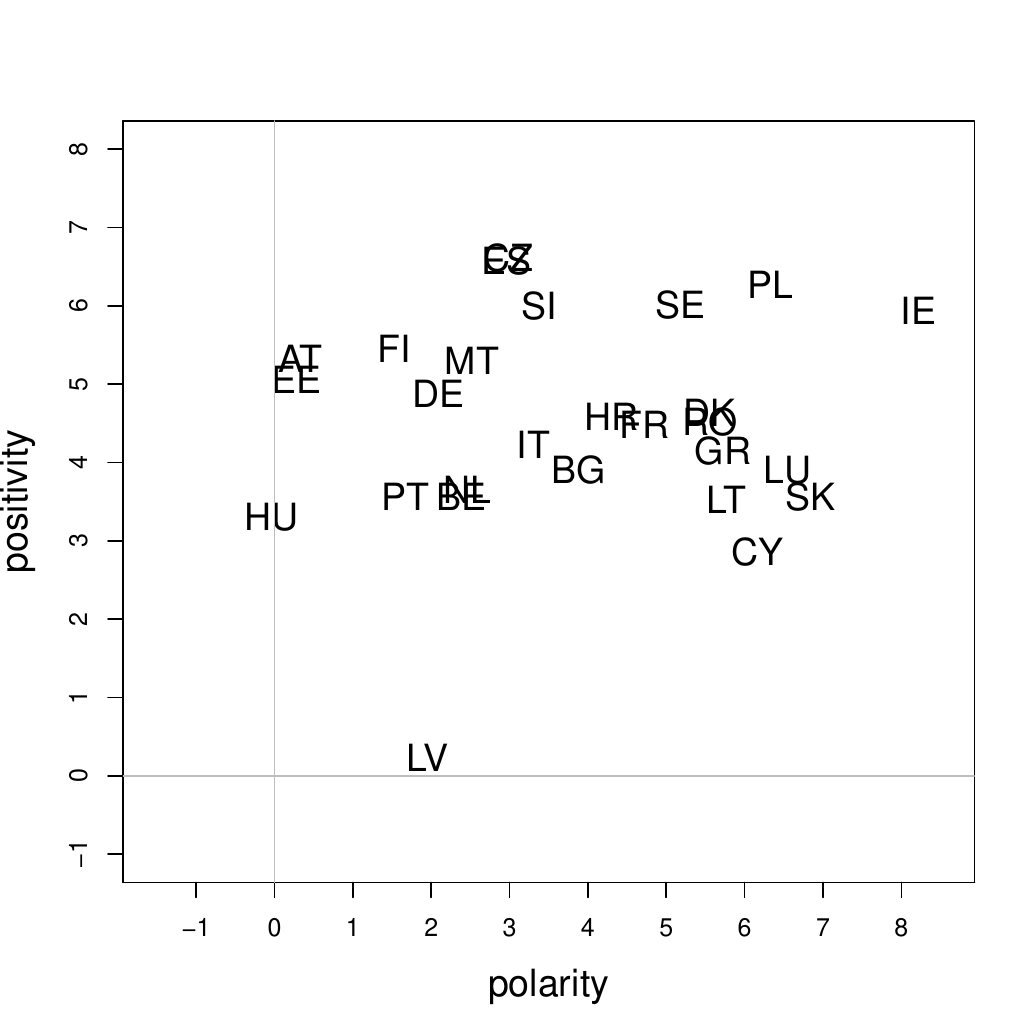}
\caption{Energy data. Scatter-plot of the standardized distances from the mean of the baseline depth for polarity (x-axis) and positivity (y-axis), for the 27 EU countries. 
\label{fig:scores}}
\end{figure}

To synthesize the information displayed in Fig~\ref{fig:depth_sbp}, we compute the standardized distance of the depth of the energy curve from the mean of the depths of the resampled baseline curves. The obtained values are represented in Fig~\ref{fig:scores}, where the x coordinate reports the standardized distances for polarity and the y coordinate reports the standardized distances for positivity. 
The distribution of the countries in this plot tells us whether the sentiment in the analyzed topic, energy in this case, is similar or dissimilar to the baseline. The position in the horizontal dimension tells us if the country has a polarity curve similar to the baseline if it is positioned in the left part of the plot or different from it in the right part of the plot. The position in the vertical dimension vertical gives us insights on the positivity curves, indicating similarities with the baseline in the bottom part and differences in the top part. A similar curve indicates no relevant difference in the sentiment labels found (whether polarity or positivity) in the thematic data and the general discourse. See Appendix~\ref{sec:appendix_intra_1_energy} for some examples.

Moving beyond qualitative observations which enabled us to visualize a greater difference between baseline and energy data in positivity rather than in polarity for the majority of the countries, the quantitative comparison of the depth of the curves enrich the results obtained enabling reliable inter-country comparisons through intra-country analyses. Indeed, as shown in Fig~\ref{fig:scores}, the standardized distances computed for polarity are more spread (x axis, with countries positioned from 0 to more than 8) than the standardized distances for positivity (y axis, with all the countries except one positioned from almost 3 to 7), which are more compact and almost all above the value 3, indicating that the depth of the energy curve is more than 3 standard deviations from the mean of the depth of the baseline curves for all the considered countries except Latvia (LV).
More details on these results on energy data are reported in Appendix~\ref{sec:appendix_intra_1_energy}.

\subsection{Quantitative multi-country sentiment analysis for trend signi\-ficance assessment}
\label{sec:intra_2_energy}

As observed in Fig~\ref{fig:energy_sbp_smooth}, the polarity curves for the energy data are almost flat, indicating no significant temporal evolution for the monthly timescale we use, while the positivity curves show a trend (common to the majority of the countries) with a downward step in the middle of 2021.

To assess the significance of the trend in the positivity curves, we perform bootstrap tests on the first derivative of the energy positivity curve against the first derivative of the baseline curve. The same test could have been performed also for the polarity curves if qualitative observations highlighted temporal changes. The test is conducted pointwise on a fine, uniformly spaced temporal grid by comparing the observed derivative of the energy curve with the empirical distribution obtained from bootstrap resampling of the baseline curve.

\begin{figure}[t!]
\begin{center}
\includegraphics[trim= 0 12mm 500 47,clip,width=0.5\textwidth]{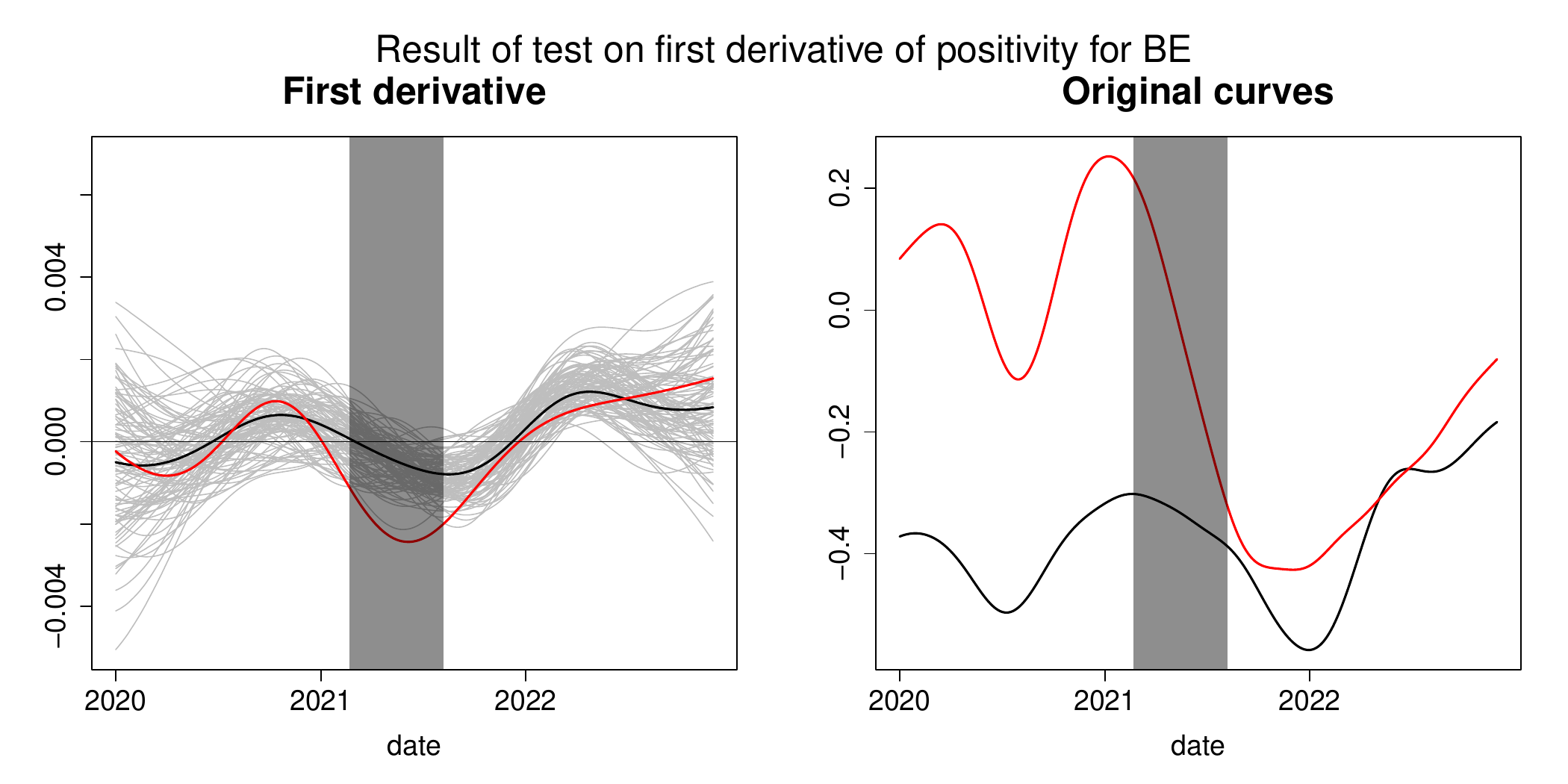}
\end{center}
\caption{Energy data. Results of the bootstrap tests on the first derivatives of the positivity curves for Belgium.
\label{fig:BE_energy_derivative}}
\end{figure}

Fig~\ref{fig:BE_energy_derivative} shows one example of this analysis (for Belgium): the red curve is the first derivative of the energy data and the gray curves are the first derivatives of the resampled baseline data (for reference, the first derivative of the baseline data is reported in black). The bootstrap tests determine, for each time instant, whether there is a significant difference between the value of the red curve and the distribution of the values of the gray curves with a significance level of 1\%. The time period where a significant difference is found is highlighted with a gray vertical band. The gray vertical band in the figure than indicates when the slope of the thematic corpus data is significantly different from the slope of the baseline data.
Additional examples are reported in Appendix~\ref{sec:appendix_intra_2_energy}.

\begin{figure}[tb!]
\begin{center}
\includegraphics[trim= 0 12mm 0 0,clip,width=0.55\textwidth]{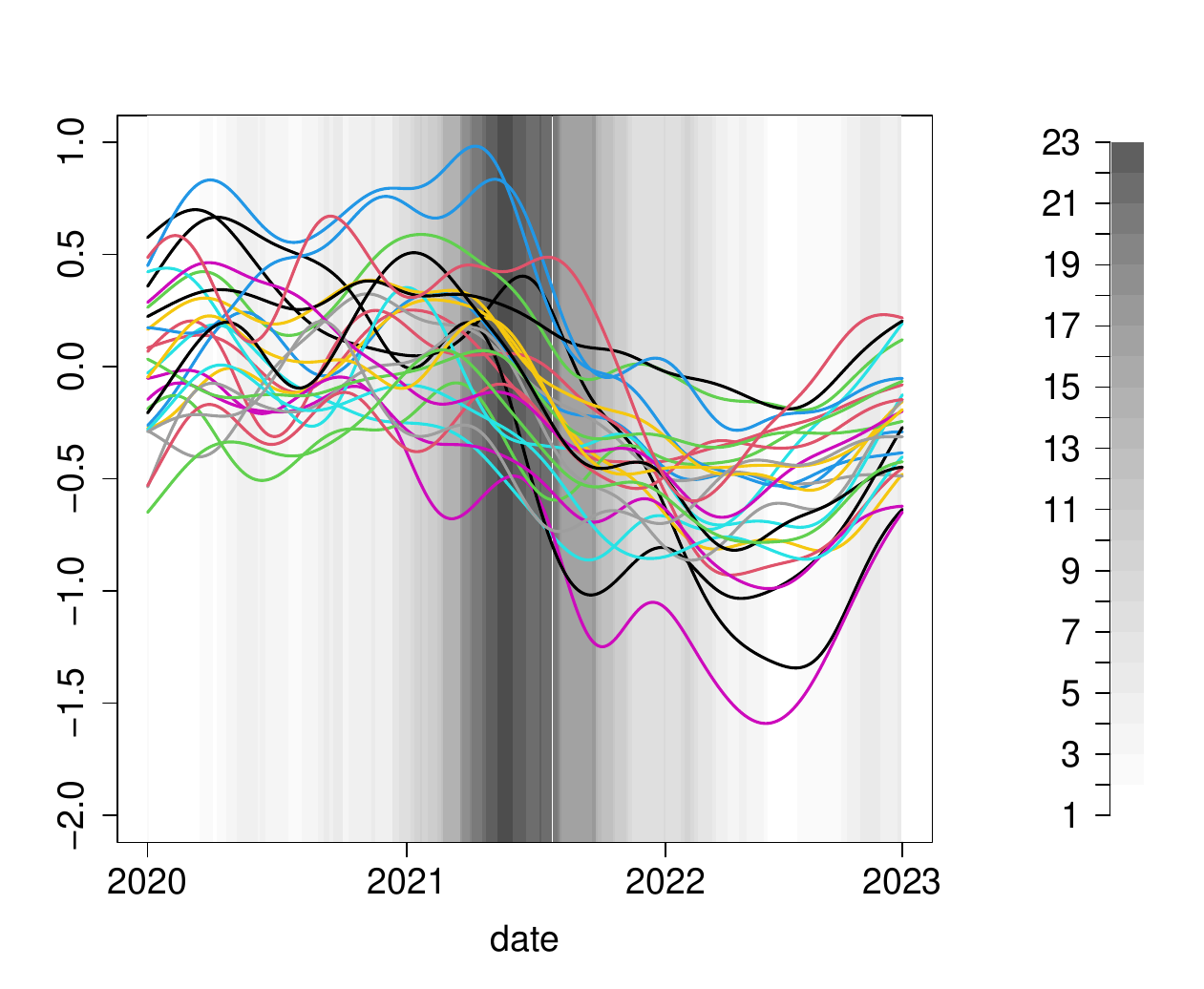}
\end{center}
\caption{Energy data. Summary of the results of the bootstrap tests on the first derivatives of the positivity curves for all the countries. Gray background colors are proportional to the count of countries for which a significant difference has been found.
\label{fig:ALL_energy_derivative}}
\end{figure}

To summarize the results obtained in this analysis performed country by country, and obtain a inter-country comparison of the results, we count how many countries have a significant difference in each point of the fine grid spanning uniformly the time window considered, and the result in shown in Fig~\ref{fig:ALL_energy_derivative}. 

The change towards negativity in 2021 mirrors the sharp increase in energy prices \citep{EC2021report,EC2022report}. The escalation at beginning 2022, instead, receives less significant results in the news sentiment, and it might indicate that the EU response has worked.

\begin{figure}[tb!]
\begin{center}
\hspace{-12mm}\includegraphics[trim= 0 12mm 0 60,clip,width=0.5\textwidth]{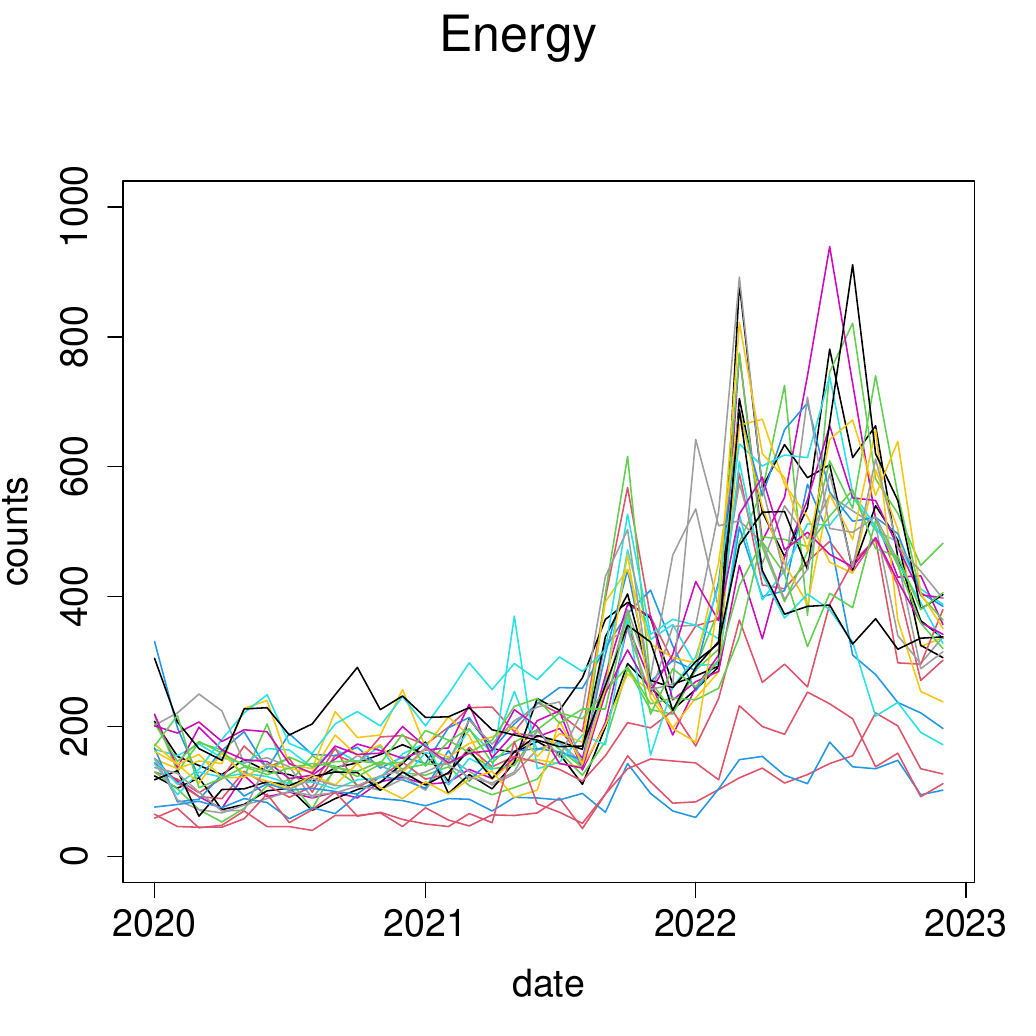}
\end{center}
\caption{Energy data. Total counts of news articles for the 27 EU countries. 
\label{fig:energy_total_counts}}
\end{figure}

The effect of the events mentioned at the beginning of this section is also observable in the volume of news items related to the energy topic, as can be seen in Fig~\ref{fig:energy_total_counts}, which reports the total counts of news articles in the energy dataset for the 27 EU countries with monthly aggregation. A drastic increase in the number of news articles speaking about energy is clearly visible in mid-2021, showing an increase in the interest of the media in the topic in that period. This trend is accentuated in early 2022, with the military aggression of Russia against Ukraine, creating strong uncertainties on energy supply in the EU and again in mid-2022. However, as presented above, the analysis of the sentiment via \textit{sentiment sensing} enables to characterize this increasing interest in the energy topic in the news by providing information on the evolution of the sentiment of the news items concerning this topic.

\section{AI case study through the lens of sentiment sensing}
\label{sec:AI}

The AI case study collects data from 1-1-2020 to 12-15-2024. In this case study, we use the sentiment annotations obtained using XLM-RLnews-5, which is the model solving the German bias towards neutrality identified in XLM-RLnews-8 and applied in the energy case study to demonstrate the model-agnosticity of \textit{sentiment sensing}. The full details on data collection and annotation are reported in Appendix~\ref{sec:appendix_data_collection}.

The case study period has seen several landmark developments in AI. NLP tasks---from text mining (more a language understanding task) to machine translation (more a language generation task)---have achieved remarkable performance improvements, largely thanks to transformer-based approaches, such as masked language models and GPT-like autoregressive models. This led to a wide adoption of these models in different domains, creating the necessity for regulation. The EU has launched the AI Act, firstly proposed in April 2021 by the European Commission. Then, discussions with the European Parliament started in June 2023, and a provisional agreement was reached between the Parliament and the European Council in December 2023, and finally approved by the Parliament in March 2024.
In the meanwhile, at the end of 2022, the tool ChatGPT was publicly launched, making the use of AI free to anyone with an internet connection. AI has since become common knowledge. The state of AI, published annually since 2018, in 2022 report for the first time a \textit{safety} section for the identification and mitigation of ``catastrophic risks that highly-capable future AI systems could pose to us'' \citep{stateof.ai2022}. This signals the raising concerns culminated in March 2023, with the petition advocating for a pause from giant AI experiments, signed also by leading AI researchers as Yoshua Bengio. This petition was then followed in May 2023 by a concise statement signed by over 100 AI professors (Bengio included) called Statement on AI Risk, which emphasizes the need for AI risk mitigation as a global priority. In the next sections, we apply \textit{sentiment sensing} to see whether the AI perception in the news features peculiar characteristics or trends in this time period. 

\subsection{Qualitative multi-country sentiment analysis}
\label{sec:intra_1_AI}

\begin{figure}[tb!]
\begin{center}
\includegraphics[width=\textwidth,trim=0 12mm 0 15mm,clip]{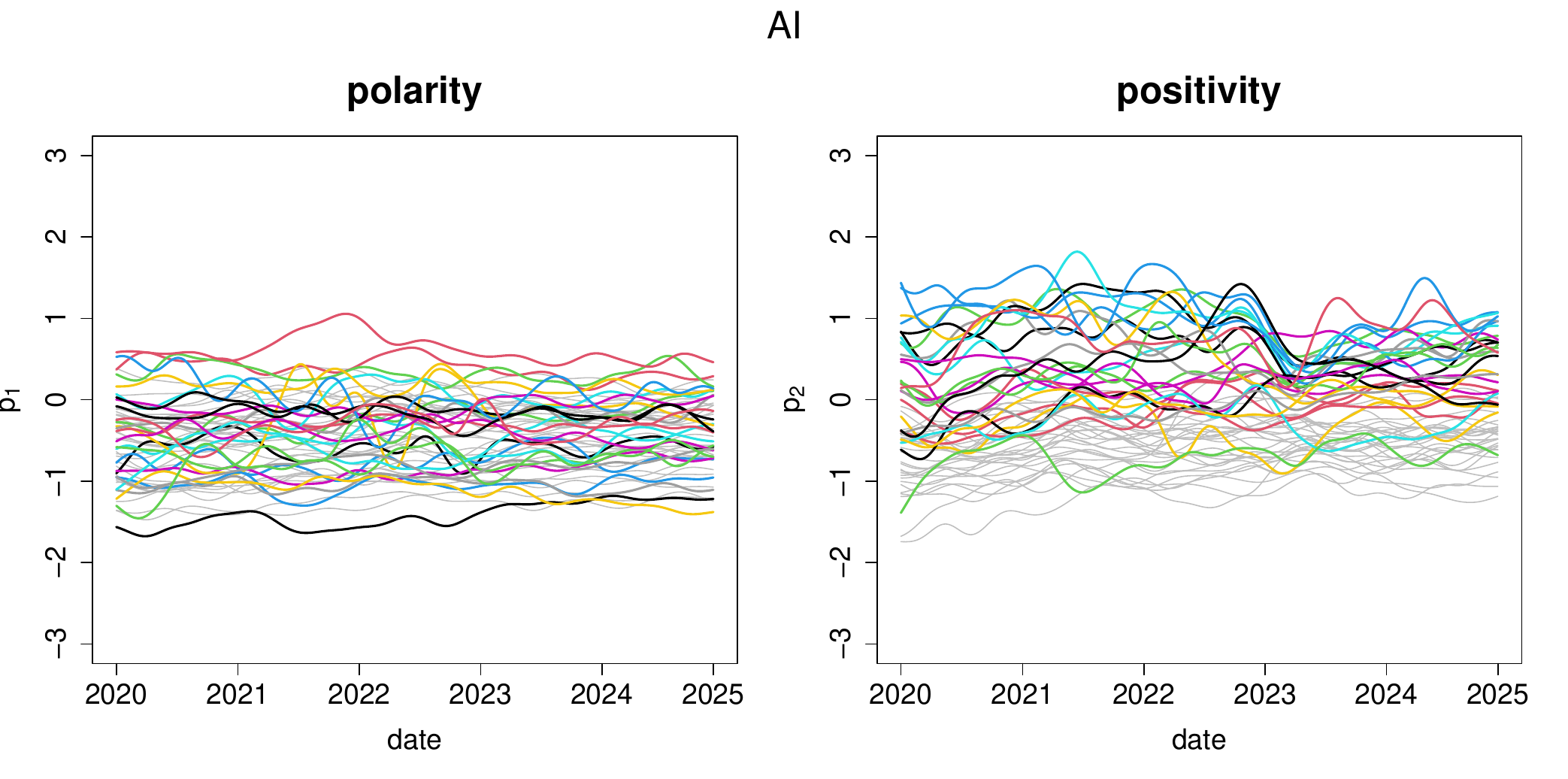}
\end{center}
\caption{AI data. Curves of polarity and positivity for the 27 EU countries. The thematic corpus data are in colored curves, while the baseline corpus data are in gray.
\label{fig:AI_sbp_smooth}}
\end{figure}

As done for the energy case study, we start from the qualitative exploration of the polarity and positivity curves comparing AI (colored curves) and baseline (gray curves) data, showed in Fig~\ref{fig:AI_sbp_smooth}. 

We can qualitatively observe similar values although with a slightly greater dispersion for the polarity curves of the AI data with respect to the baseline data, while maintaining an almost flat temporal evolution. The positivity curves of the AI data appear to have higher values with respect to the baseline data.

To understand quantitatively these differences, we use the depth of polarity and positivity curves of the two datasets, thematic and baseline, to meaningfully compare the news sentiment representing the EU countries (Section~\ref{sec:AI_depth}), and then, since the major differences are qualitatively observed in the positivity panel, we carry out bootstrap tests to identify statistically significant trends on the positivity first derivative curves (Section~\ref{sec:intra_2_AI}).

\subsection{Quantitative multi-country sentiment analysis via centrality and outlyingness in the notion of depth}
\label{sec:AI_depth}

\begin{figure}[tb!]
\begin{center}
\includegraphics[width=0.50\textwidth,trim=0 0 0 15mm,clip]{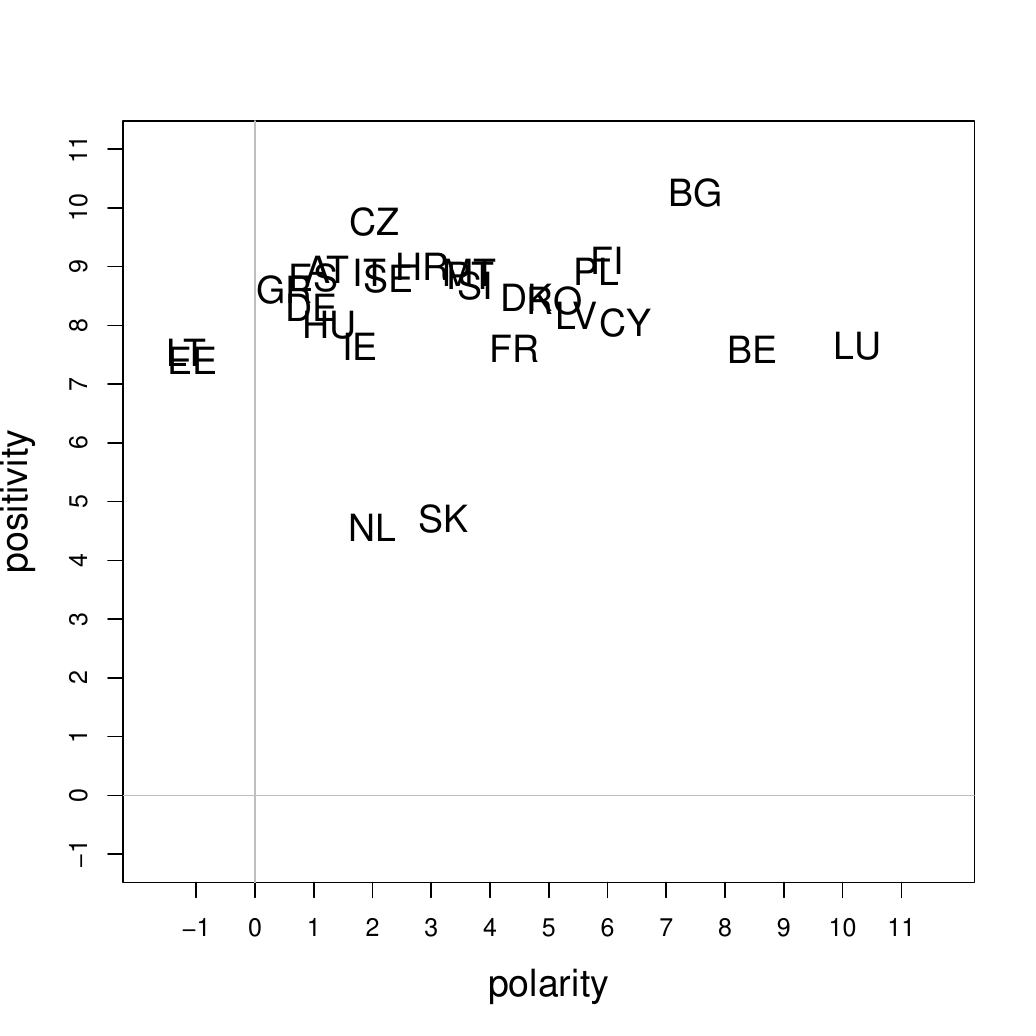}
\end{center}
\caption{AI data. Scatter-plot of the standardized distances from mean of the baseline depth for polarity (x-axis) and positivity (y-axis), for the 27 EU countries. 
\label{fig:AI_scores}}
\end{figure}

Fig~\ref{fig:AI_scores} shows the standardized distances of the depth of the AI curves from the means of the depths of the resampled baseline curves. All of the countries have a high score on positivity, which derives from a positivity curve for AI always above the resampled baseline curves, reflecting the qualitative observation that baseline and AI data strongly differ on positivity. The scores of polarity range from very low values (such as Lithuania, LT, and Estonia, EE), signaling a polarity curve for AI very similar to the baseline, to high values (such as Belgium, BE, and Luxembourg, LU), where there is a strong difference between the AI curve and the baseline. This dispersion reflects the larger variability in the polarity curves of the AI data with respect to the baseline data, as qualitatively observed in Fig~\ref{fig:AI_sbp_smooth}.

More details on the results on AI data are reported in Appendix~\ref{sec:appendix_intra_1_AI}.
 
\subsection{Quantitative multi-country sentiment analysis for trend significance assessment}
\label{sec:intra_2_AI}

As can be observed in Fig~\ref{fig:AI_sbp_smooth}, the polarity curves for the AI data do not present common temporal patterns, while many positivity curves feature a drop at the beginning of 2023. The same analysis described in Section~\ref{sec:intra_2_energy} is performed here on the first derivative of AI data.

\begin{figure}[tb!]
\begin{center}
\includegraphics[trim= 0 12mm 0 0,clip,width=0.55\textwidth]{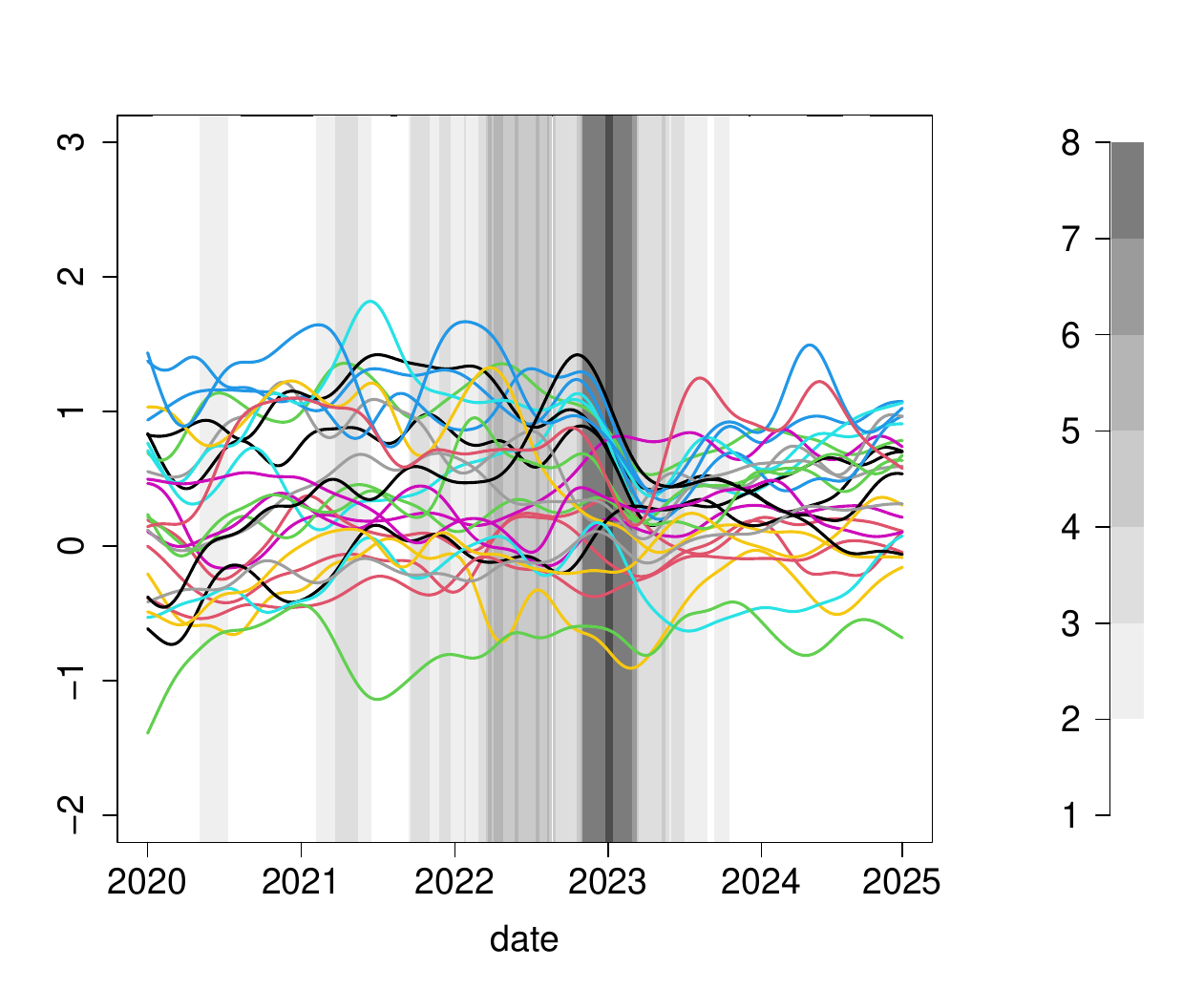}
\end{center}
\caption{AI data. Summary of the results of the bootstrap tests on the first derivatives of the positivity curves for all the countries. Gray background colors are proportional to the count of countries for which a significant difference has been found.
\label{fig:ALL_AI_derivative}}
\end{figure}



Fig~\ref{fig:ALL_AI_derivative} summarizes the results of the bootstrap tests for all the countries (as in Fig~\ref{fig:ALL_energy_derivative}), while additional examples are provided in Appendix~\ref{sec:appendix_intra_1_AI}. We obtain smaller counts with respect to the energy data, signaling that the effect we observe is less common or less pronounced in some countries. The largest counts observed (8) occur between the end of 2022 and the beginning of 2023, corresponding to an increase in popularity and debate of the topic due to the advancements in generative AI and availability of user-friendly AI applications (such as the release of OpenAI's ChatGPT) and the related concerns emerging in the news and in the state of AI report \citep{stateof.ai2022}.

Other events connected to AI, like all the events connected to the AI act (its publication in June 2024 or the European Parliament adoption of its negotiating position in June 2023), or the release of GPT3.5, did not result in significant signals in the data considered, suggesting that these events did not trigger a shift in the sentiment of the news landscape.

\begin{figure}[tb!]
\begin{center}
\includegraphics[trim= 0 12mm 0 60,clip,width=0.6\textwidth]{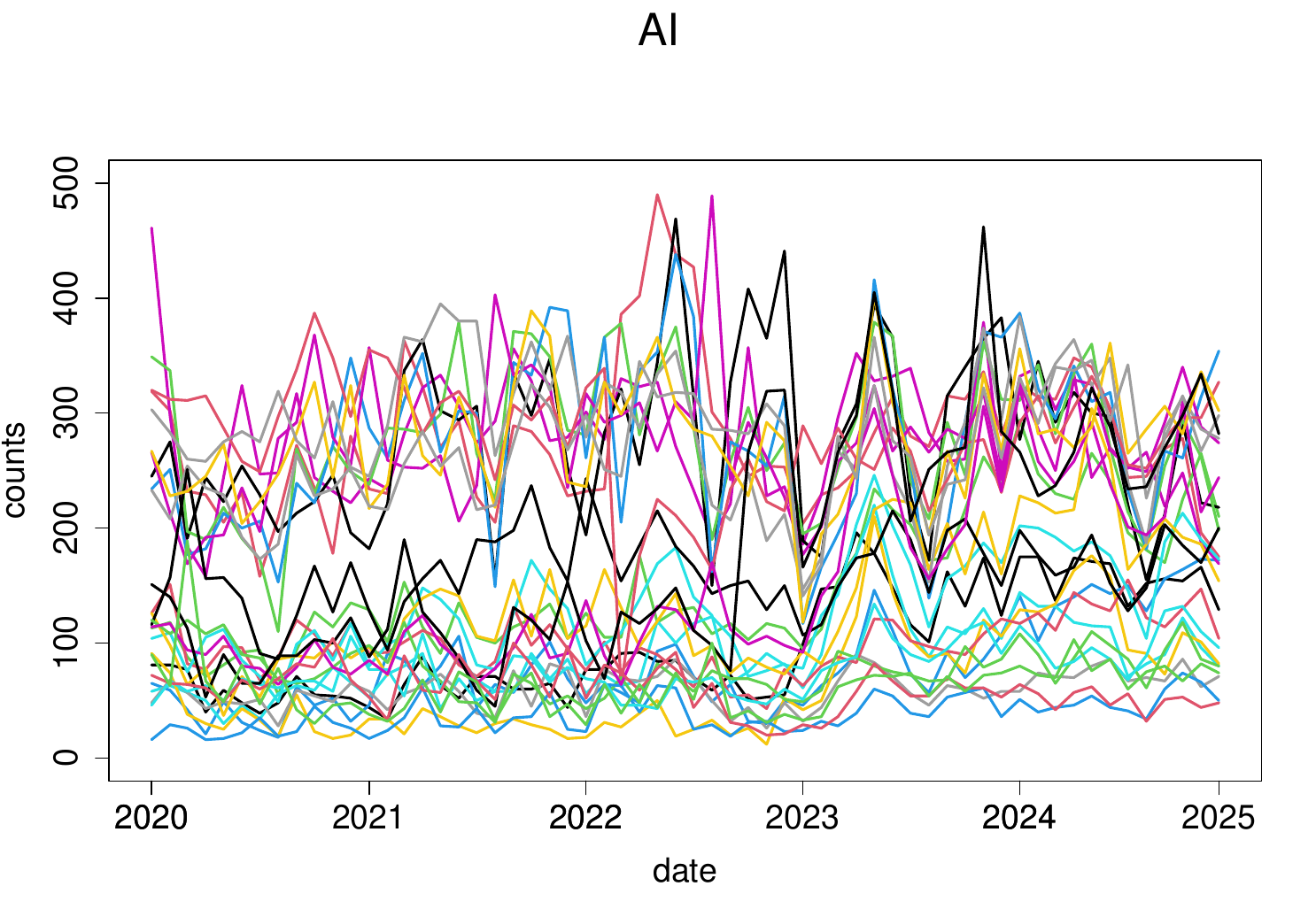}
\end{center}
\caption{AI data. Total counts of news articles for the 27 EU countries. 
\label{fig:AI_total_counts}}
\end{figure}

Differently than in the energy case study, where there were peaks in the total counts of news, here the significant change observed in positivity curves is not accompanied by a strong signal in the time series of total counts of news articles about AI, as shown in~Fig~\ref{fig:AI_total_counts}. 

\section{Discussion}
\label{sec:disc}

This study proposes \textit{sentiment sensing} as a framework for post-classification analysis of sentiment data and demonstrates how it addresses several methodological limitations that recur in applied sentiment analysis.

Cross-language analyses are complicated by systematic differences in classifier performance across languages and by cultural variation in media tone. Standard evaluation metrics such as accuracy or F1 score quantify classification quality but do not directly address how such biases affect downstream comparative analyses.
The comparative baseline design introduced here mitigates these effects by assessing each country’s thematic sentiment relative to its own contemporaneous baseline. This controls for stable language-specific or cultural biases and enables cross-country comparisons to focus on deviations that are specific to the topic of interest. The energy case study illustrates that this design yields interpretable results even when applied to sentiment labels generated by models that qualitatively exhibit language-specific biases, while improved models further enhance data quality without altering the validity of the comparative approach.

Sentiment proportions constitute compositional data and are therefore subject to the constant-sum constraint, which violates assumptions underlying many standard statistical procedures. While this property is well understood in compositional data analysis, it is often not accounted for in sentiment studies, which commonly analyze raw proportions or derived summary scores.
By operating in a logratio-transformed space, the proposed framework enables statistically valid inference while preserving interpretability through polarity and positivity indices. The consistency between these indices and the structure revealed by correspondence analysis suggests that the transformations capture intrinsic relationships among sentiment categories rather than artifacts of aggregation.

The functional data approach smooths the temporal evolution, thus capturing long-term patterns with strong temporal persistence while filtering out single-month fluctuations. Appendix \ref{app:finitedifferences} presents the trend‑significance analysis of the unsmoothed monthly AI data, applying the bootstrap test to the finite differences rather than to the pointwise evaluations of the smooth first derivative (as done in Section \ref{sec:intra_2_AI}). The results show that signals identified as significant by the functional‑data method are missed when the raw monthly data are analysed, because of the high noise level in the original data. Conversely, the functional approach does not capture brief, month‑specific spikes, as it is not intended for fine‑grained temporal resolution. Such isolated fluctuations are detected in the analysis performed on the raw monthly data, although their statistical significance may be affected by the same noise that masks longer-term trends. When the expected temporal scale of the phenomenon under study requires a finer temporal detail, the proposed methodology should be applied to weekly or daily aggregates, provided that enough data are available to perform the analysis at those resolutions.

Temporal sentiment analyses frequently rely on descriptive inspection of trends or on informal interpretations of time series plots. In contrast, the present framework integrates formal hypothesis testing and changepoint detection to identify statistically significant temporal shifts.
In the case studies, this approach identifies mid-2021 as a significant shift point for energy sentiment and late-2022 to early-2023 for AI sentiment in multiple countries. These time points align with major external developments, but the use of formal tests ensures that the reported shifts are reproducible and not dependent on subjective selection.

Analyzing sentiment trajectories in isolation makes it difficult to distinguish topic-specific effects from broader changes in media tone, platform dynamics, or seasonal patterns. The baseline-referenced design introduced here addresses this limitation by contrasting thematic corpora against contemporaneous reference corpora. This setting ensures that the detected signals are specific of the considered topic and not induced by other factors that could affect the overall media landscape.

The analysis of the temporal evolution of the sentiment complements and characterizes patterns that can be detected in the total count of news over time.
For example, the analysis reveals that AI sentiment changed without a corresponding increase in publication volume, indicating a reframing of coverage rather than a change in attention, while the sentiment change for the energy topic corresponds to a rise of the popularity of the topic in the media. Such effects are not visible in frequency-based analyses and illustrate the value of comparative post-classification analysis.

Previous studies have addressed elements of multilingual sentiment analysis, compositional data treatment, or temporal modeling, but typically in isolation. For example, large-scale multilingual analyses emphasize model training and evaluation but often rely on simple aggregation of outputs \citep{wang2022global,farquharson2025multilingual,RODRIGUEZIBANEZ2023review}; temporal studies introduce regression or changepoint methods but are frequently limited to single languages \citep{rozado2022longitudinal,wang2022global,farquharson2025multilingual}; and compositional approaches demonstrate appropriate transformations without embedding them in a broader analytical framework  \citep{10.1007/978-3-031-83705-0_13,10.3389/fpubh.2025.1597941}.
\textit{Sentiment sensing} integrates these components into a unified post-classification framework that is multilingual, compositional, comparative, and temporally explicit. Importantly, the framework is model-agnostic and domain-agnostic: it can be applied to any sentiment-annotated dataset with temporal metadata and an appropriate baseline corpus.

Although demonstrated on news corpora, the proposed framework is applicable to any setting where sentiment-annotated texts can be organized into reference and thematic corpora with temporal structure. Potential applications include disinformation analysis, political communication, crisis response, brand monitoring, and policy evaluation. The minimal requirements are sentiment annotations, timestamps, and a comparative corpus design.

Several limitations should be noted. First, the smoothing procedures used in the functional data analysis may introduce edge effects, limiting the suitability of the current implementation for real-time early-warning applications. Future work should explore alternative smoothing and online detection methods optimized for streaming data.
Second, bootstrap resampling and functional methods are computationally intensive and may require optimization for large-scale or real-time deployment.
Third, while temporal alignment with external events is suggestive, the framework does not establish causality. Integration with causal inference approaches such as synthetic control or quasi-experimental designs could strengthen event attribution.
Fourth, the comparison with a same-language baseline relies on the assumption, adopted in prior work (e.g. \cite{tao2025academic}), that sentiment performance is not influenced by topic variation. However, since different topics may be characterized by distinct linguistic features that affect sentiment label assignment \citep{kellert2024unveiling}, we aim to mitigate potential topic-induced bias via baseline resampling.
Finally, statistical significance does not necessarily imply substantive importance. Future research should explore effect size measures and domain-specific interpretation frameworks to improve the practical relevance of detected changes.


\section{Conclusion}
\label{sec:conc}

This study introduces \textit{sentiment sensing} as a framework for rigorous post-classification analysis of multilingual sentiment data. While advances in sentiment classification have enabled large-scale annotation across languages and domains, analytical practices have often relied on simple aggregation and descriptive interpretation. The proposed framework addresses this gap by integrating compositional data analysis, comparative baseline design, formal temporal inference, and multilingual applicability into a unified framework.

The case studies demonstrate that this approach reveals sentiment dynamics that are aligned with relevant events, including topic-specific reframing without concurrent change in volume of articles and statistically significant temporal shifts across countries. These findings arise from how sentiment labels are analyzed rather than from changes in how they are generated.

By providing a statistically defensible, reproducible, and interpretable post-classification analysis framework, \textit{sentiment sensing} complements existing sentiment analysis techniques and supports more reliable comparative research across time, languages, and domains. As sentiment analysis continues to be applied in diverse policy, scientific, and commercial contexts, methodological rigor in post-classification analysis will be essential for ensuring that inferences drawn from sentiment data are trustworthy and meaningful.

\section*{Declaration of Generative AI use}
The authors used LLMs via the GPT@JRC Institutional test bed (now called GENESIS) \cite{de2025proof} to rephrase parts of the article (especially abstract, introduction, related work and conclusion), and to carry out literature research. Kiro with Claude Opus 4.5, institutionally accessed, was used to generate Fig~\ref{fig:diagram}. Fig~\ref{fig:diagram_intro} is generated with ChatGPT. After using these tools, the authors reviewed and edited the content as needed and take full responsibility for the publication’s content. 

\section*{Data availability statement}
To allow reproducibility of the results presented in this article, the data and code used are available here: \url{https://github.com/MaraBernardiJRC/SentimentSensing}. The news dataset are distributed as aggregates, as they are protected by copyright.

\section*{Acknowledgments}
The work of the authors aligns with the institutional program of the Joint Research Centre (JRC) of the European Commission, under Horizon Europe (2021-2027).

\bibliography{bibliography.bib}

\begin{thebibliography}{}

\bibitem[A~Semary et~al., 2024]{semary2024sentimentPlos}
A~Semary, N., Ahmed, W., Amin, K., P{\l}awiak, P., and Hammad, M. (2024).
\newblock Enhancing machine learning-based sentiment analysis through feature extraction techniques.
\newblock {\em PLOS One}, 19(2):e0294968.

\bibitem[Barbieri et~al., 2022]{barbieri-etal-2022-xlm}
Barbieri, F., Espinosa~Anke, L., and Camacho-Collados, J. (2022).
\newblock {XLM}-{T}: Multilingual language models in {T}witter for sentiment analysis and beyond.
\newblock In Calzolari, N., B{\'e}chet, F., Blache, P., Choukri, K., Cieri, C., Declerck, T., Goggi, S., Isahara, H., Maegaard, B., Mariani, J., Mazo, H., Odijk, J., and Piperidis, S., editors, {\em Proceedings of the Thirteenth Language Resources and Evaluation Conference}, pages 258--266, Marseille, France. European Language Resources Association.

\bibitem[Benaich and Hogarth, 2022]{stateof.ai2022}
Benaich, N. and Hogarth, I. (2022).
\newblock State of ai 2022.
\newblock \url{https://www.stateof.ai/2022}.
\newblock Accessed: 2025-10-23.

\bibitem[Conneau et~al., 2020]{conneau-etal-2020-unsupervised}
Conneau, A., Khandelwal, K., Goyal, N., Chaudhary, V., Wenzek, G., Guzm{\'a}n, F., Grave, E., Ott, M., Zettlemoyer, L., and Stoyanov, V. (2020).
\newblock Unsupervised cross-lingual representation learning at scale.
\newblock In Jurafsky, D., Chai, J., Schluter, N., and Tetreault, J., editors, {\em Proceedings of the 58th Annual Meeting of the Association for Computational Linguistics}, pages 8440--8451, Online. Association for Computational Linguistics.

\bibitem[De~Longueville et~al., 2025]{de2025proof}
De~Longueville, B., Sanchez, I., Kazakova, S., Luoni, S., Zaro, F., Daskalaki, K., and Inchingolo, M. (2025).
\newblock The proof is in the eating: Lessons learnt from one year of generative ai adoption in a science-for-policy organisation.
\newblock {\em AI}, 6(6):128.

\bibitem[Di~Nuovo et~al., 2026]{dinuovo2026sentimentbias}
Di~Nuovo, E., Bernardi, M., Colla, D., Stefanovitch, N., Perrotta, D., Torti, F., Longueville, B.~D., and Macmillan, C. ((in press) 2026).
\newblock Multilingual sentiment analysis: identifying and addressing language-dependent bias without the need for gold labels.
\newblock In {\em Proceedings of the Twelveth Italian Conference on Computational Linguistics (CLiC-it 2026)}.

\bibitem[Di~Nuovo et~al., 2024]{10.1007/978-3-031-70242-6_3}
Di~Nuovo, E., Cartier, E., and De~Longueville, B. (2024).
\newblock Meet xlm-rlnews-8: Not just another sentiment analysis model.
\newblock In {\em Natural Language Processing and Information Systems: 29th International Conference on Applications of Natural Language to Information Systems, NLDB 2024, Turin, Italy, June 25–27, 2024, Proceedings, Part II}, page 24–35, Berlin, Heidelberg. Springer-Verlag.

\bibitem[{Directorate-General for Economic and Financial Affairs}, 2021]{EC2021report}
{Directorate-General for Economic and Financial Affairs} (2021).
\newblock European economic forecast: Autumn 2021.
\newblock European Economy Institutional Papers 160, European Commission, Luxembourg.
\newblock Accessed: 2025-10-16.

\bibitem[{Directorate-General for Economic and Financial Affairs}, 2022]{EC2022report}
{Directorate-General for Economic and Financial Affairs} (2022).
\newblock European economic forecast: Summer 2022.
\newblock European Economy Institutional Papers 183, European Commission, Luxembourg.
\newblock Accessed: 2025-10-16.

\bibitem[Farquharson, 2025]{farquharson2025multilingual}
Farquharson, J.~I. (2025).
\newblock Multilingual social media analysis reveals global patterns and language imbalances in volcanic eruption coverage.
\newblock {\em Communications Earth \& Environment}, 6(1):777.

\bibitem[Granger, 1969]{granger1969causalrelations}
Granger, C. W.~J. (1969).
\newblock Investigating causal relations by econometric models and cross-spectral methods.
\newblock {\em Econometrica}, 37(3):424--438.

\bibitem[Greenacre, 2017]{greenacre2017}
Greenacre, M. (2017).
\newblock {\em Correspondence analysis in practice}.
\newblock 3rd edition. BocaRaton, FL: Chapman and Hall CRC Press.

\bibitem[Haines et~al., 2025]{10.3389/fpubh.2025.1597941}
Haines, H., Sursely, A., Pham, H., Beltran, R., Sewell, D.~K., Reinders, L.~T., Askelson, N., Dixon, S., Emrich, K., Estle, C., Lewis, M., Parker, E., Reyes, J., Walkner, L., Sittig, K.~W., and Afifi, R.~A. (2025).
\newblock Insights from the front line: uplifting stories of the covid-19 pandemic through the eyes of the public health workforce in iowa.
\newblock {\em Frontiers in Public Health}, Volume 13 - 2025.

\bibitem[Hartmann et~al., 2023]{HARTMANN202375}
Hartmann, J., Heitmann, M., Siebert, C., and Schamp, C. (2023).
\newblock More than a feeling: Accuracy and application of sentiment analysis.
\newblock {\em International Journal of Research in Marketing}, 40(1):75--87.

\bibitem[Ieva et~al., 2019]{roahd}
Ieva, F., Paganoni, A.~M., Romo, J., and Tarabelloni, N. (2019).
\newblock {roahd Package: Robust Analysis of High Dimensional Data}.
\newblock {\em {The R Journal}}, 11(2):291--307.

\bibitem[Jabbari~Tofighi and Alhajj, 2025]{jabbari2025sentimentPlos}
Jabbari~Tofighi, N. and Alhajj, R. (2025).
\newblock Investigating the impact of social media images on users’ sentiments towards sociopolitical events based on deep artificial intelligence.
\newblock {\em PLOS One}, 20(7):e0326936.

\bibitem[Kellert et~al., 2024]{kellert2024unveiling}
Kellert, O., G{\'o}mez-Rodr{\'\i}guez, C., and Uz~Zaman, M. (2024).
\newblock Unveiling factors influencing judgment variation in sentiment analysis with natural language processing and statistics.
\newblock {\em Plos one}, 19(5):e0304201.

\bibitem[Kryvasheyeu et~al., 2015]{kryvasheyeu2015performance}
Kryvasheyeu, Y., Chen, H., Moro, E., Van~Hentenryck, P., and Cebrian, M. (2015).
\newblock Performance of social network sensors during hurricane sandy.
\newblock {\em PLOS One}, 10(2):e0117288.

\bibitem[L{\'o}pez-Pintado and Romo, 2007]{lopez2007depth}
L{\'o}pez-Pintado, S. and Romo, J. (2007).
\newblock Depth-based inference for functional data.
\newblock {\em Computational Statistics \& Data Analysis}, 51(10):4957--4968.

\bibitem[L{\'o}pez-Pintado and Romo, 2009]{lopez2009concept}
L{\'o}pez-Pintado, S. and Romo, J. (2009).
\newblock On the concept of depth for functional data.
\newblock {\em Journal of the American statistical Association}, 104(486):718--734.

\bibitem[Marti-Ochoa et~al., 2025]{10.1007/978-3-031-83705-0_13}
Marti-Ochoa, J., Ferrer-Rosell, B., and Martin-Fuentes, E. (2025).
\newblock Analysis of sentiments in airbnb experiences.
\newblock In Nixon, L., Tuomi, A., and O'Connor, P., editors, {\em Information and Communication Technologies in Tourism 2025}, pages 155--164, Cham. Springer Nature Switzerland.

\bibitem[Mozeti{\v{c}} et~al., 2016]{mozetivc2016multilingual}
Mozeti{\v{c}}, I., Gr{\v{c}}ar, M., and Smailovi{\'c}, J. (2016).
\newblock Multilingual twitter sentiment classification: The role of human annotators.
\newblock {\em PLOS One}, 11(5):e0155036.

\bibitem[Pawlowsky-Glahn et~al., 2015]{pawlowsky2015modeling}
Pawlowsky-Glahn, V., Egozcue, J.~J., and Tolosana-Delgado, R. (2015).
\newblock {\em Modeling and analysis of compositional data}.
\newblock John Wiley \& Sons.

\bibitem[Přibáň et~al., 2024]{PRIBAN2024123247}
Přibáň, P., Šmíd, J., Steinberger, J., and Mištera, A. (2024).
\newblock A comparative study of cross-lingual sentiment analysis.
\newblock {\em Expert Systems with Applications}, 247:123247.

\bibitem[Riani et~al., 2022]{riani2022}
Riani, M., Atkinson, A.~C., Torti, F., and Corbellini, A. (2022).
\newblock Robust correspondence analysis.
\newblock {\em Journal of the Royal Statistical Society Series C: Applied Statistics}, 71(5):1381--1401.

\bibitem[Rodríguez-Ibánez et~al., 2023]{RODRIGUEZIBANEZ2023review}
Rodríguez-Ibánez, M., Casánez-Ventura, A., Castejón-Mateos, F., and Cuenca-Jiménez, P.-M. (2023).
\newblock A review on sentiment analysis from social media platforms.
\newblock {\em Expert Systems with Applications}, 223:119862.

\bibitem[Rozado et~al., 2022]{rozado2022longitudinal}
Rozado, D., Hughes, R., and Halberstadt, J. (2022).
\newblock Longitudinal analysis of sentiment and emotion in news media headlines using automated labelling with transformer language models.
\newblock {\em PLOS One}, 17(10):e0276367.

\bibitem[Stefanovitch et~al., 2022]{stefanovitch-etal-2022-resources}
Stefanovitch, N., Piskorski, J., and Kharazi, S. (2022).
\newblock Resources and experiments on sentiment classification for {G}eorgian.
\newblock In Calzolari, N., B{\'e}chet, F., Blache, P., Choukri, K., Cieri, C., Declerck, T., Goggi, S., Isahara, H., Maegaard, B., Mariani, J., Mazo, H., Odijk, J., and Piperidis, S., editors, {\em Proceedings of the Thirteenth Language Resources and Evaluation Conference}, pages 1613--1621, Marseille, France. European Language Resources Association.

\bibitem[Steinberger et~al., 2015]{EMM2015}
Steinberger, R., Podavini, A., Balahur, A., Jacquet, G., Tanev, H., Linge, J., Atkinson, M., Chinosi, M., Zavarella, V., Steiner, Y., and van~der Goot, E. (2015).
\newblock Observing trends in automated multilingual media analysis.
\newblock In {\em Proceedings of the Symposium on New Frontiers of Automated Content Analysisin the Social Sciences (ACA'2015)}, pages 1--20, Zürich, Switzerland.

\bibitem[Tao and Shen, 2025]{tao2025academic}
Tao, Y. and Shen, Q. (2025).
\newblock Academic discourse on chatgpt in social sciences: A topic modeling and sentiment analysis of research article abstracts.
\newblock {\em PLOS One}, 20(10):e0334331.

\bibitem[Wang et~al., 2022]{wang2022global}
Wang, J., Fan, Y., Palacios, J., Chai, Y., Guetta-Jeanrenaud, N., Obradovich, N., Zhou, C., and Zheng, S. (2022).
\newblock Global evidence of expressed sentiment alterations during the covid-19 pandemic.
\newblock {\em Nature Human Behaviour}, 6(3):349--358.

\bibitem[Yan and Liu, 2024]{yan2024has}
Yan, T. and Liu, F. (2024).
\newblock Has sentiment returned to the pre-pandemic level? a sentiment analysis using us college subreddit data from 2019 to 2022.
\newblock {\em PLOS One}, 19(3):e0299837.

\bibitem[Youvan, 2024]{youvan2024understanding}
Youvan, D.~C. (2024).
\newblock Understanding sentiment analysis with vader: a comprehensive overview and application.

\bibitem[Zhao et~al., 2024]{10.1145/3645106}
Zhao, C., Wu, M., Yang, X., Zhang, W., Zhang, S., Wang, S., and Li, D. (2024).
\newblock A systematic review of cross-lingual sentiment analysis: Tasks, strategies, and prospects.
\newblock {\em ACM Comput. Surv.}, 56(7).

\end{thebibliography}

\newpage

\appendix
\label{sec:appendix}

\section{Corpus collection}
\label{sec:appendix_data_collection}

We analyze data collected from the Europe Media Monitor pipeline \citep{EMM2015}, which continuously gathers and analyses news items from around the world. We use a selection of around 500 mainstream, influential media sources from the 27 EU countries, collecting the following datasets.
\begin{itemize}
    \item The \emph{energy data}, a set of thematic news items related to the topic of energy. In this case, the news articles are collected by looking for the presence of multilingual keywords that define the specific topic (e.g. ``energy'', ``energija'', ``courant électrique''). In addition, English keywords (e.g. ``energy'', ``gas'', ``electricity'') were searched in the English translation of the news. For this dataset, we consider the time period from 1-1-2020 to 12-31-2022.
    \item The \emph{AI data} contains a set of thematic news items related to the topic of AI. In this case, the news are collected by looking for the presence of multilingual keywords that define the specific topic (e.g. ``artificial intelligence'', ``künstliche Intelligenz'', ``inteligencia artificial''), and selected English keywords in the English translation (i.e. ``AI'', ``artificial intelligence''). For this dataset, we consider the time period from 1-1-2020 to 12-15-2024.
    \item The \emph{baseline data}, a set of randomly selected news items covering the periods used for both the energy and AI data, from 1-1-2020 to 12-15-2024. For the energy case study we use only data up to 12-31-2022.
\end{itemize}

The number of news items sampled per country is chosen to have a maximum of 10,000 news items over a three-year period, i.e., about 3,333 news per year. In the thematic datasets, the query returns fewer than this maximum number of news pieces in some of the smaller countries (such as Malta, Latvia and Luxembourg). 
Table~\ref{tab:keywords} reports the full list of keywords used to search the news articles.

\setcounter{table}{0} 
\begin{longtable}{@{} l L @{}}
\toprule
\textbf{Dataset} & \textbf{Keywords} \\
\midrule
\endfirsthead                     
\toprule
\textbf{Dataset} & \textbf{Keywords} \\
\midrule
\endhead
\midrule
\multicolumn{2}{r}{\textit{continued on next page}}\\
\bottomrule
\endfoot
\bottomrule
\endlastfoot

Energy &
``energy'' ``energiya'' (BG) ``energie'' ``energi'' ``energieversgung'' ``energeia'' (EL) ``energ\'ia'' ``energia'' ``\'energie'' ``fuinneamh'' ``energija'' ``l'energia'' ``energijos'' ``energetikis'' (LV) ``energija'' (MT) ``gas'' ``gaz'' (BG) ``plynov\'y'' ``gase'' ``aerio'' (EL) ``gaas'' ``kaasu'' ``gaz'' ``g\'as'' ``plin'' ``g\'az'' ``duju'' (LT) ``gazes'' (LV) ``gaze'' (LV) ``gass'' ``gaze'' ``plyn'' ``plinski'' ``heating'' ``otoplenie'' (BG) ``vyt\'ap\v{e}n\'i'' ``opvarmning'' ``Heizung'' ``Thermansi'' (EL) ``calefacci\'on'' ``K\"ute'' ``L\"ammitys'' ``chauffage'' ``t\'eamh'' ``grijanje'' ``f\H{u}t\'es'' ``riscaldamento'' ``\v{s}ildymas'' ``apkures sistemas'' (LV) ``tishin'' (MT) ``verwarmen'' ``ogrzewanie'' ``aquecimento'' ``\^inc\u{a}lzire'' ``vykurovanie'' ``ogrevanje'' ``uppv\"armning'' ``electricity'' ``elektrichestvo'' (BG) ``elekt\v{r}ina'' ``elektricitet'' ``elektrische energie'' ``ilektriki energeia'' (EL) ``electricidad'' ``elekter'' ``s\"ahk\"o'' ``\'electricit\'e'' ``leictreachas'' ``elektri\v{c}na energija'' ``energia elettrica'' ``elektros energija'' ``elektriba'' (LV) ``elettriku'' ``elektriciteit'' ``elektryczno\'s\'c'' ``eletricidade'' ``electricitate'' ``elektrina'' ``elektrika'' ``electricitet'' ``electric current'' ``elektricheski tok'' (BG) ``elektrick\'y proud'' ``elektrisk str{\o}m'' ``elektrischer Strom'' ``ilektriko revma'' (EL) ``criente el\'ectrica'' ``elektrivool'' ``s\"ahk\"ovirta'' ``courant \'electrique'' ``leictreach at\'a ann faoi l\'athair'' ``elektri\v{c}na struja'' ``elektromos \'aram'' ``crente elettrica'' ``elektros srove'' (LT) ``elektriska strava'' (LV) ``kurrent elettriku'' ``elektrische stroom'' ``prad elektryczny'' (PL) ``crente el\'etrica'' ``curent electric'' ``elektrick\'y pr\'ud'' ``elektri\v{c}ni tok'' ``elektrisk str\"om'' ``light'' ``svetlina'' (BG) ``sv\v{e}tlo'' ``lys'' ``licht'' ``fos'' (EL) ``luz'' ``valgus'' ``valo'' ``lumi\`ere'' ``solas'' ``svjetlo'' ``vil\'ag\'it\'as'' ``luce'' ``\v{s}viesos'' ``gaismas lukturis'' ``dawl'' ``licht'' ``\'swiat{\l}o'' ``lumin\u{a}'' ``svetlo'' ``svetloba'' ``ljus'' ``fuel'' ``gorivo'' (BG) ``pohonn\'e hmoty'' ``br{\ae}ndstof'' ``Kraftstoffe'' ``kafsima'' (EL) ``combustible'' ``k\"utus'' ``polttoaine'' ``carburant'' ``breosla'' ``givo'' ``\"uzemanyag'' ``carburante'' ``kuras'' ``degvielas uzpildes stacijas'' ``fjuwil'' ``brandstof'' ``paliwo'' ``combust\'ivel'' ``combustibil'' ``palivo'' ``br\"ansle'' ``gasoline'' ``benzin'' (BG) ``benz\'in'' ``benzin'' ``Benzin'' ``venzini'' (EL) ``gasolina'' ``bensiin'' ``bensiini'' ``essence'' ``g\'asail\'in'' ``benzina'' ``benzinas'' ``benzins'' (LV) ``gazolina'' (MT) ``benzine'' ``benzyna'' ``benzin\u{a}'' ``bencin'' ``bensin'' ``diesel'' ``dizel'' (BG) ``nafta'' ``Dieselkraftstoff'' ``ntizel'' (EL) ``di\'esel'' ``diislik\"utus'' ``d\'iosal'' ``dizel'' ``d\'izel'' ``gasolio'' ``Dyzelinas'' ``Dizeldegviela'' (LV) ``dizil'' (MT) ``olej nap\k{e}dowy'' ``gas\'oleo'' ``motin\u{a}'' ``dizelsko givo'' \\

\midrule

AI &
``ai'' ``artificial intelligence'' ``izkustven intelekt'' (BG) ``um\v{e}l\'a inteligence'' ``kunstig intelligens'' ``k\"unstliche Intelligenz'' ``techniti noimosini'' (EL) ``inteligencia artificial'' ``tehisintellekt'' ``teko\"aly'' ``intelligence artificielle'' ``intleacht shaorga'' ``umjetna inteligencija'' ``mesters\'eges intelligencia'' ``intelligenza artificiale'' ``dirbtinis intelektas'' ``maksl\=igais intelekts'' (LV) ``intelligenza artificjali'' (MT) ``kunstmatige intelligentie'' ``sztuczna inteligencja'' ``intelig\^encia artificial'' ``inteligenta artificiala'' (RO) ``umel\'a inteligencia'' ``umetna inteligenca'' ``artificiell intelligens'' \\

\midrule

baseline & * \\

\bottomrule
\caption{\label{tab:keywords}Keywords used to search news on the thematic and baseline data.}
\end{longtable}


\vspace{2em}
For each news item, we consider title and lead-line. The latter, containing about 300 characters, is included to capture the overall tonality of the article by adding context to the title, without relying on the entire news piece. 

The collected data mainly consist of news in one predominant language per country, with very few exceptions such as Luxembourg. For simplicity, in this article we compare sentiment across countries, although the framework supports a comparison across languages, language-country pairs or indeed any populations.

\section{Sentiment annotation}
\label{app:dataAnn}

Having collected the corpora, we have the raw news data, consisting of the baseline and two thematic corpora. These are enriched by automatically annotating them with a sentiment label (Positive, Negative or Neutral), using a news-domain-adapted sentiment analysis model. 

Our first choice of multilingual news sentiment analysis model is XLM-RLnews-8 \citep{10.1007/978-3-031-70242-6_3}, a sentiment classifier based on XLM-RoBERTa-Large \citep{conneau-etal-2020-unsupervised}, which has been domain-adapted on multilingual news data and then fine-tuned using the Unified Multilingual Sentiment Analysis Benchmark (UMSAB) \citep{barbieri-etal-2022-xlm}. 

During the course of this work, it became apparent that XLM-RLnews-8 has a language specific bias, which means that text in German is more likely to be classified as neutral than the same message expressed in another language. A set of experiments including comparing labels assigned to texts translated from German to Italian, English and from these languages into German confirmed that the differences were due to model bias rather than cultural differences. The source of the bias was traced to the fine-tuning dataset used, the UMSAB, and a new model, XLM-RLnews-5 was trained on a subset of UMSAB, using the same training notebook in \cite{10.1007/978-3-031-70242-6_3} and selecting the sentiment datasets in English, Spanish, Portuguese, Italian and French. This new model is evaluated on the same datasets in \cite{10.1007/978-3-031-70242-6_3}, achieving comparable results and fixing the bias towards Neutral label for the German language. A complete report on the experiments following the identification of the bias is presented in \cite{dinuovo2026sentimentbias}. 

The importance of \textit{sentiment sensing} is that the framework supports visualizations which can reveal such biases, and allows meaningful comparisons to be made despite the presence of biases. To show that the framework works with biased models as well, we use the biased XLM-RLnews-8 model in the first case study on energy, and the corrected XLM-RLnews-5 model in the second case study on AI.

\section{Functional Principal Component Analysis for inter- country analysis of sentiment}
\label{sec:appendix_PCA}

Here we report the results of the functional Principal Component Analysis (PCA) applied to both baseline and thematic data.

\subsection{PCA: energy case study}
\label{sec:baseline_energy}

Observing the temporal evolution of polarity and positivity for the baseline (represented in Fig~\ref{fig:baseline_sbp_smooth}), we see that, while the overall temporal trend is almost constant for every country, there are differences among countries: for example, looking at the polarity curves, we can see that some countries tend to be more polarized, while other tend to be more neutral. To explore and better understand this characterization of the different countries, we analyze the data via dimensionality reduction obtaining a synthetic, low-dimensional description that eases their qualitative exploration and interpretation.

\begin{figure}[t!]
\begin{minipage}{\textwidth}
\centering\includegraphics[width=0.5\textwidth]{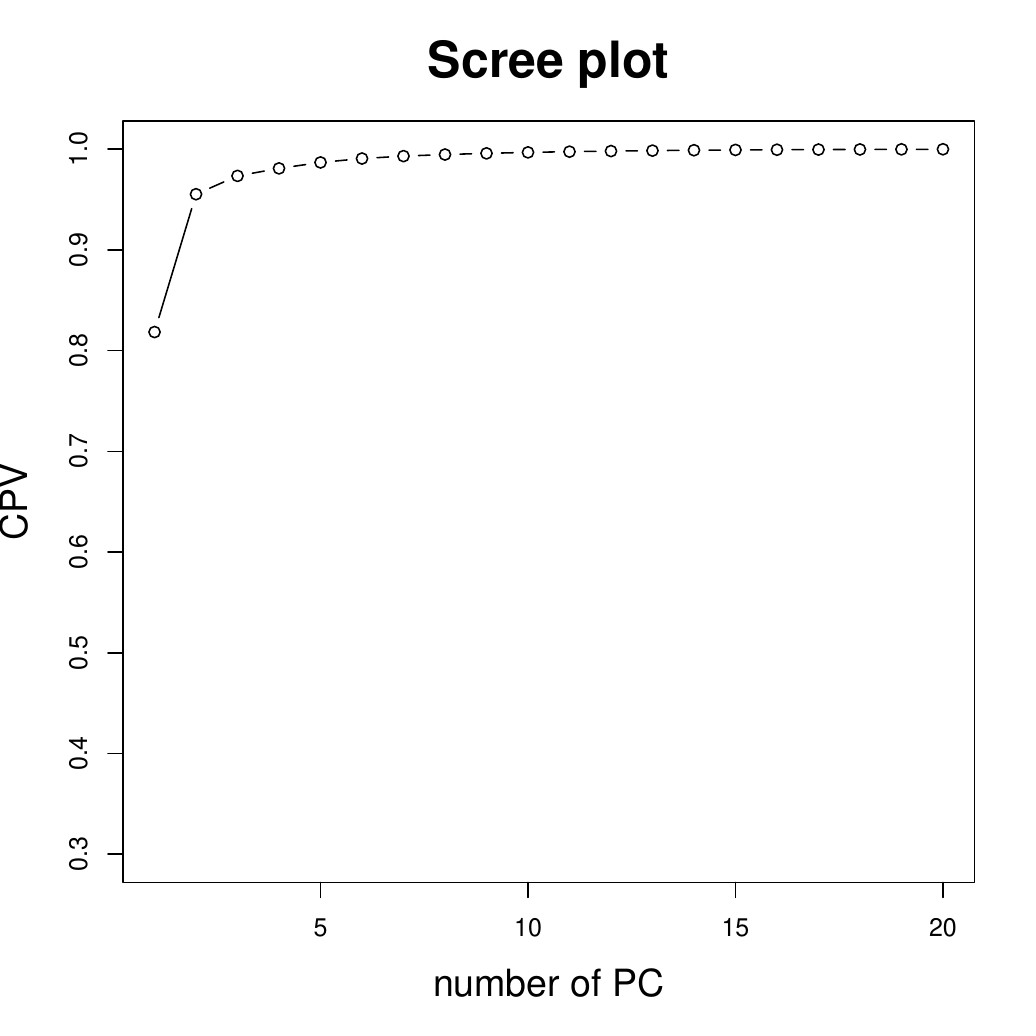}
\caption{XLM-RLnews-8 baseline data. Plot of the Cumulative Proportion of Variance (CPV) versus the number of PCs of the functional PCA on the baseline data.
\label{fig:scree_plot}\vspace{3mm}}
\end{minipage}
%
\begin{minipage}{\textwidth}
\centering\includegraphics[width=\textwidth,trim=0 12mm 0 15mm,clip]{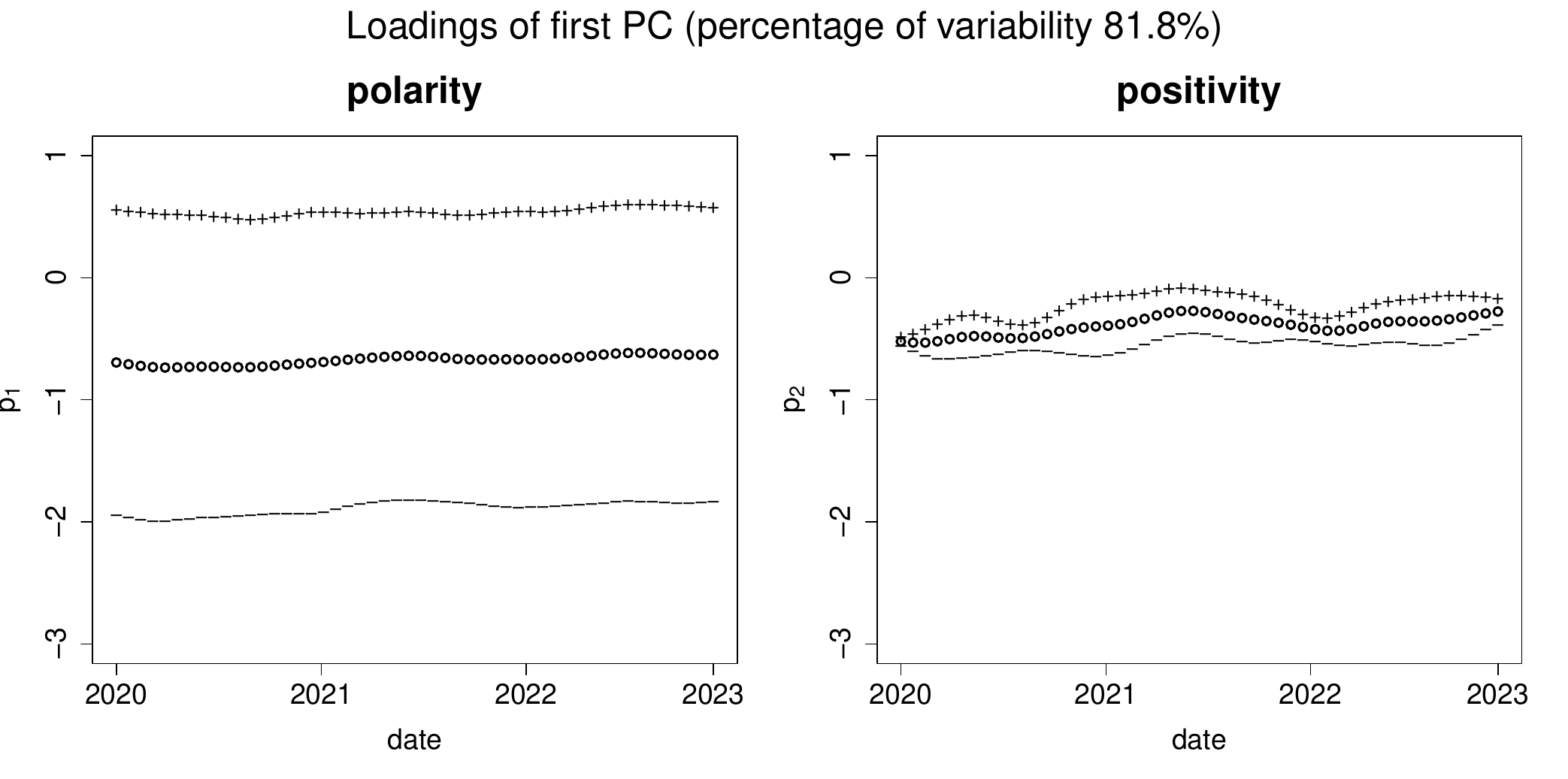}
\caption{XLM-RLnews-8 baseline data. Loadings of the first principal component of the functional PCA (accounting for the 81.8\% of the total variance) represented as perturbations of the mean function (dotted line) obtained by adding (line of + signs) or subtracting (line of - signs) two standard deviations of the first principal component.
\label{fig:loadings_PC1}}
\end{minipage}
\end{figure}

We start with the baseline data, performing dimensionality reduction via functional PCA. We carry out the analysis on the bivariate functional data whose components are polarity and positivity. 
Fig~\ref{fig:scree_plot} shows the obtained scree-plot: the first principal component explains more than 80\% of the variability, meaning that the distribution of the scores along this principal component captures most of the variation characterizing the different countries,  leading to a considerable dimensionality reduction (from two curves to two numbers). Fig~\ref{fig:loadings_PC1} shows the loadings of the first principal component, which accounts for the difference in magnitude of the curves, with high scores corresponding to high polarity and high positivity and low scores corresponding to low polarity and low positivity. 

\begin{figure}[t!]
\begin{center}
\includegraphics[width=0.5\textwidth]{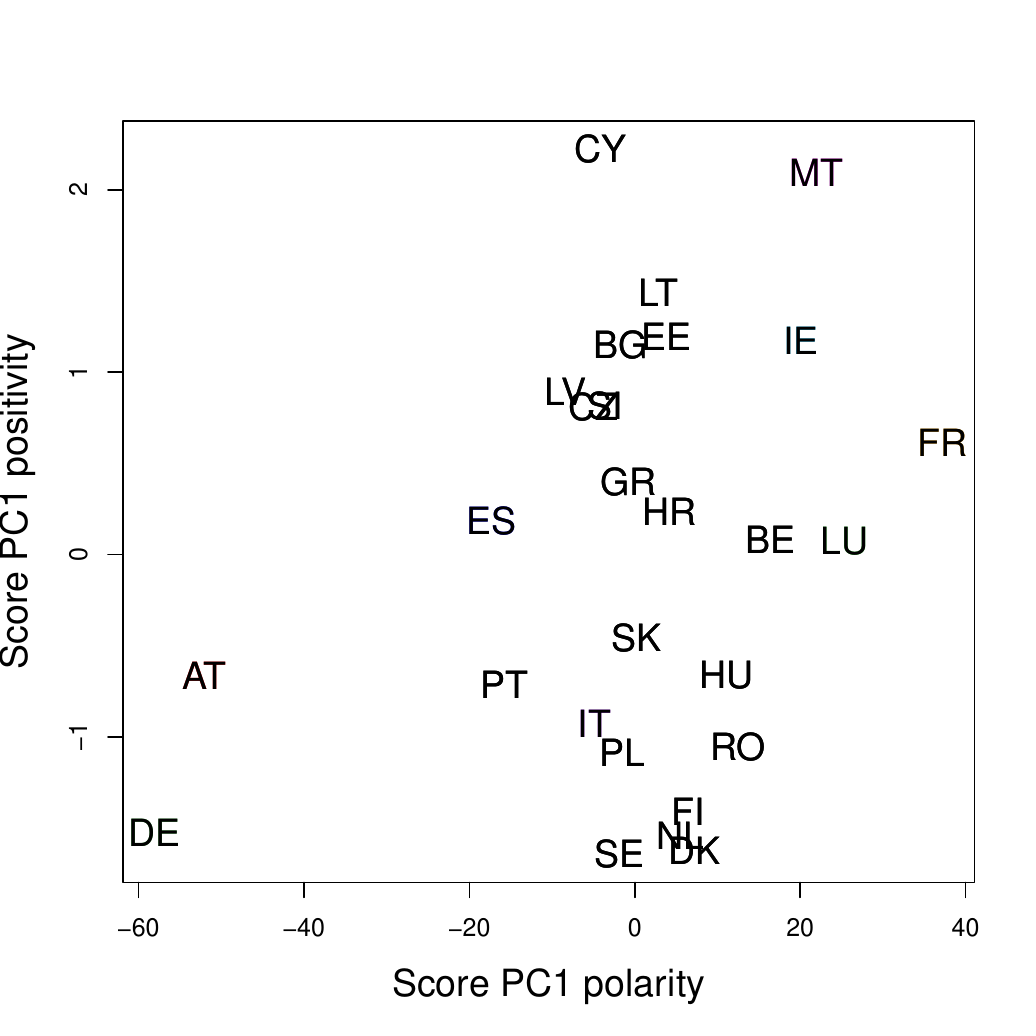}
\end{center}
\caption{XLM-RLnews-8 baseline data. Scatterplot of the scores of the 27 EU countries along the first principal component (polarity dimension on the x axis and positivity dimension on the y axis).
\label{fig:scores_sbp}}
\end{figure}

We organize the polarity and positivity scores along the first principal component in a scatterplot (polarity on the x-axis, positivity on the y-axis) for all the 27 EU countries, as shown in Fig~\ref{fig:scores_sbp}. This two-dimensional plot aims to provide a synthetic representation of the original data, retaining the main characteristic of the countries and providing information about similarities and dissimilarities between them by their relative position.

On the left side of the plot, there are the countries whose media sentiment is characterized by neutrality (such as Germany and Austria), while on the right those characterized by a more polarized media landscape (such as France). Countries with a media landscape characterized by positivity (such as Cyprus and Malta) are on the upper part, and those characterized by negativity (such as the Scandinavian countries) on the lower part. The countries with a media landscape unbiased towards positivity or negativity are in the central part (such as Belgium and Luxembourg). 

Two countries stand out from the rest for their extremely low score on polarity: Germany and Austria. Indeed, as stated in Appendix~\ref{app:dataAnn}, the model used to annotate it, XLM-RLnews-8, has a bias towards neutrality when German data is received. This bias is clearly evident in this synthetic representation. This framework allowed us to spot this bias, which we then tackled and solved by training XLM-RLnews-5, the model that we apply to the AI case study, and, indeed, to re-annotate the baseline. The characterization and resolution of this language-dependent bias is reported in \cite{dinuovo2026sentimentbias}. In this article, we use both a biased and unbiased model to demonstrate that this framework can be effective in two ways: first, in qualitatively spotting biases, and second, in showing that regardless of the model, the quantitative analyses remain valid even with annotations generated by a plausibly biased model.

\begin{figure}[ht]
\begin{minipage}{\textwidth}
\centering\includegraphics[width=0.5\textwidth]{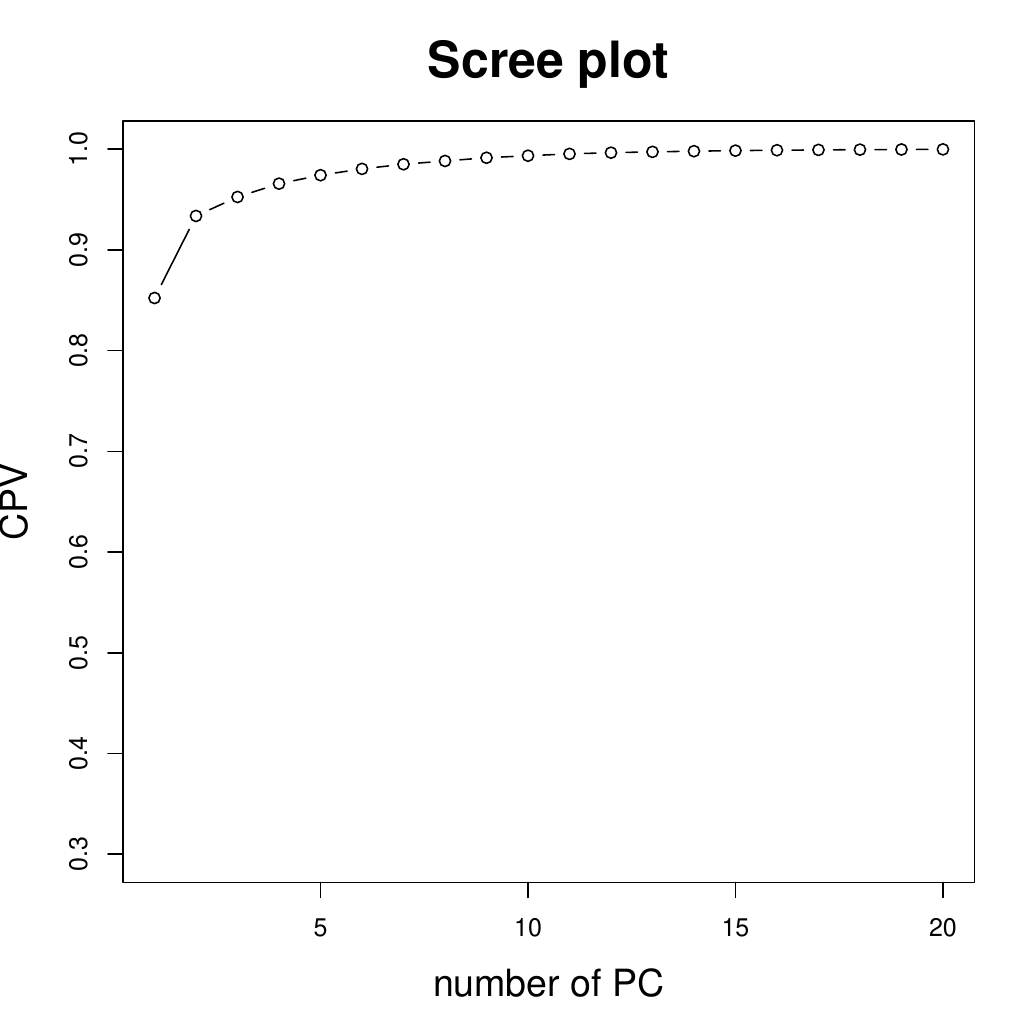}
\caption{Energy data. Plot of the CPV versus the number of PCs of the functional PCA.
\label{fig:scree_plot_energy}}
\end{minipage}
\vskip 5mm
\begin{minipage}{\textwidth}
\includegraphics[width=\textwidth,trim=0 12mm 0 15mm,clip]{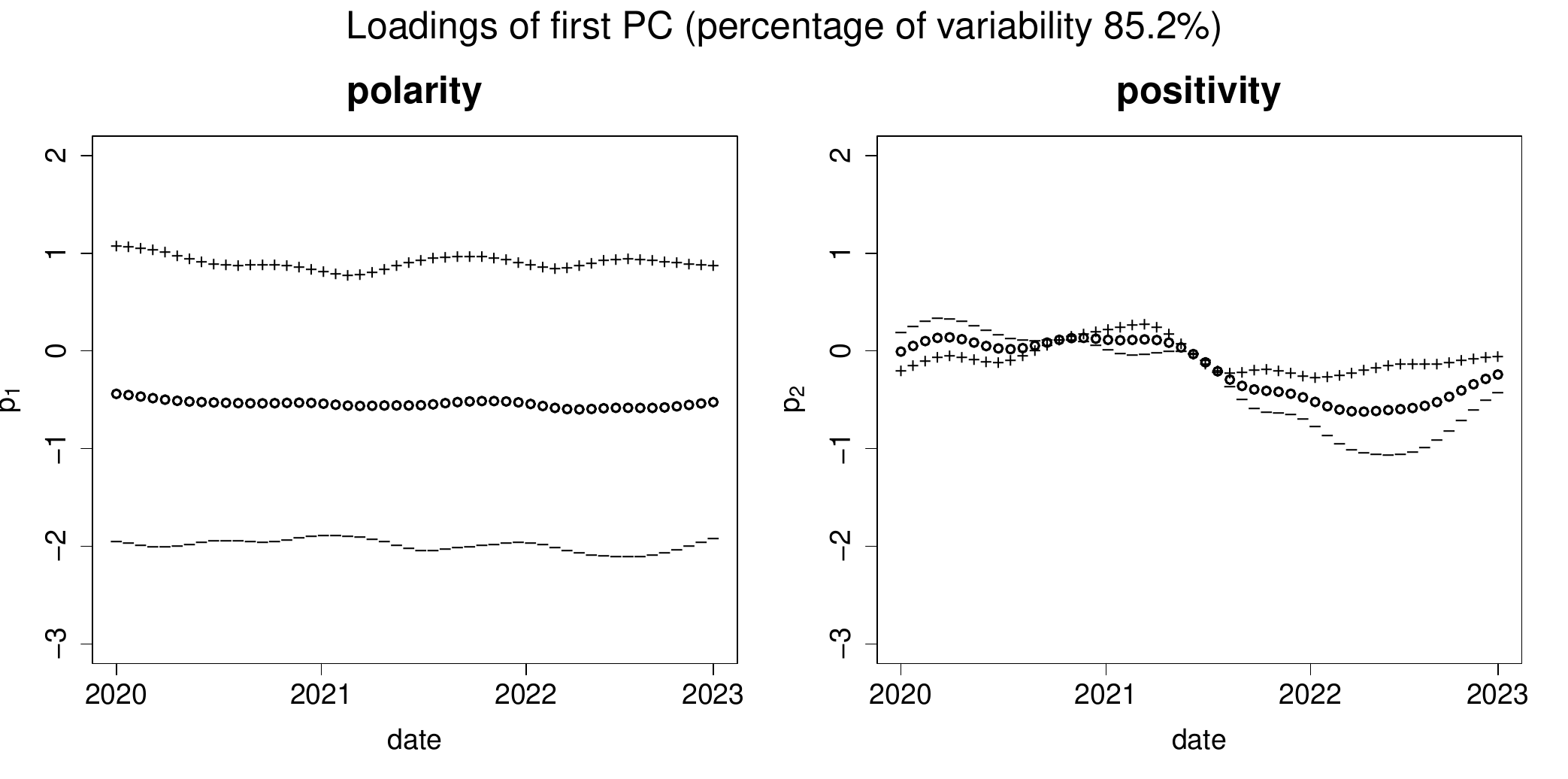}
\caption{Energy data. Loadings of the first principal component of the functional PCA, capturing 85.2\% of the total variability.
\label{fig:loadings_PC1_energy}}
\end{minipage}
\end{figure}

We perform the same dimensionality reduction analysis via functional PCA on the thematic corpus on energy data (Fig~\ref{fig:scree_plot_energy} and Fig~\ref{fig:loadings_PC1_energy}), obtaining comparable values of variance explained, but, in this case, the loadings of the first principal component for the positivity curve feature a temporal evolution that contrasts the beginning of year 2021 with the rest of the time window.
The first principal component accounts for more than 85\% of the variability of the energy data (see the scree-plot in Fig~\ref{fig:scree_plot_energy}). However, unlike what happens for the baseline, the effect of the first principal component on the positivity curves is not just a vertical shift, but features a contrast between different time periods (see the loadings of the first principal component in Fig~\ref{fig:loadings_PC1_energy}); this behavior confirms the findings reported in Section 5 of the main article.

\subsection{PCA: AI case study}

\label{sec:baseline_AI}
Since the AI case study incorporates two years more than the energy case study, we perform again the analysis including the full time range.



\begin{figure}[tb!]
\begin{minipage}{\textwidth}
\centering\includegraphics[width=0.5\textwidth]{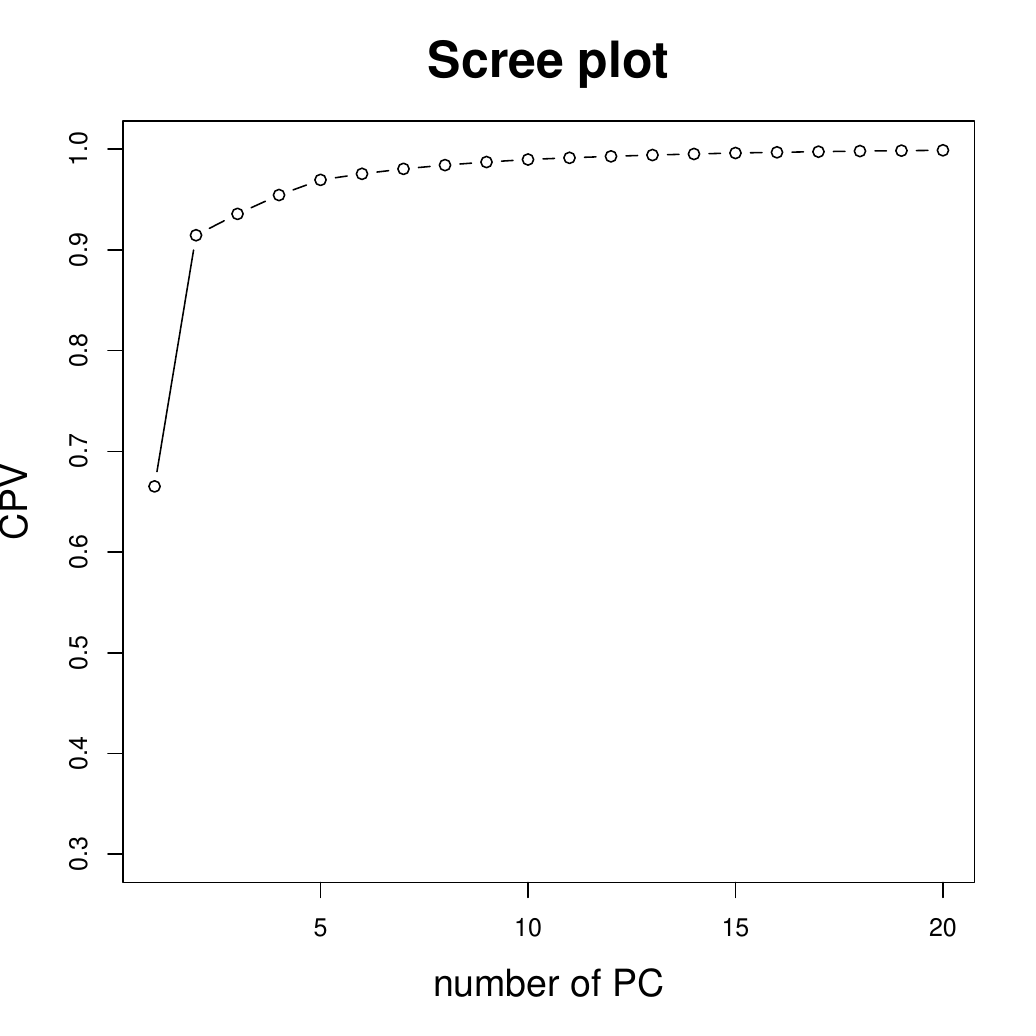}
\caption{XLM-RLnews-5 baseline data. Plot of the CPV versus the number of PCs of the functional PCA on the baseline data.
\label{fig:scree_plot_2}\vspace{3mm}}
\end{minipage}
%
\begin{minipage}{\textwidth}
\centering\includegraphics[width=\textwidth,trim=0 12mm 0 15mm,clip]{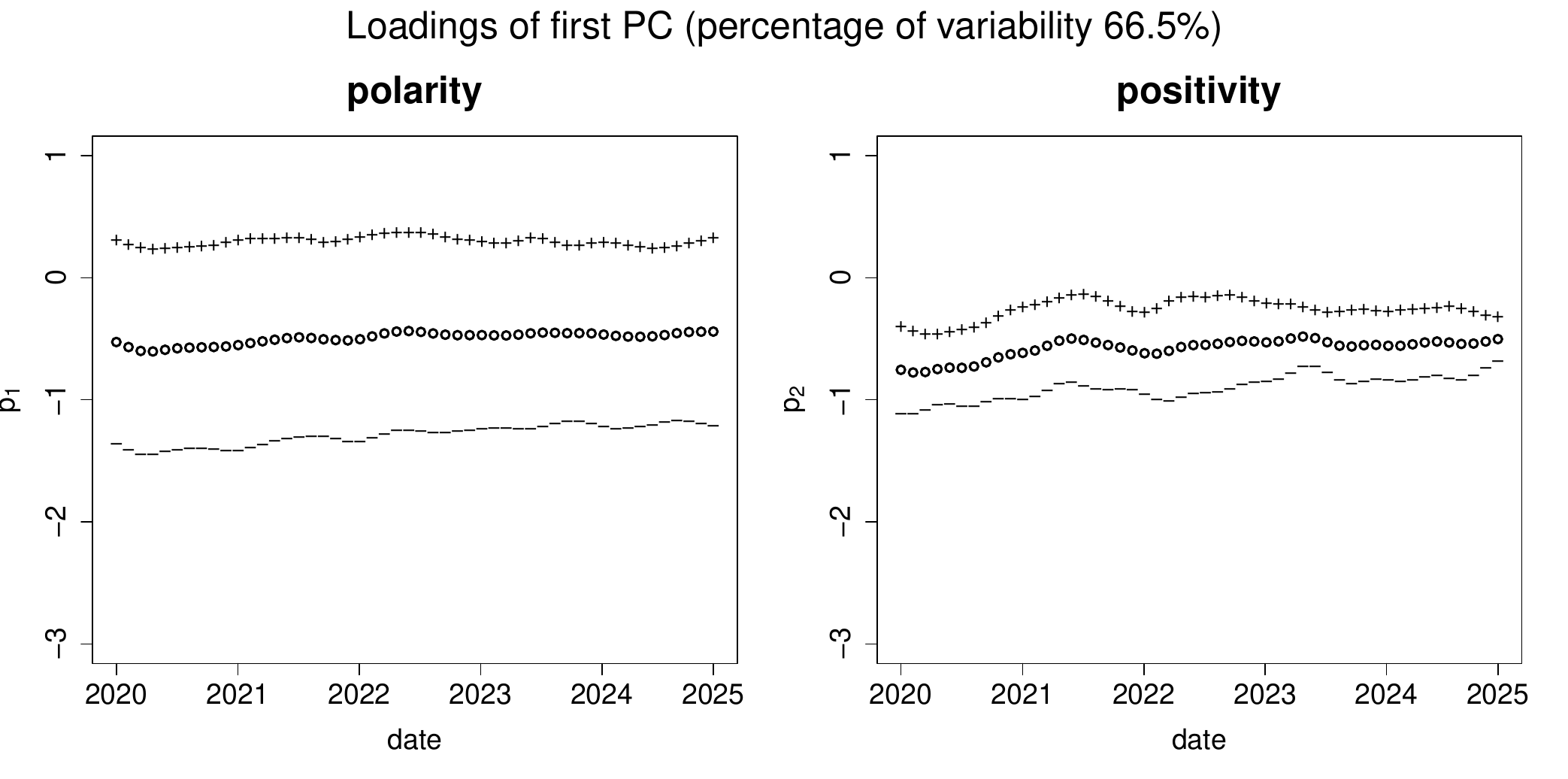}
\caption{XLM-RLnews-5 baseline data. Loadings of the first principal component (explaining 66.5\% of the total variability) of the functional PCA on the baseline data represented as perturbations of the mean function (dotted line) obtained by adding (line of + signs) or subtracting (line of - signs) two standard deviations of the first principal component. Polarity and positivity on the left and right, respectively.
\label{fig:loadings_PC1_2}}
\end{minipage}
\end{figure}

\begin{figure}[tb!]
\begin{center}
\includegraphics[width=0.5\textwidth]{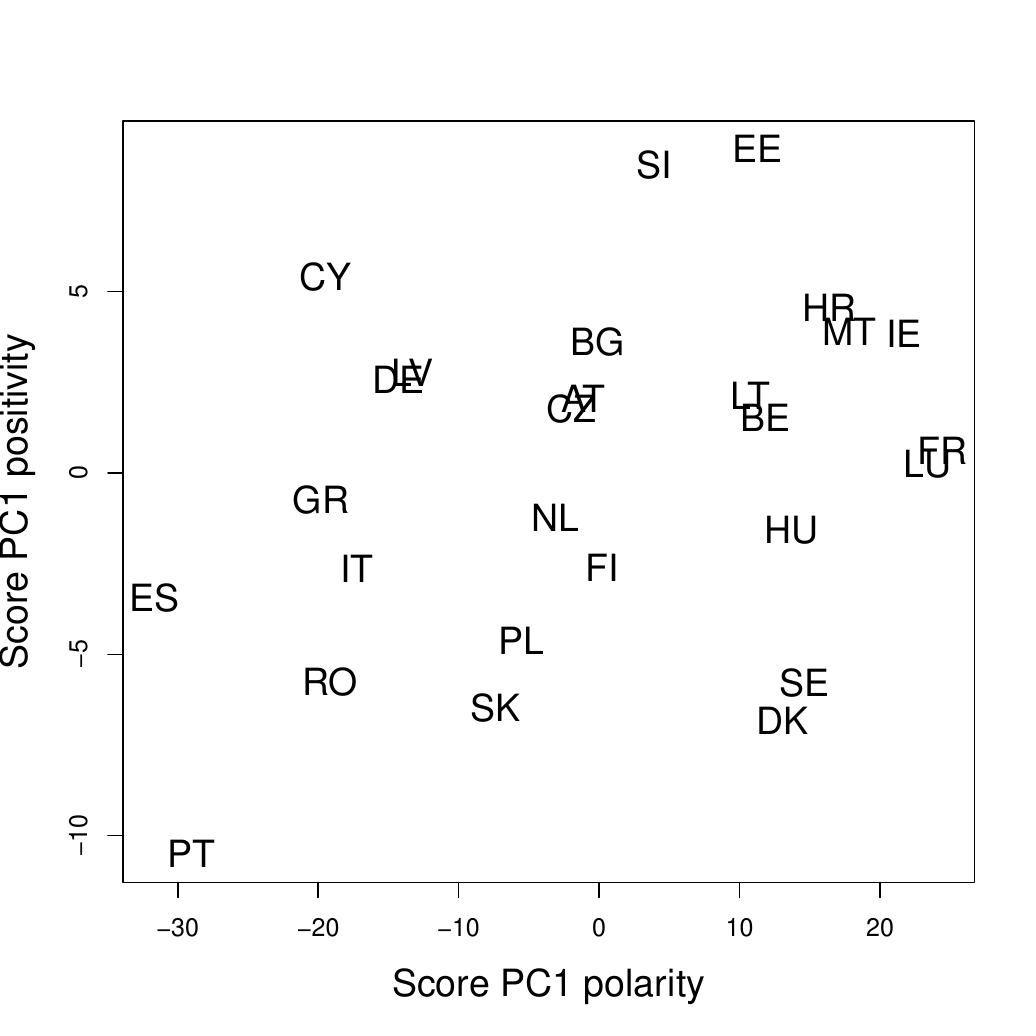}
\end{center}
\caption{XLM-RLnews-5 baseline data. Scatterplot of the scores of the 27 EU countries along the first principal component (polarity dimension on the x-axis and positivity dimension on the y-axis) on the baseline data.
\label{fig:scores_sbp_new}}
\end{figure}

We carry out functional PCA to obtained a reduced but still informative representation of the data.
The functional PCA performed on the baseline data annotated using XLM-RLnews-5 and including two more years (2023--2024) provides conclusions similar to those obtained on the baseline annotated with XLM-RLnews-8 and ranging only 2020--2022 (see Fig~\ref{fig:scree_plot_2} and Fig~\ref{fig:loadings_PC1_2}): the first principal component accounts for the majority of the variability of the data (the percentage of variance explained is more than 65\%) and its loadings curves account for the vertical shift of the curves.
However, the scatterplot of the polarity and positivity scores along the first principal component (Fig~\ref{fig:scores_sbp_new}) is no more affected by the language-related bias, with Germany and Austria lying within the range of values of all the other countries, allowing an unbiased evaluation of the differences among countries. 


In the plot, on the left there are the countries whose media sentiment analysis is characterized by neutrality (such as Spain and Portugal), while on the right those characterized by a more polarized media landscape (such as Luxembourg and France). Countries with a media landscape characterized by positivity (such as Slovenia and Estonia) are on the upper part, and those characterized by negativity (such as Portugal) on the lower part. The countries with a media landscape unbiased towards positivity or negativity are in the central part (such as Greece, the Netherlands, France and Luxembourg). 

\begin{figure}[ht]
\begin{minipage}{\textwidth}
\centering\includegraphics[width=0.5\textwidth]{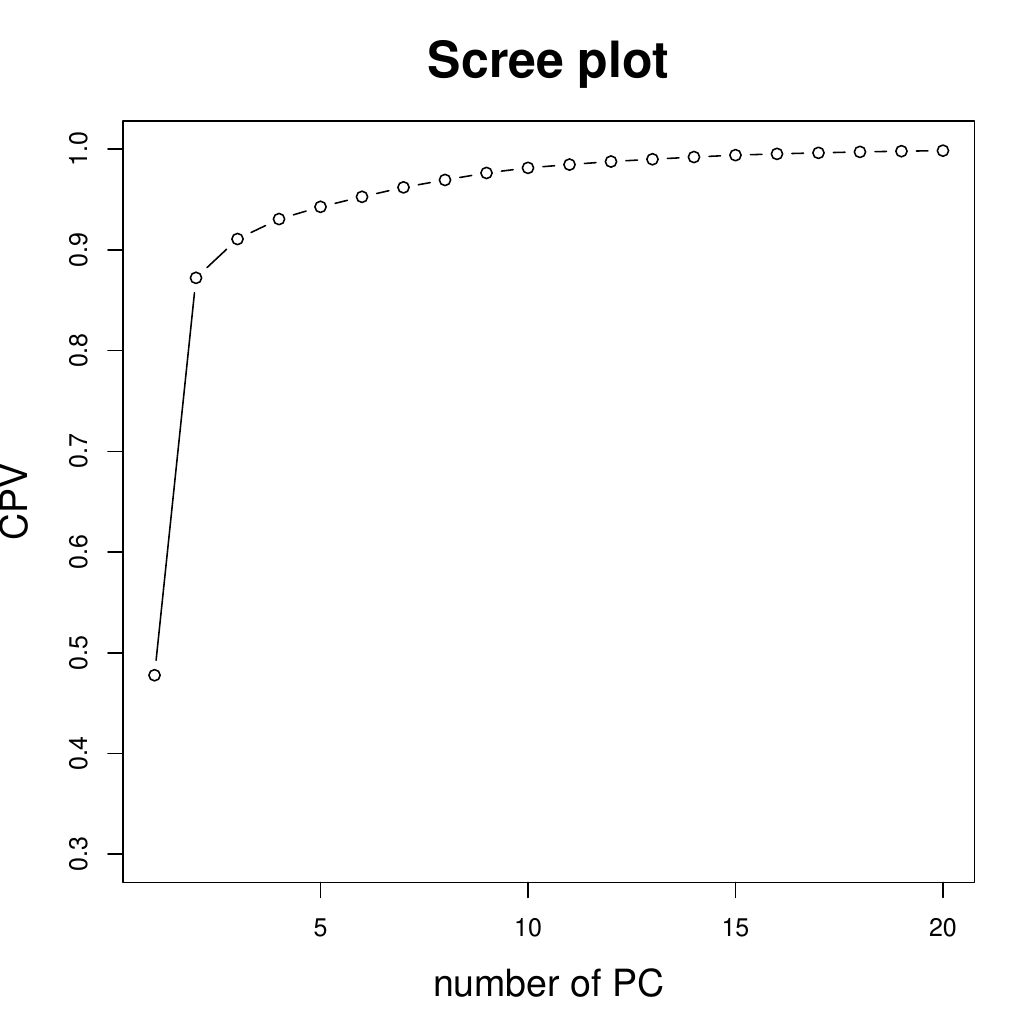}
\caption{AI data. Plot of the CPV versus the number of PCs of the functional PCA.
\label{fig:scree_plot_AI}}
\end{minipage}
%
\vskip 5mm
\begin{minipage}{\textwidth}
\includegraphics[width=\textwidth,trim=0 12mm 0 15mm,clip]{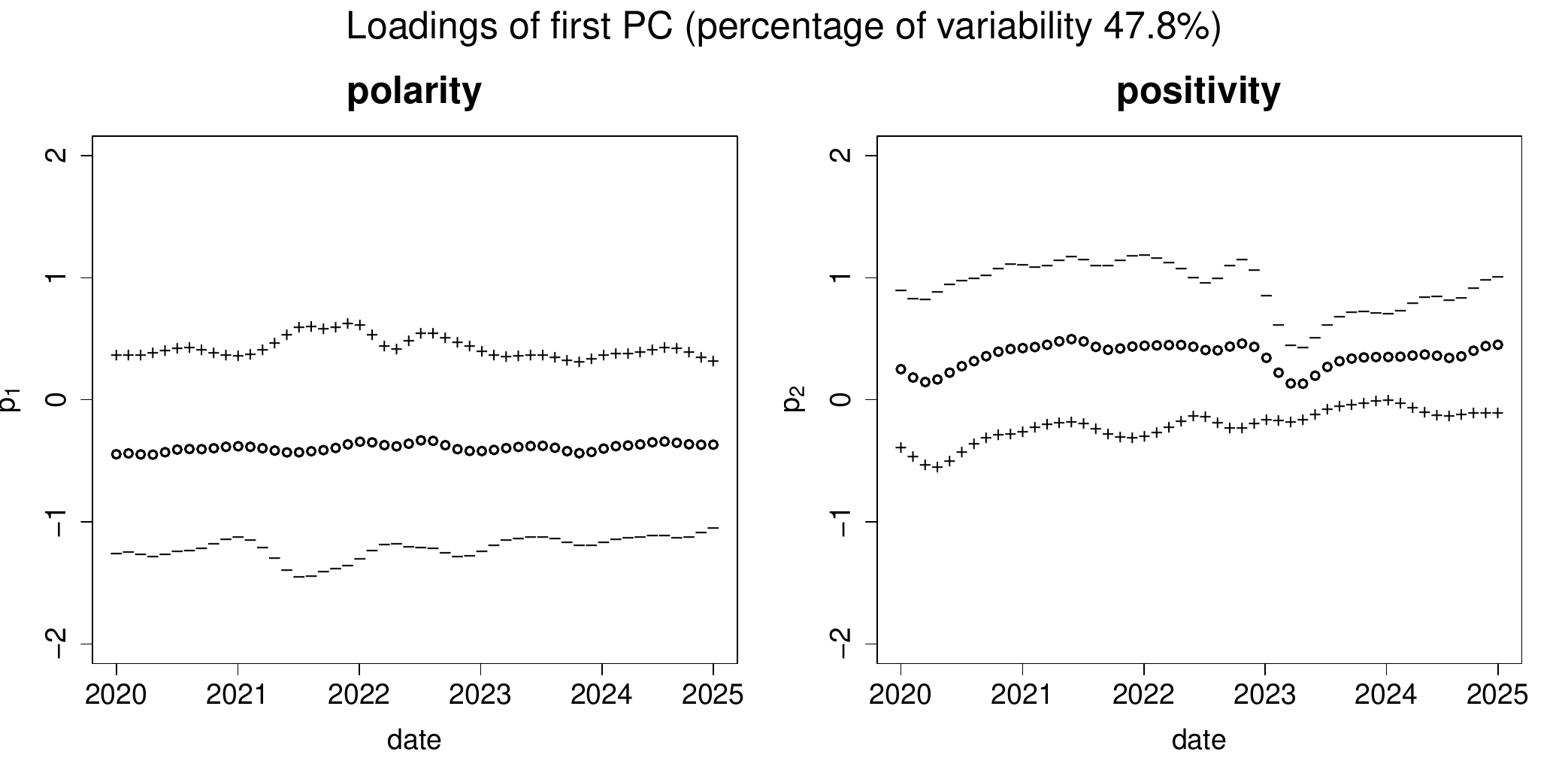}
\caption{AI data. Loadings of the first principal component of the functional PCA, capturing 47.8\% of the total variability. 
\label{fig:loadings_PC1_AI}}
\end{minipage}
\end{figure}

We conduct the same dimensionality reduction analysis on the AI dataset (Fig~\ref{fig:scree_plot_AI} and Fig~\ref{fig:loadings_PC1_AI}),  obtaining a first principal component, accounting also in this case mainly for a vertical shift of the curves, that explains less than 50\% of the variability. This suggests a more complex situation than the baseline, which is quantitatively analyzed via \textit{sentiment sensing} in the main article.

\section{Additional results comparing the energy curves' depth with respect to the baseline}
\label{sec:appendix_intra_1_energy}

\begin{figure}[ht]
\begin{minipage}[c]{0.03\textwidth}
\centering
\textbf{ } 
\end{minipage}
\hspace{2mm} \hfill
\begin{minipage}[c]{0.45\textwidth}
\centering\textbf{polarity} 
\end{minipage}
\hfill
\begin{minipage}[c]{0.45\textwidth}
\centering\textbf{positivity} 
\end{minipage}
\\[1mm]
\begin{minipage}[c]{0.03\textwidth}
\centering
\rotatebox{90}{%
\textbf{Belgium} 
}
\end{minipage}
\hspace{3mm} 
\begin{minipage}[c]{0.9\textwidth}
\centering
\includegraphics[width=\textwidth,trim=0 12mm 0 28mm,clip]{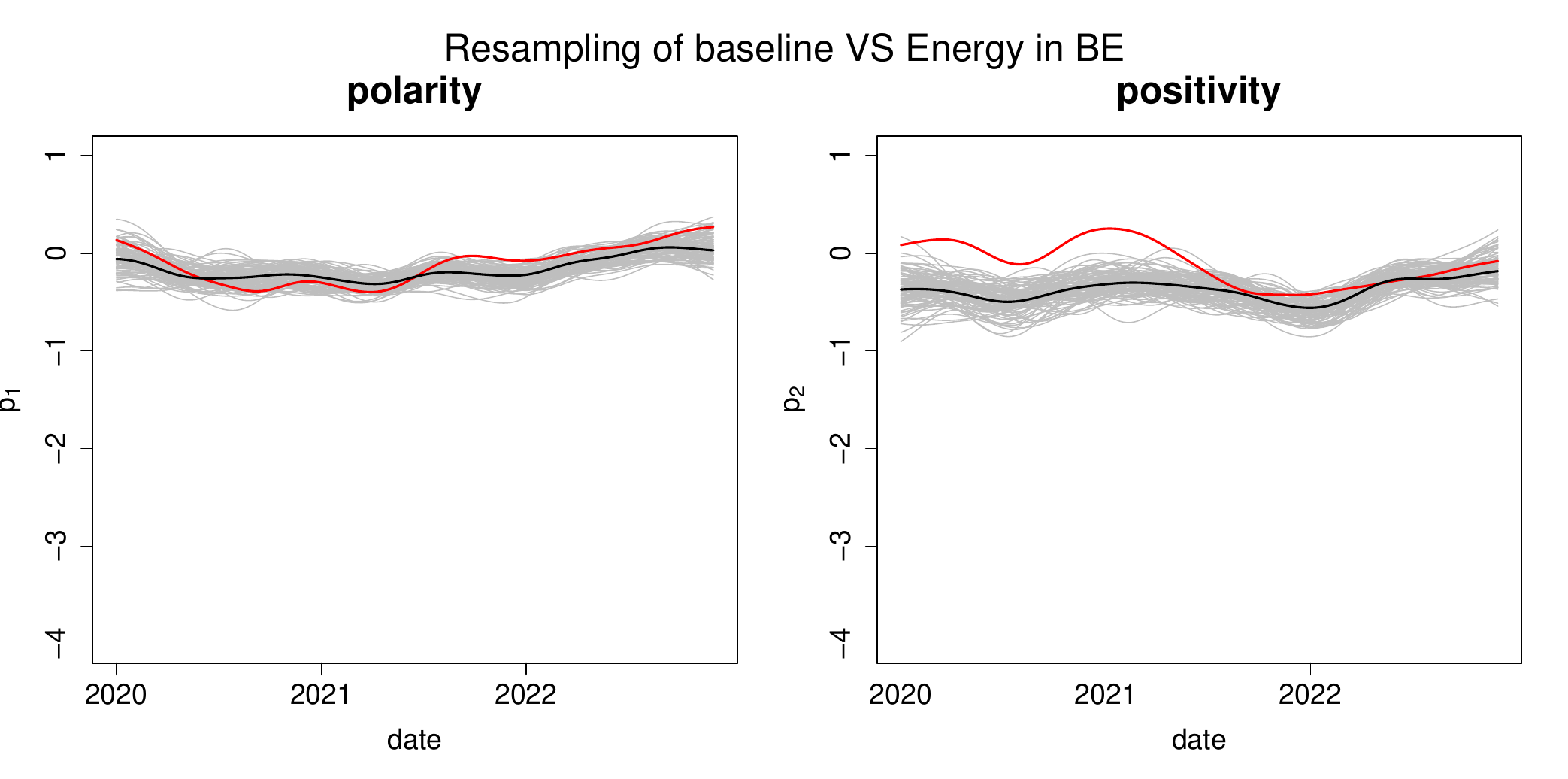}
\end{minipage}
\caption{Energy data. Curves obtained from the re-sampling of the baseline data (gray) and curves obtained from the energy data (red) for Belgium. The curves for the original baseline data are in black. Polarity on the left panel and positivity on the right panel. 
\label{fig:BE}}
\end{figure}

Fig~\ref{fig:BE} shows, as an example, the curves obtained for Belgium: the 100 gray curves are obtained from the re-sampling of the baseline data (whose curves are reported in black), while the red curves are obtained from the energy data. We can see that polarity of the energy data mostly falls within the distribution of the baseline, while positivity of the energy data falls above the distribution of the baseline until mid-2021 and within it after mid-2021. This means that, for the energy topic, the proportion of polarized news articles with respect to the neutral ones is similar to the one of the overall media landscape, but it was spoken of in a more positive way with respect to the overall media landscape until mid-2021, when a drop in positivity occurred.

\begin{figure}[ht]
\begin{minipage}[c]{0.04\textwidth}
\centering
\textbf{ } 
\end{minipage}
\hspace{2mm} 
\begin{minipage}[c]{0.45\textwidth}
\centering\textbf{polarity} 
\end{minipage}
\hfill
\begin{minipage}[c]{0.45\textwidth}
\centering\textbf{positivity} 
\end{minipage}
\\[1mm]
\begin{minipage}[c]{0.04\textwidth}
\centering
\rotatebox{90}{%
\textbf{Hungary} 
}
\end{minipage}
\hspace{3mm} 
\begin{minipage}[c]{0.9\textwidth}
\centering
\includegraphics[width=\textwidth,trim=0 12mm 0 28mm,clip]{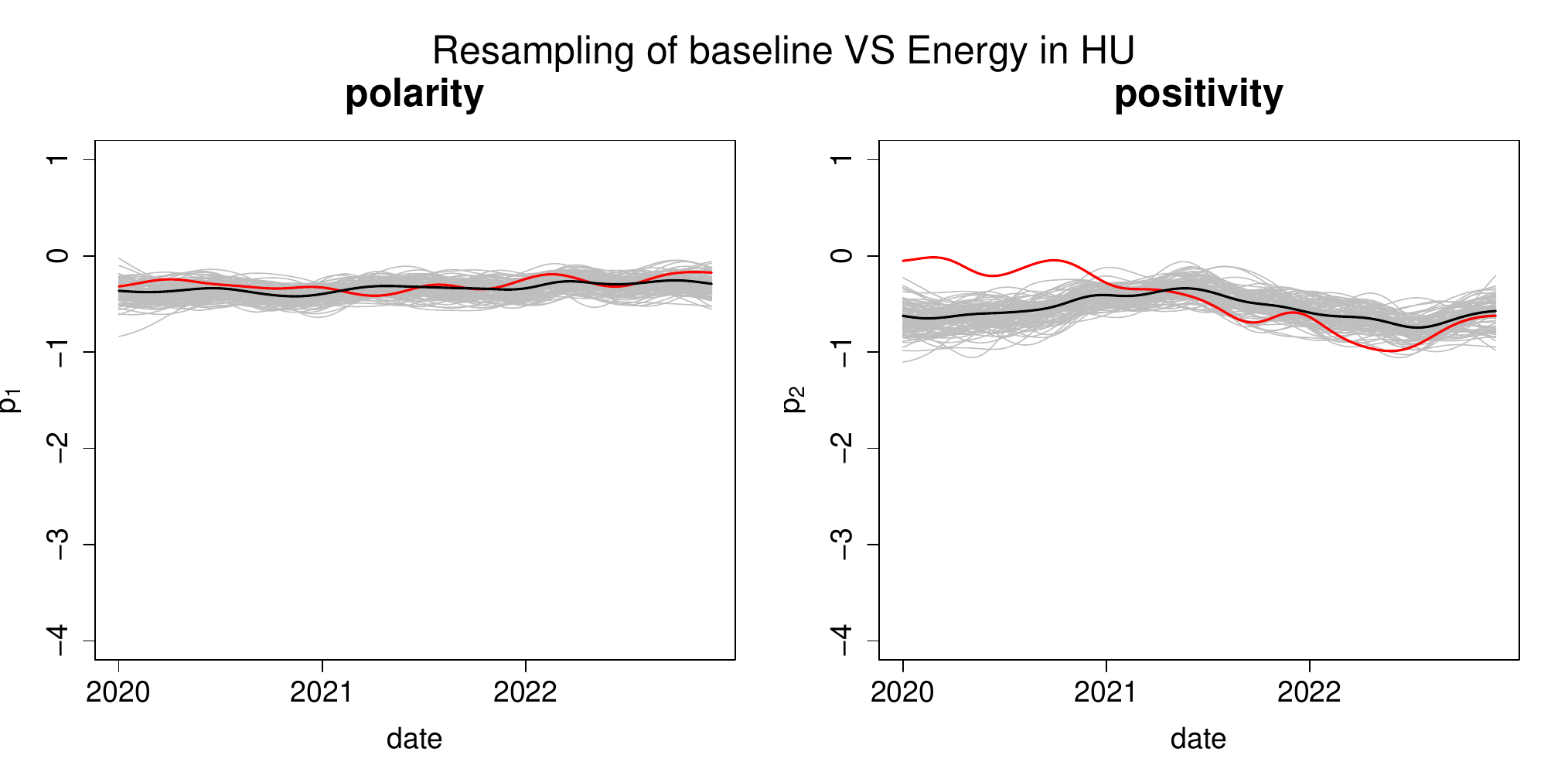}
\end{minipage}
\\[1mm]
\begin{minipage}[c]{0.04\textwidth}
\centering
\rotatebox{90}{%
\textbf{Ireland} 
}
\end{minipage}
\hspace{3mm} 
\begin{minipage}[c]{0.9\textwidth}
\includegraphics[width=\textwidth,trim=0 12mm 0 28mm,clip]{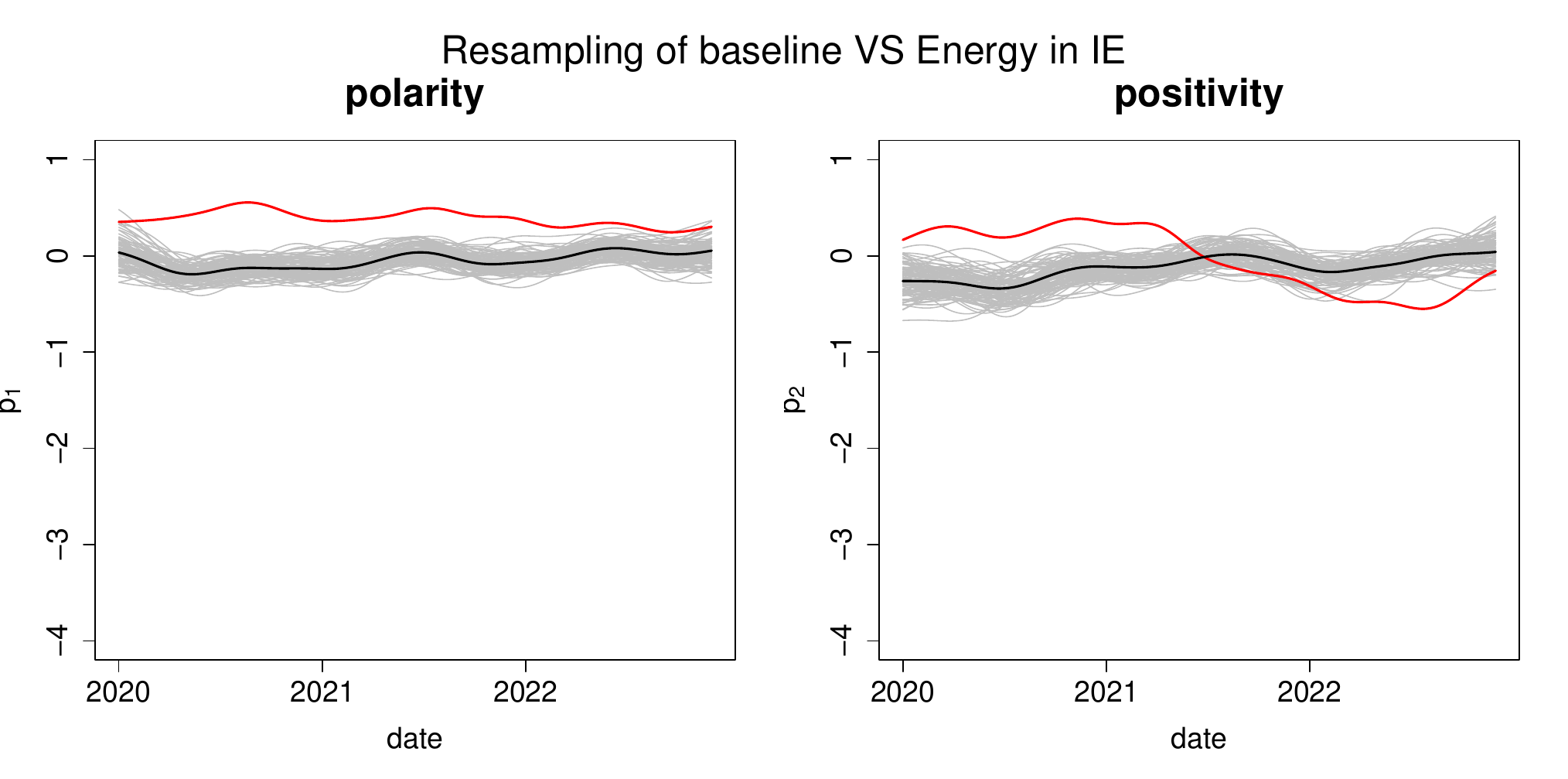}
\end{minipage}
\\[1mm]
\begin{minipage}[c]{0.04\textwidth}
\centering
\rotatebox{90}{%
\textbf{Latvia} 
}
\end{minipage}
\hspace{3mm} 
\begin{minipage}[c]{0.9\textwidth}
\includegraphics[width=\textwidth,trim=0 12mm 0 28mm,clip]{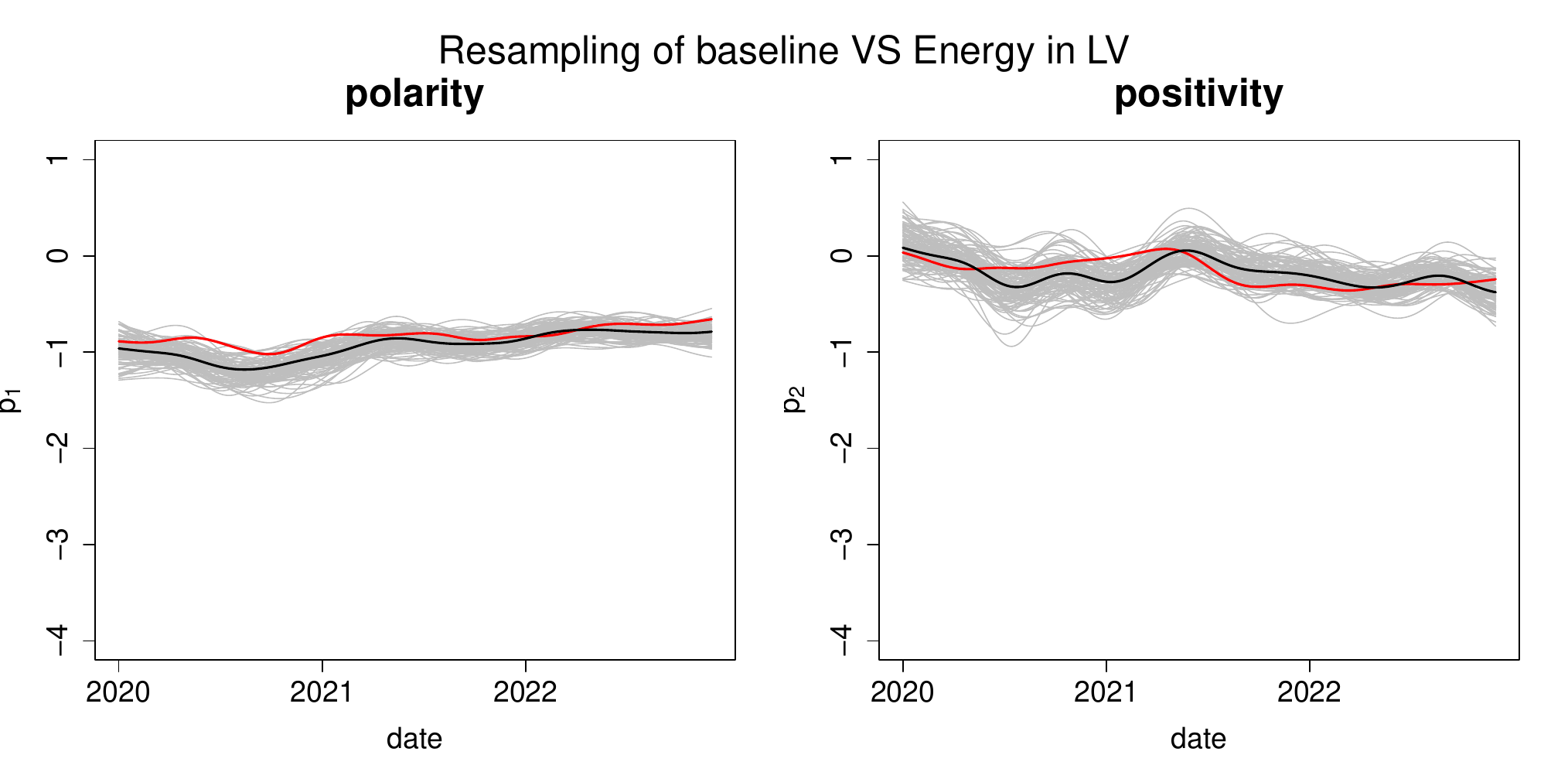}
\end{minipage}
\\[1mm]
\begin{minipage}[c]{0.04\textwidth}
\centering
\rotatebox{90}{%
\textbf{Czech Republic} 
}
\end{minipage}
\hspace{3mm} 
\begin{minipage}[c]{0.9\textwidth}
\includegraphics[width=\textwidth,trim=0 12mm 0 28mm,clip]{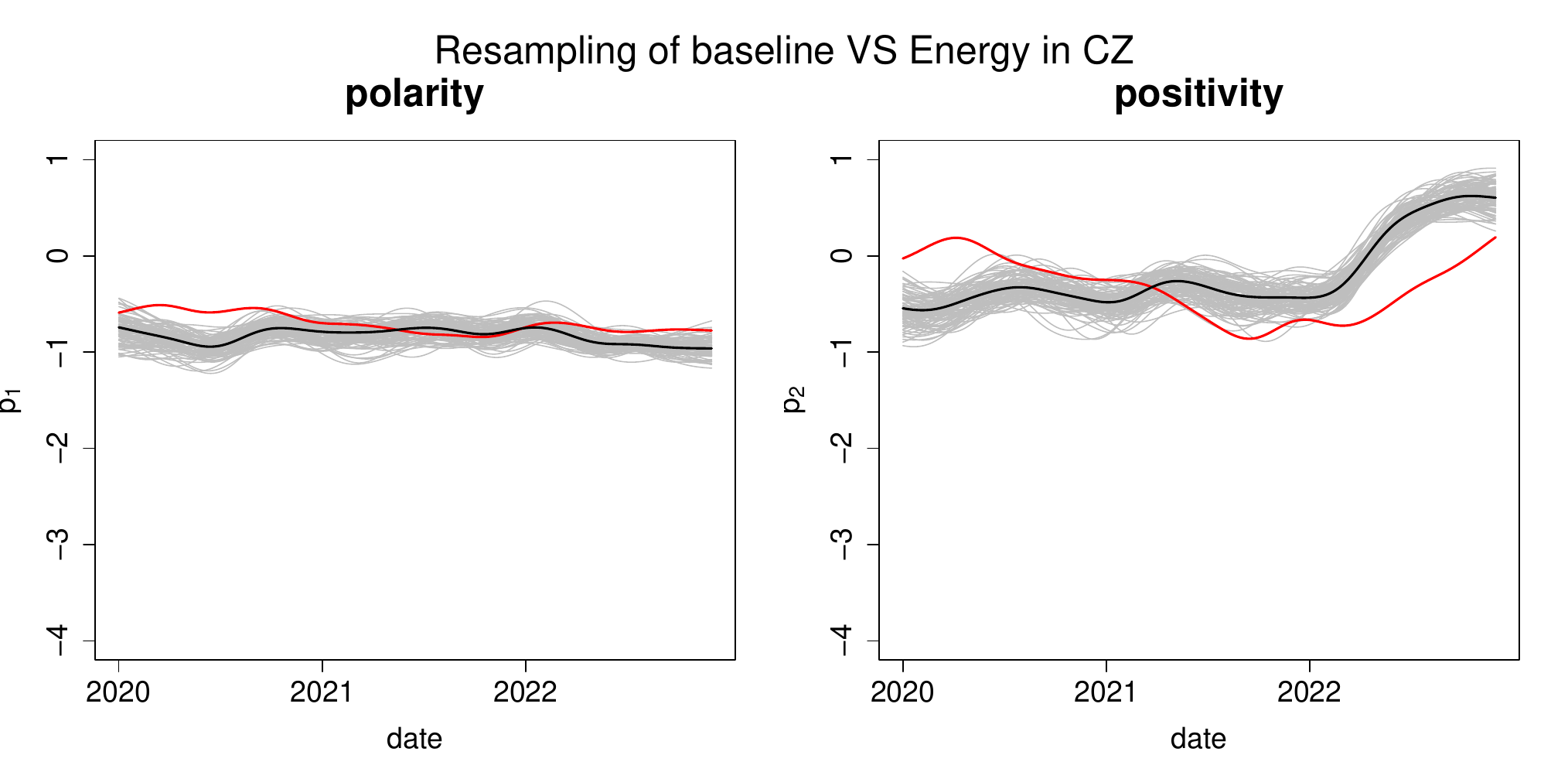}
\end{minipage}
\caption{Energy data. Curves obtained from the re-sampling of the baseline data (gray) and curves obtained from the energy data (red) for Hungary, Ireland, Latvia and Czech Republic. Polarity on the left panel and positivity on the right panel.
\label{fig:HU_IE_LV_CZ}}
\end{figure}

\begin{itemize}
\item
Those in the left part (i.e. countries with a low standardized distance for polarity) are characterized by a polarity curve for the energy data which falls in the middle of the set of baseline curves. 
See, for example the polarity curves at
Fig~\ref{fig:HU_IE_LV_CZ} for Hungary, which has a negative standardized distance for polarity (-0.04).
\item
Those in the right part (i.e. countries with a high standardized distance for polarity) are characterized by a polarity curve for the energy data which falls outside or at the boundary of the set of baseline curves. See, for example the polarity curves at 
Fig~\ref{fig:HU_IE_LV_CZ} for Ireland, which has a high standardized distance for polarity (8.22).
\item
Those in the bottom part (i.e. countries with standardized distance for positivity) are characterized by a positivity curve for the energy data which falls in the middle of the set of baseline curves. See, for example the positivity curves at 
Fig~\ref{fig:HU_IE_LV_CZ}  for Latvia, which has a small standardized distance for positivity (0.23).
\item
Those in the top part (i.e. countries with a high standardized distance for positivity) are characterized by a positivity curve for the energy data which falls outside or at the boundary of the set of baseline curves (in this case, above it in 2020 and below it after 2020). See, for example the positivity curves at 
Fig~\ref{fig:HU_IE_LV_CZ}  for Czech Republic, which has a high standardized distance for positivity (6.61).
\end{itemize}

\section{Additional results on the statistical analysis of temporal trends in energy data}
\label{sec:appendix_intra_2_energy}

\begin{figure}[ht]
\begin{minipage}[c]{0.04\textwidth}
\centering
\textbf{ } 
\end{minipage}
\hspace{2mm} 
\begin{minipage}[c]{0.45\textwidth}
\centering\textbf{First derivative} 
\end{minipage}
\begin{minipage}[c]{0.45\textwidth}
\centering\textbf{Original curves} 
\end{minipage}
\\[1mm]
\begin{minipage}[c]{0.04\textwidth}
\centering
\rotatebox{90}{%
\textbf{Germany} 
}
\end{minipage}
\hspace{3mm} 
\begin{minipage}[c]{0.9\textwidth}
\centering
\includegraphics[width=\textwidth,trim=0 12mm 0 28mm,clip]{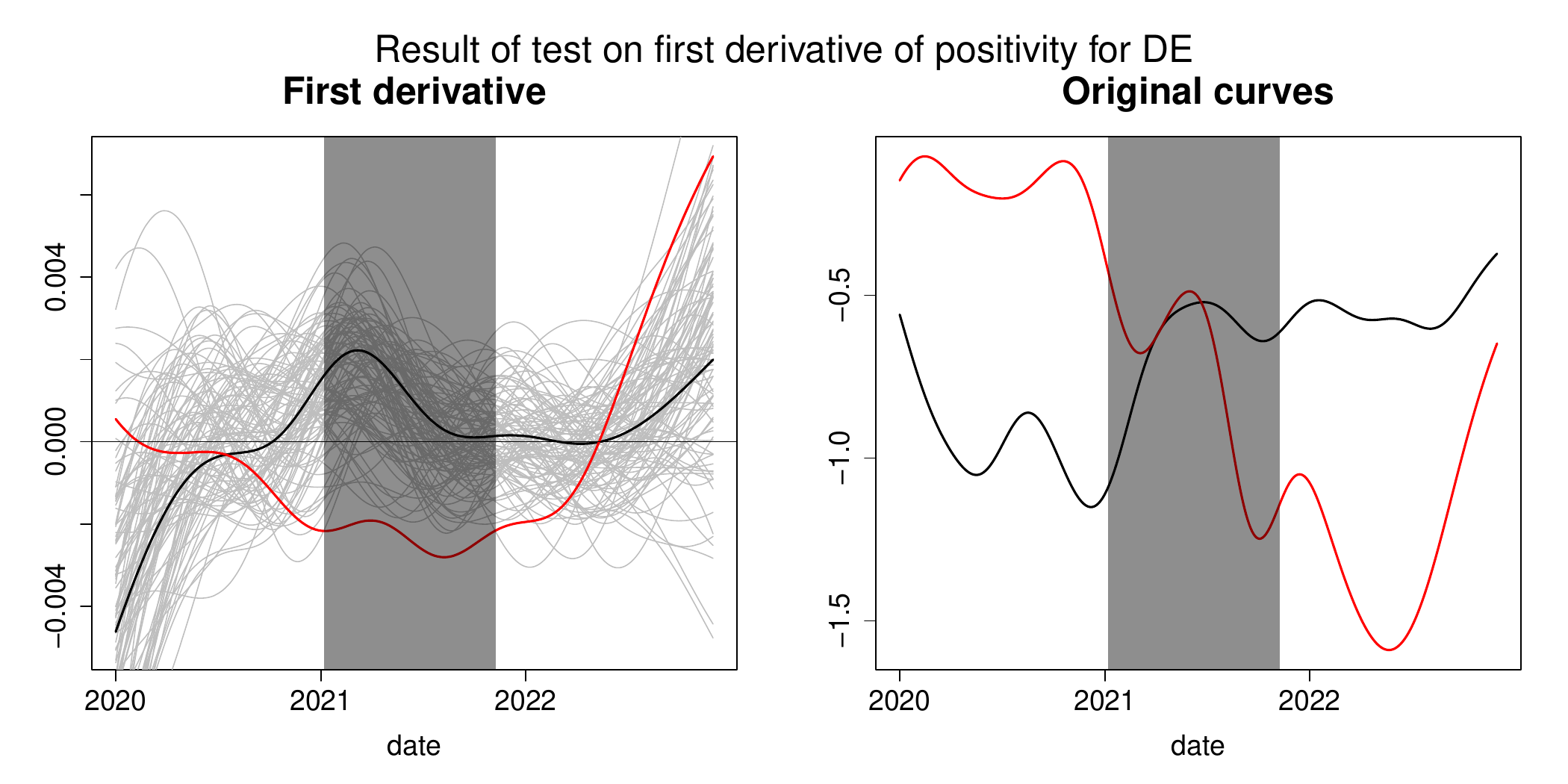}
\end{minipage}
\\[1mm]
\begin{minipage}[c]{0.04\textwidth}
\centering
\rotatebox{90}{%
\textbf{Italy} 
}
\end{minipage}
\hspace{3mm} 
\begin{minipage}[c]{0.9\textwidth}
\centering
\includegraphics[width=\textwidth,trim=0 12mm 0 28mm,clip]{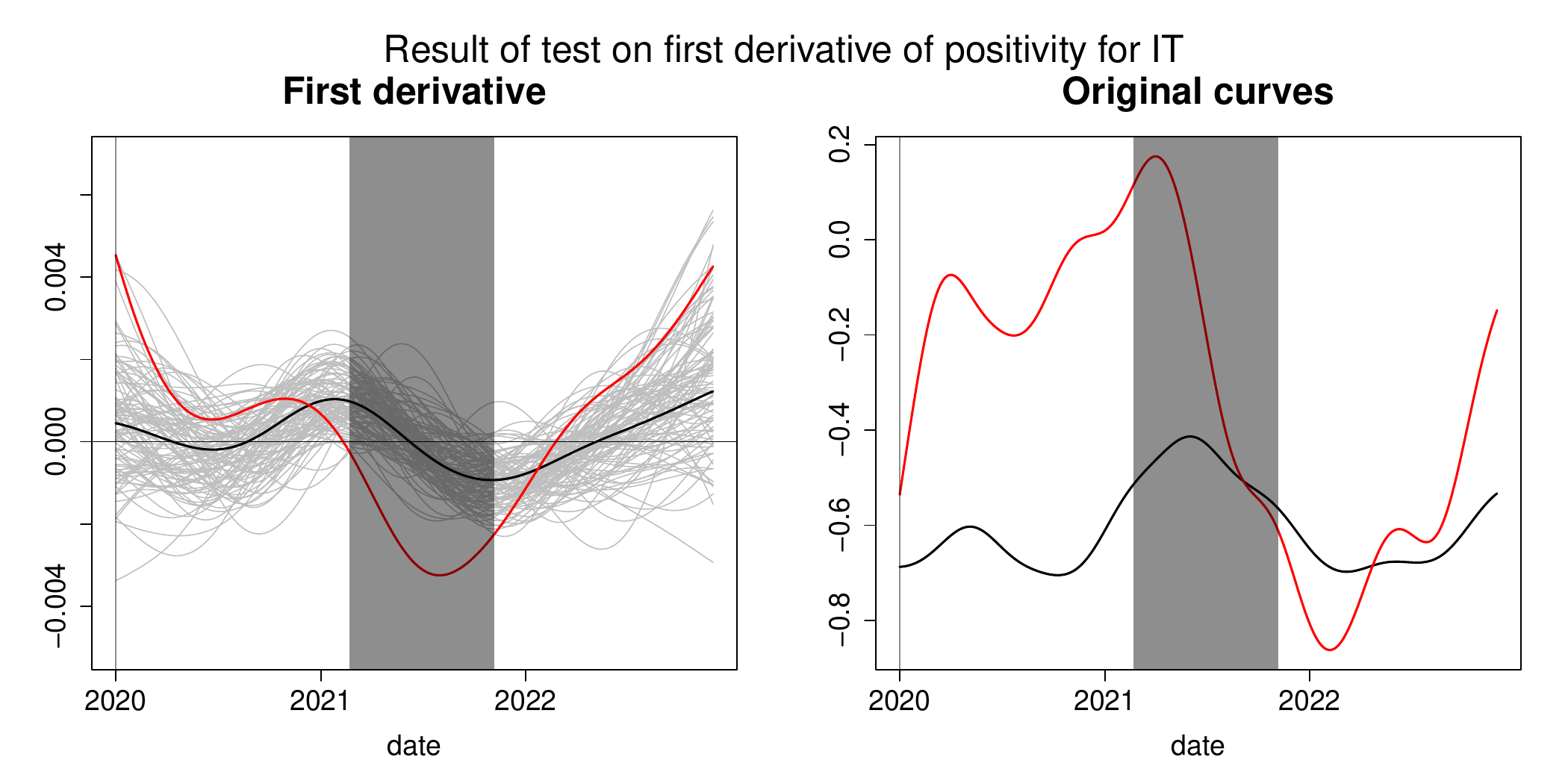}
\end{minipage}
\\[1mm]
\begin{minipage}[c]{0.04\textwidth}
\centering
\rotatebox{90}{%
\textbf{Greece} 
}
\end{minipage}
\hspace{3mm} 
\begin{minipage}[c]{0.90\textwidth}
\centering
\includegraphics[width=\textwidth,trim=0 12mm 0 28mm,clip]{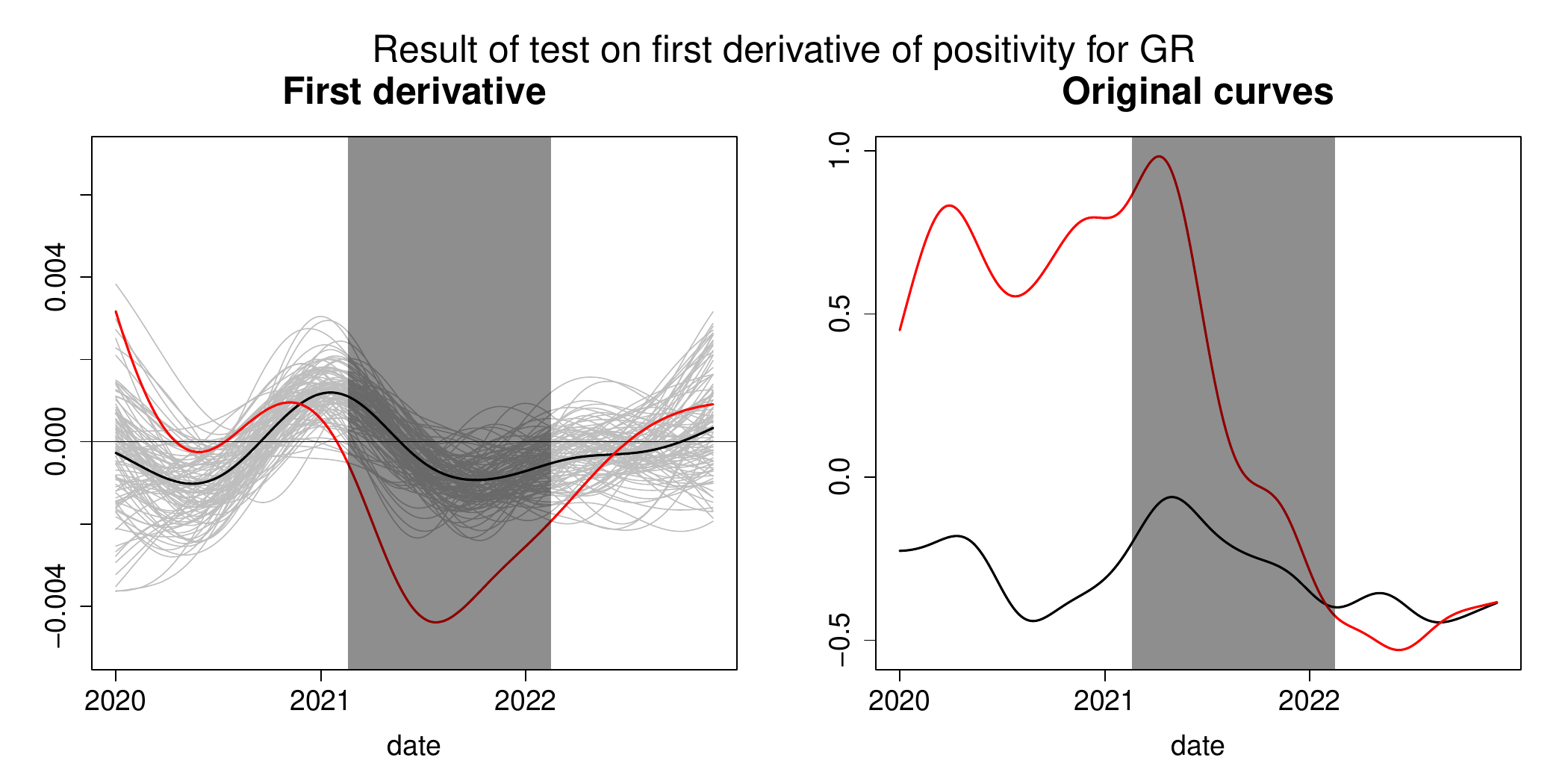}
\end{minipage}
\\[1mm]
\begin{minipage}[c]{0.04\textwidth}
\centering
\rotatebox{90}{%
\textbf{Spain} 
}
\end{minipage}
\hspace{3mm} 
\begin{minipage}[c]{0.90\textwidth}
\centering
\includegraphics[width=\textwidth,trim=0 12mm 0 28mm,clip]{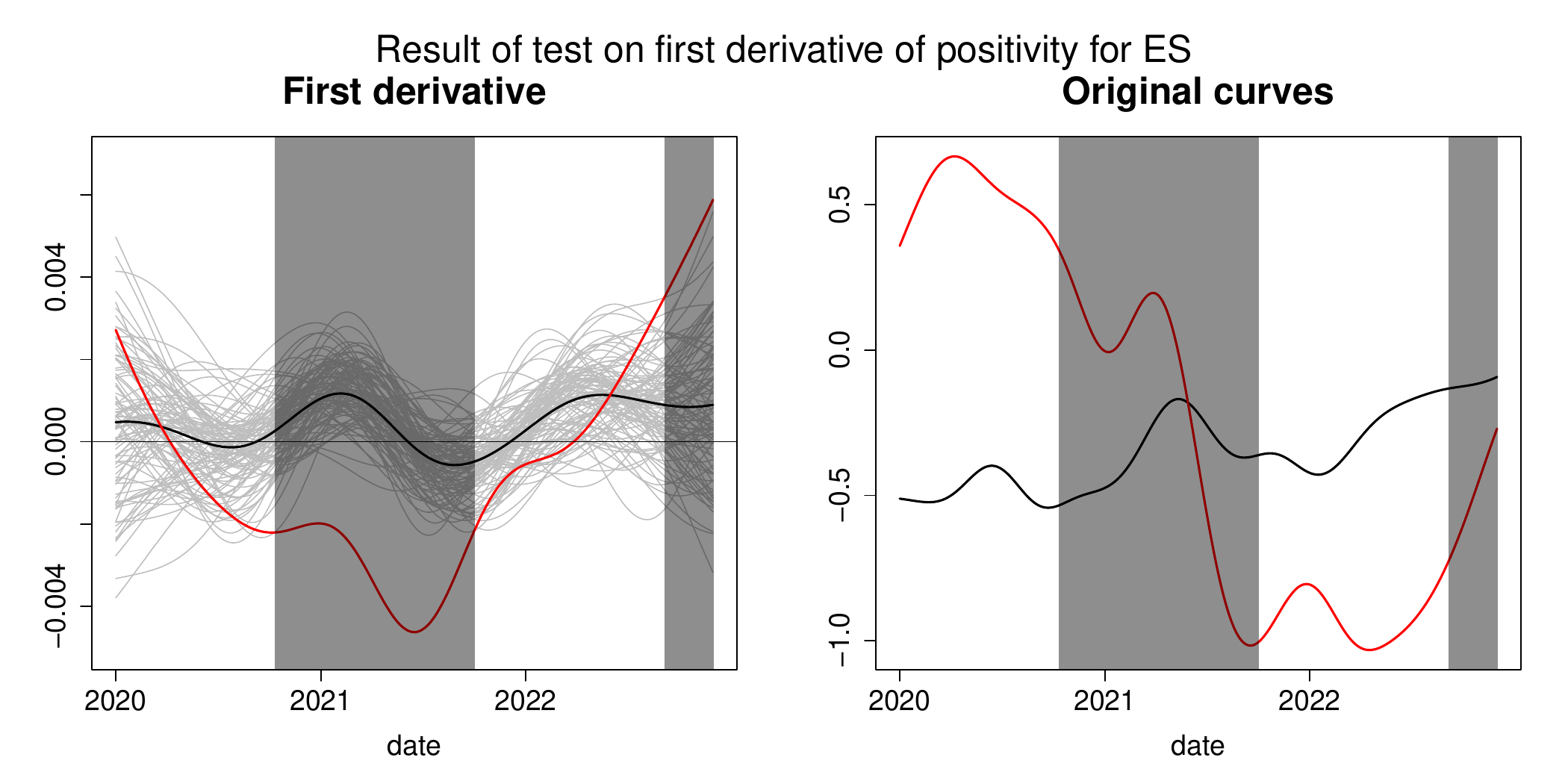}
\end{minipage}
\caption{Energy data. Results of the bootstrap tests on the first derivatives of the positivity curves for Germany, Italy, Greece and Spain. The red curves correspond to energy data, the black curves corresponds to baseline data and the gray curves correspond to resampled baseline data.
\label{fig:DE_IT_GR_ES_energy_derivative}}
\end{figure}

The results for some of the countries (namely, Germany, Italy, Greece and Spain) are reported in 
Fig~\ref{fig:DE_IT_GR_ES_energy_derivative}.
The time points where a significant difference is found are highlighted by gray vertical bands. For each country, the left panel shows the first derivatives (on which the test is performed) and the right panel shows the original curves (to better interpret the results). For instance, for Germany, we see a significant difference in 2021, when the positivity of the overall media sentiment tends to increase (in the left panel we see positive values of the re-sampled baseline and in the right panel we see an upward level shift of the baseline curve) while the positivity of the energy topic has a drastic decrease (in the left panel we see negative values of the energy curve and in the right panel we see a downward level shift of the energy curve). Similar observations can be made for Italy and Greece. For Spain, a second short period with significant differences is found at the end of 2022; however, this may be an effect of the proximity of the end of the time window, as smoothing techniques are affected by possible artifacts at the boundaries of the domain, where the estimation of the first derivative may be less accurate (as can be indeed observed in the left panel, where the variability of the curves is much higher at the boundaries of the domain with respect to the central part).

\section{Additional results comparing the AI curves' depth with respect to the baseline}
\label{sec:appendix_intra_1_AI}

\begin{figure}[ht]
\begin{center}
\includegraphics[width=\textwidth]{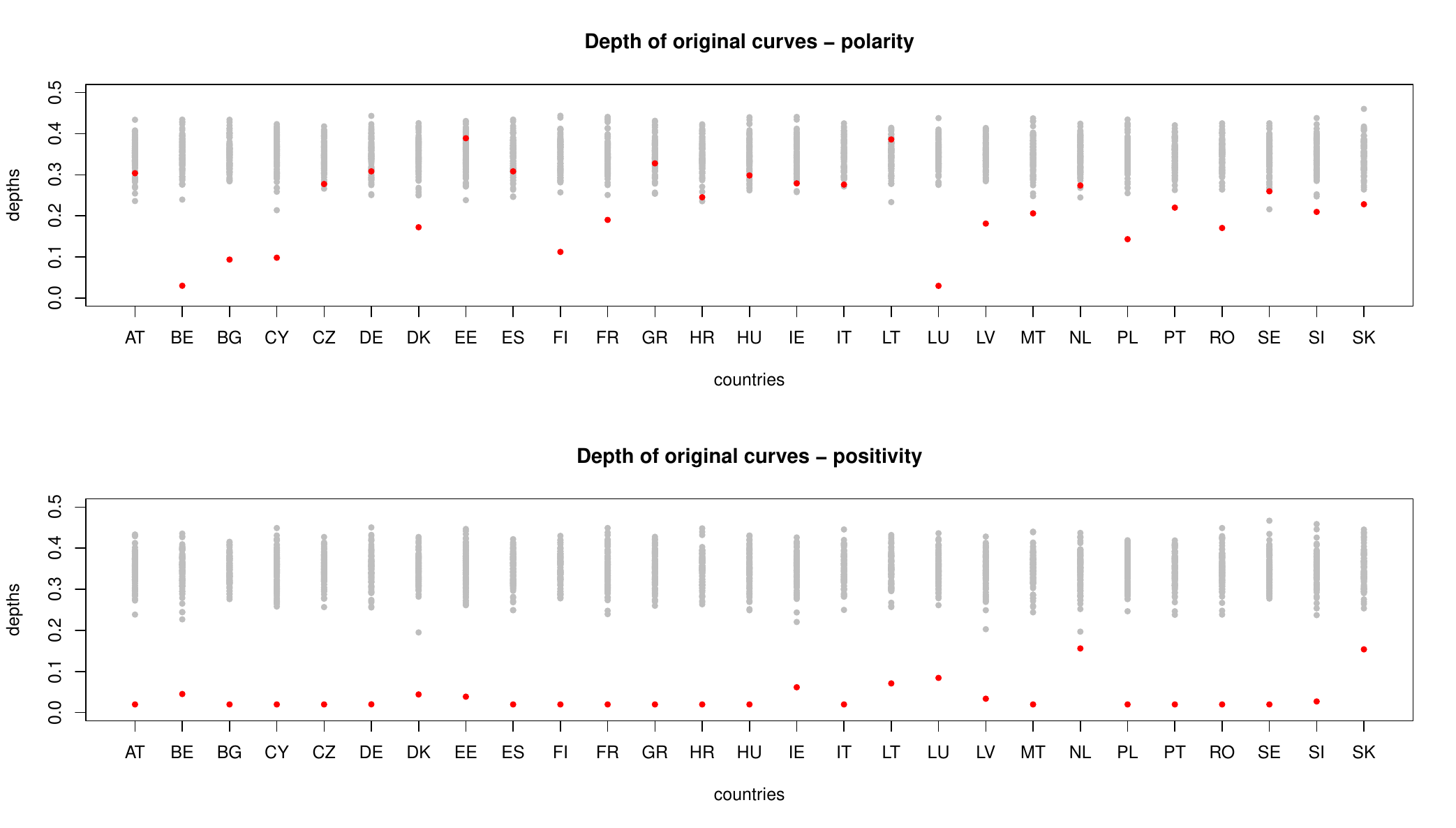}
\end{center}
\caption{AI data. Depths of the 100 curves obtained from the re-sampling of the baseline data (gray) and of the AI curve (red) for the 27 EU countries (x axis) on polarity (top panel) and positivity (bottom panel) for the AI data. 
\label{fig:AI_depth_sbp}}
\end{figure}

Fig~\ref{fig:AI_depth_sbp} shows the depths of the 100 curves obtained from the re-sampling of the baseline data (gray dots) and those of the AI curves (red dots). As expected, the positivity curves for AI have zero depth (meaning that they are outside the set of resampled baseline curves) for almost all countries.

\begin{figure}[ht]
\begin{minipage}[c]{0.04\textwidth}
\centering
\textbf{ } 
\end{minipage}
\hspace{3mm} \hfill
\begin{minipage}[c]{0.45\textwidth}
\centering\textbf{polarity} 
\end{minipage}
\hfill
\begin{minipage}[c]{0.45\textwidth}
\centering\textbf{positivity} 
\end{minipage}
\\[3mm]
\begin{minipage}[c]{0.04\textwidth}
\centering
\rotatebox{90}{%
\textbf{Netherlands} 
}
\end{minipage}
\hspace{3mm} 
\hfill
\begin{minipage}[c]{0.95\textwidth}
\includegraphics[width=\textwidth,trim=0 12mm 0 28mm,clip]{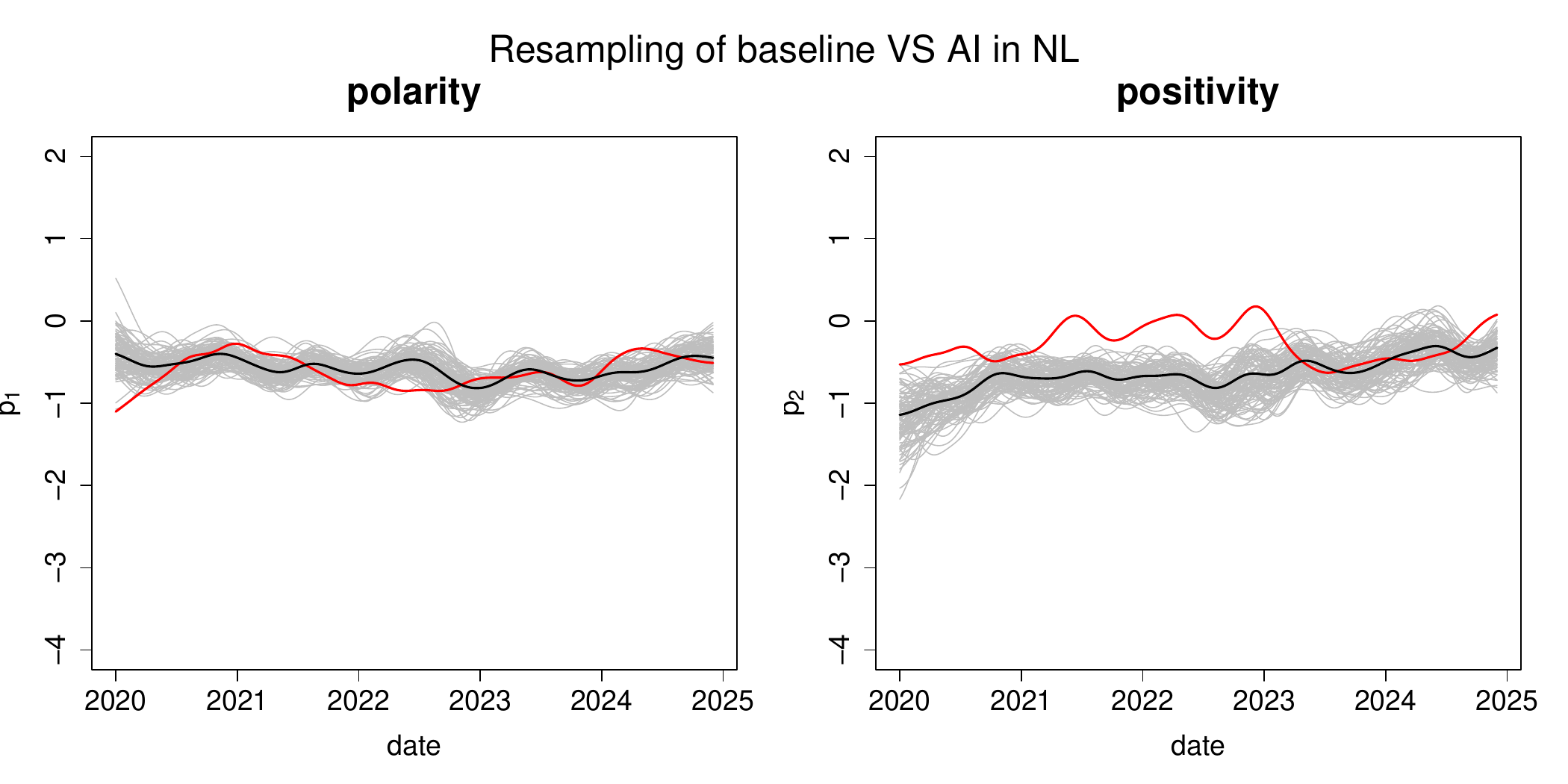}
\end{minipage}
\\[1mm]
\begin{minipage}[c]{0.04\textwidth}
\centering
\rotatebox{90}{%
\textbf{Luxembourg} 
}
\end{minipage}
\hspace{3mm} 
\hfill
\begin{minipage}[c]{0.95\textwidth}
\includegraphics[width=\textwidth,trim=0 12mm 0 28mm,clip]{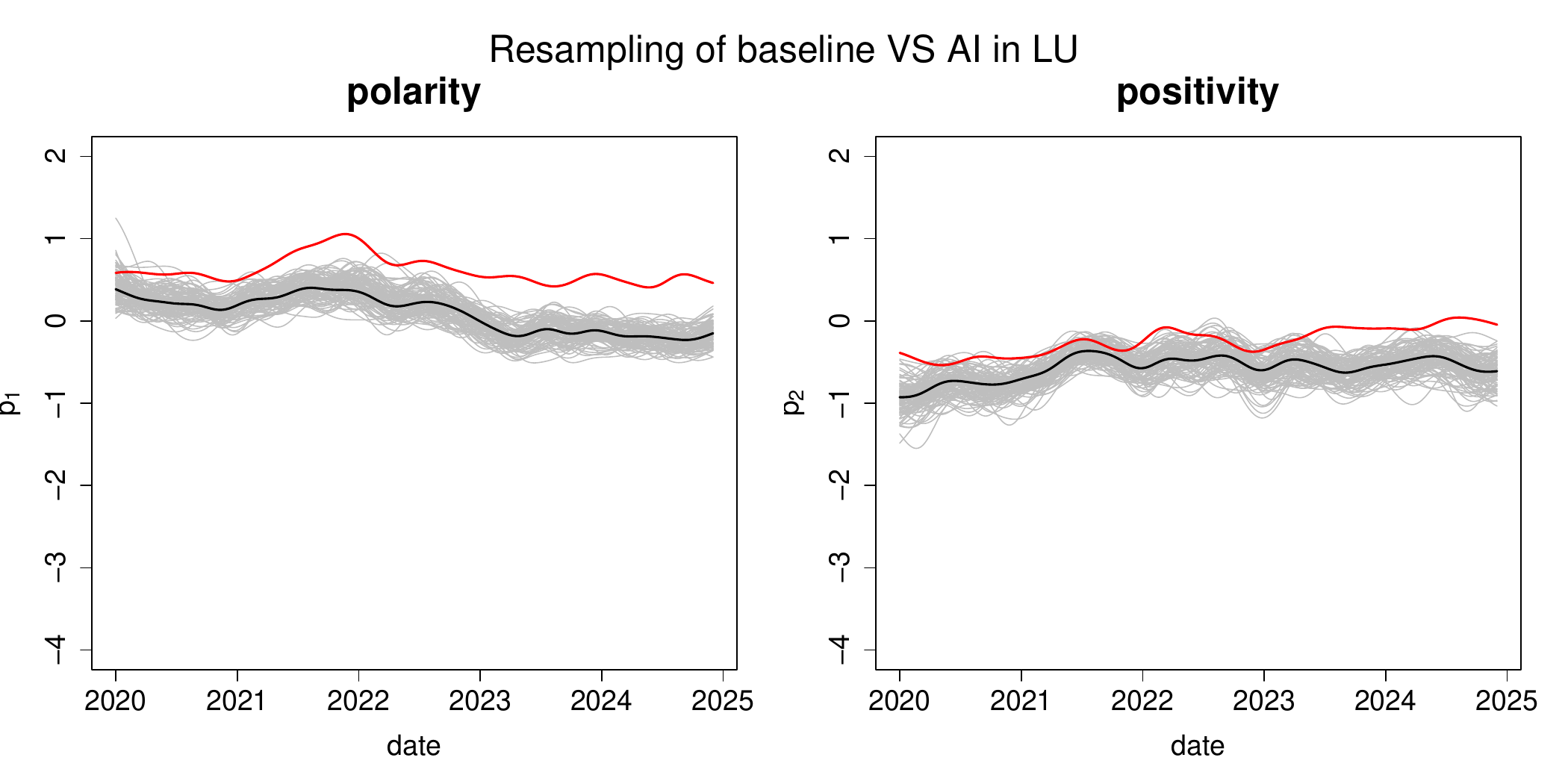}
\end{minipage}
\\[1mm]
\begin{minipage}[c]{0.04\textwidth}
\centering
\rotatebox{90}{%
\textbf{Lithuania} 
}
\end{minipage}
\hspace{3mm} 
\hfill
\begin{minipage}[c]{0.95\textwidth}
\includegraphics[width=\textwidth,trim=0 12mm 0 28mm,clip]{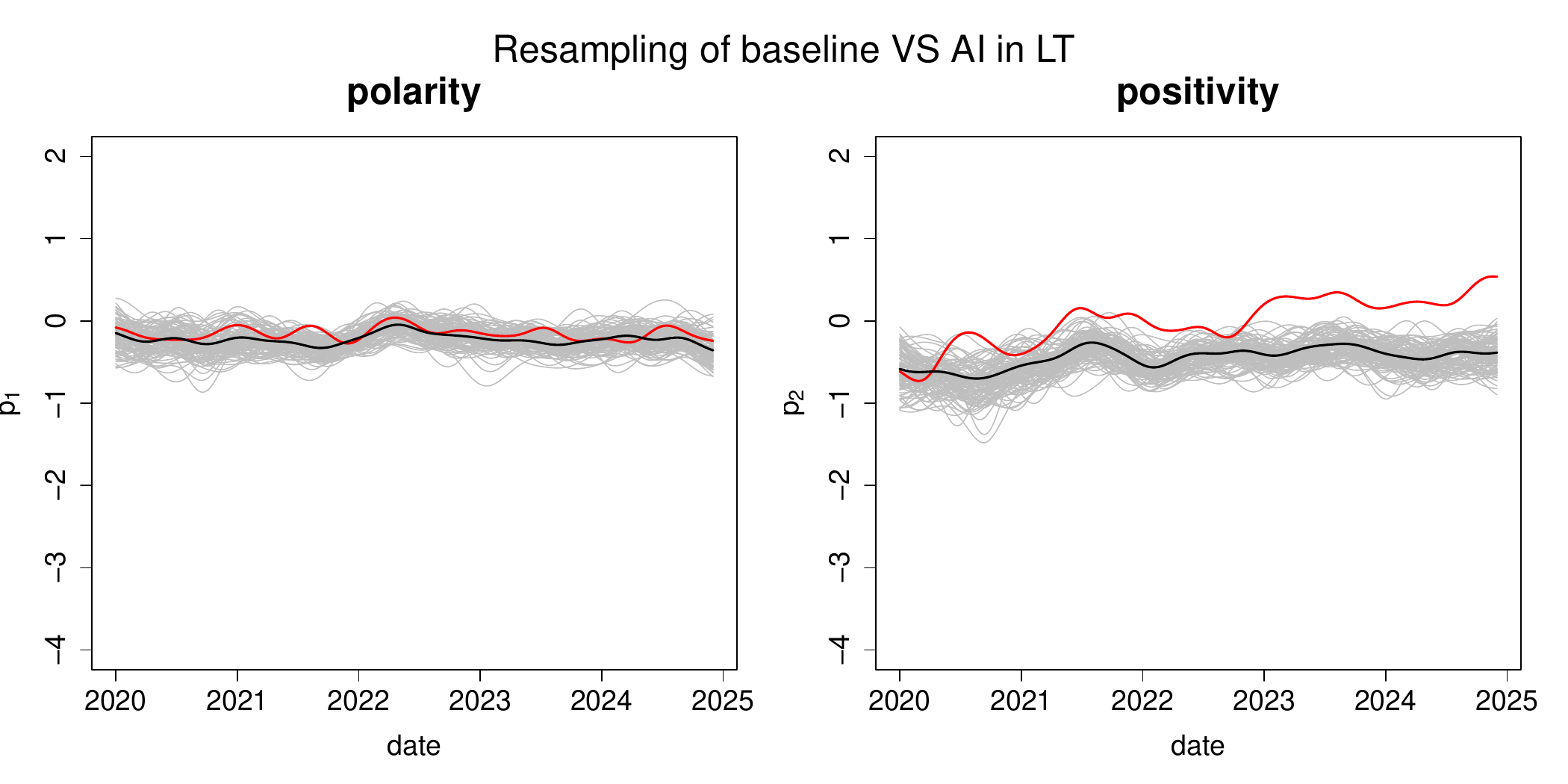}
%
\end{minipage}
\\[1mm]
\caption{AI data. Curves obtained from the re-sampling of the baseline data (gray) and curves obtained from the AI data (red) for Netherlands, Luxembourg and Lithuania. Polarity on the left panel and positivity on the right panel.
\label{fig:AI_NL_LU_LT}}
\end{figure}

The top panel of Fig~\ref{fig:AI_NL_LU_LT}
shows the Netherlands, the case with the lowest score on positivity, where the positivity curves still mostly lies above the distribution of the resampled baseline data. 
The middle and bottom panels show the countries with scores of polarity at the two extremes of the distribution: Lithuania, which has a low value (signaling a polarity curve for AI very similar to the baseline), and Luxembourg, which has a high value (signaling a strong difference between the AI curve and the baseline: in this case, the news articles speaking of AI are more polarized than the news articles of the overall media landscape).

\section{Additional results on the statistical analysis of temporal trends in AI data}
\label{sec:appendix_intra_2_AI}

\begin{figure}[ht]
\begin{minipage}[c]{0.04\textwidth}
\centering
\textbf{ } 
\end{minipage}
\hspace{3mm} \hfill
\begin{minipage}[c]{0.45\textwidth}
\centering\textbf{First derivative} 
\end{minipage}
\hfill
\begin{minipage}[c]{0.45\textwidth}
\centering\textbf{Original curves} 
\end{minipage}
\\[3mm]
\begin{minipage}[c]{0.04\textwidth}
\centering
\rotatebox{90}{%
\textbf{Spain} 
}
\end{minipage}
\hspace{3mm} \hfill
\begin{minipage}[c]{0.95\textwidth}
\includegraphics[width=\textwidth,trim=0 12mm 0 28mm,clip]{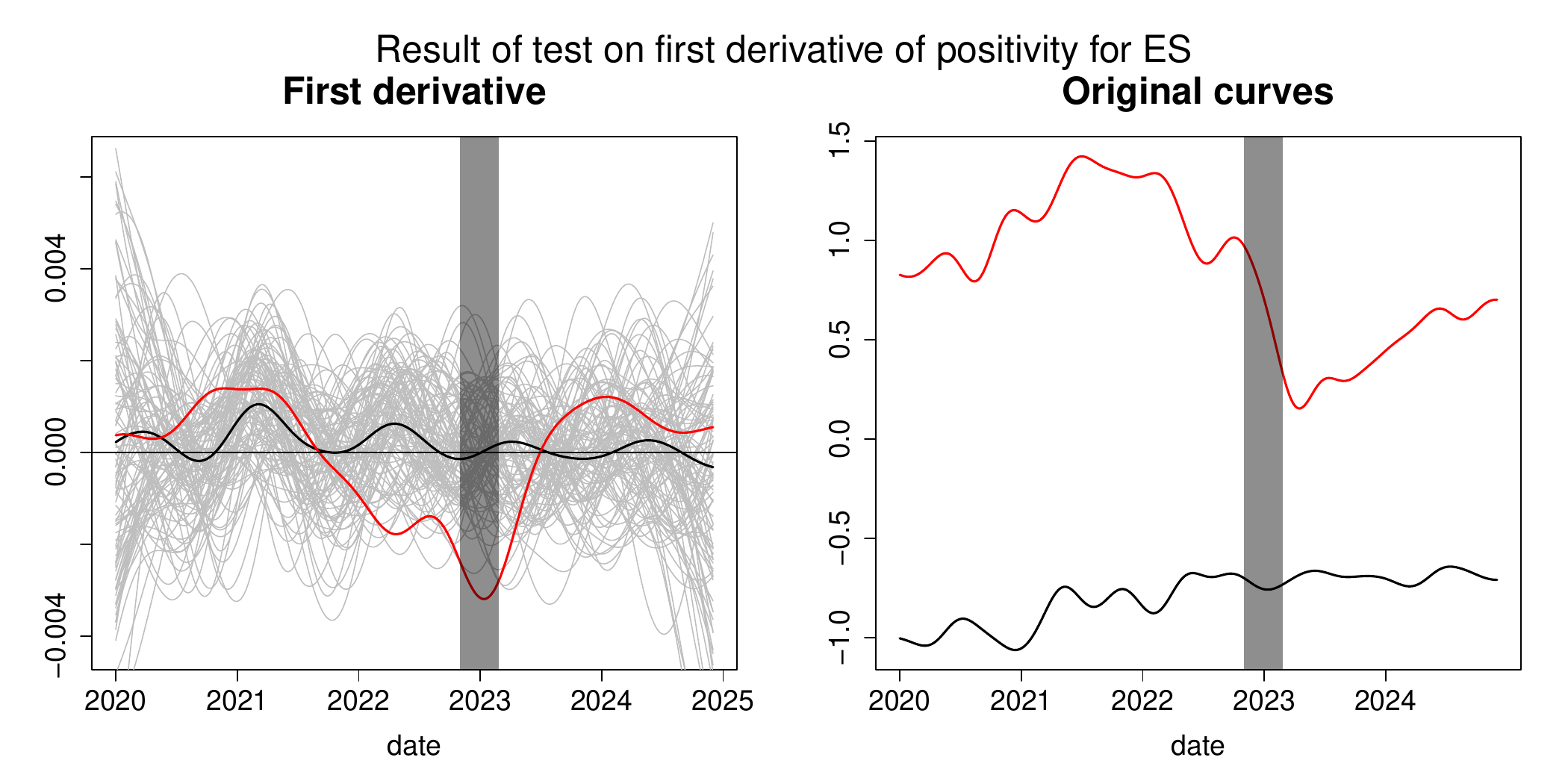}
%
%
\end{minipage}
\\[2mm]
\begin{minipage}[c]{0.04\textwidth}
\centering
\rotatebox{90}{%
\textbf{Greece} 
}
\end{minipage}
\hspace{3mm} \hfill
\begin{minipage}[c]{0.95\textwidth}
\includegraphics[width=\textwidth,trim=0 12mm 0 28mm,clip]{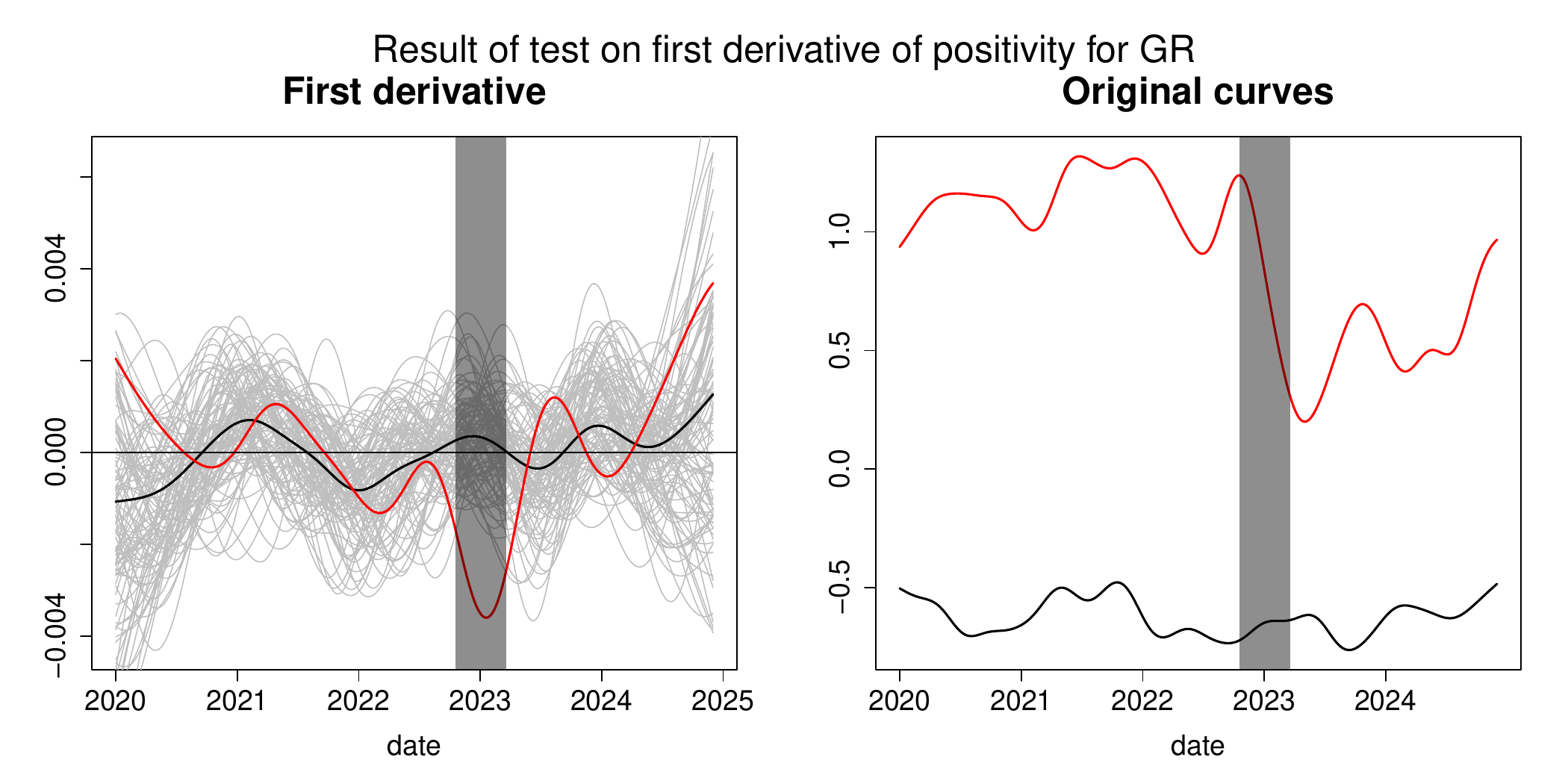}
%
%
\end{minipage}
\\[2mm]
\begin{minipage}[c]{0.04\textwidth}
\centering
\rotatebox{90}{%
\textbf{Ireland} 
}
\end{minipage}
\hspace{3mm} \hfill
\begin{minipage}[c]{0.95\textwidth}
\includegraphics[width=\textwidth,trim=0 12mm 0 28mm,clip]{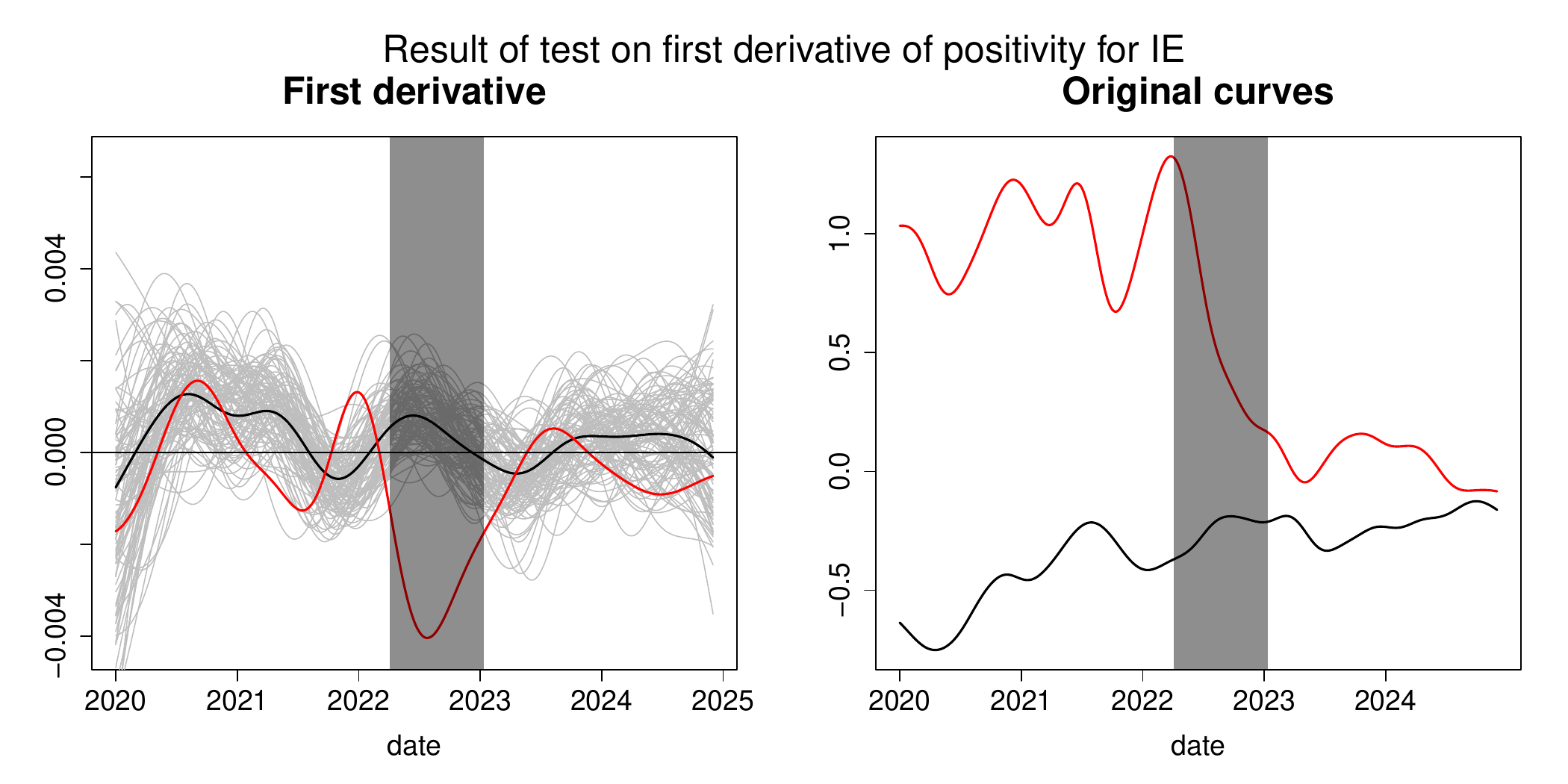}
%
\end{minipage}
\\[3mm]
\caption{AI data. Results of the bootstrap tests on the first derivatives of the positivity curves for Spain, Greece and Ireland. The red curves correspond to AI data, the black curves correspond to baseline data and the gray curves correspond to resampled baseline data.
\label{fig:ES_GR_IE_AI_derivative}}
\end{figure}

%
Fig~\ref{fig:ES_GR_IE_AI_derivative} shows the results of the bootstrap tests on the first derivatives for three example countries (Spain, Greece and Cyprus): a significant difference is found in all three cases at the end of 2022 and beginning of 2023.

%

\section{Quantitative multi-country sentiment analysis for trend significance assessment in AI unsmoothed data}
\label{app:finitedifferences}

This sections reports the results of the trend‑significance analysis of the unsmoothed monthly AI data, applying the bootstrap test to the finite differences rather than to the pointwise evaluations of the smooth first derivative (as done in Section \ref{sec:intra_2_AI}).
The finite differences are computed by subtracting from the positivity value of each month the value of the previous month. The bootstrap test is applied with the same parameters used in Section \ref{sec:intra_2_AI}.

\begin{figure}[tb!]
\begin{center}
\includegraphics[trim= 0 12mm 0 0,clip,width=0.55\textwidth]{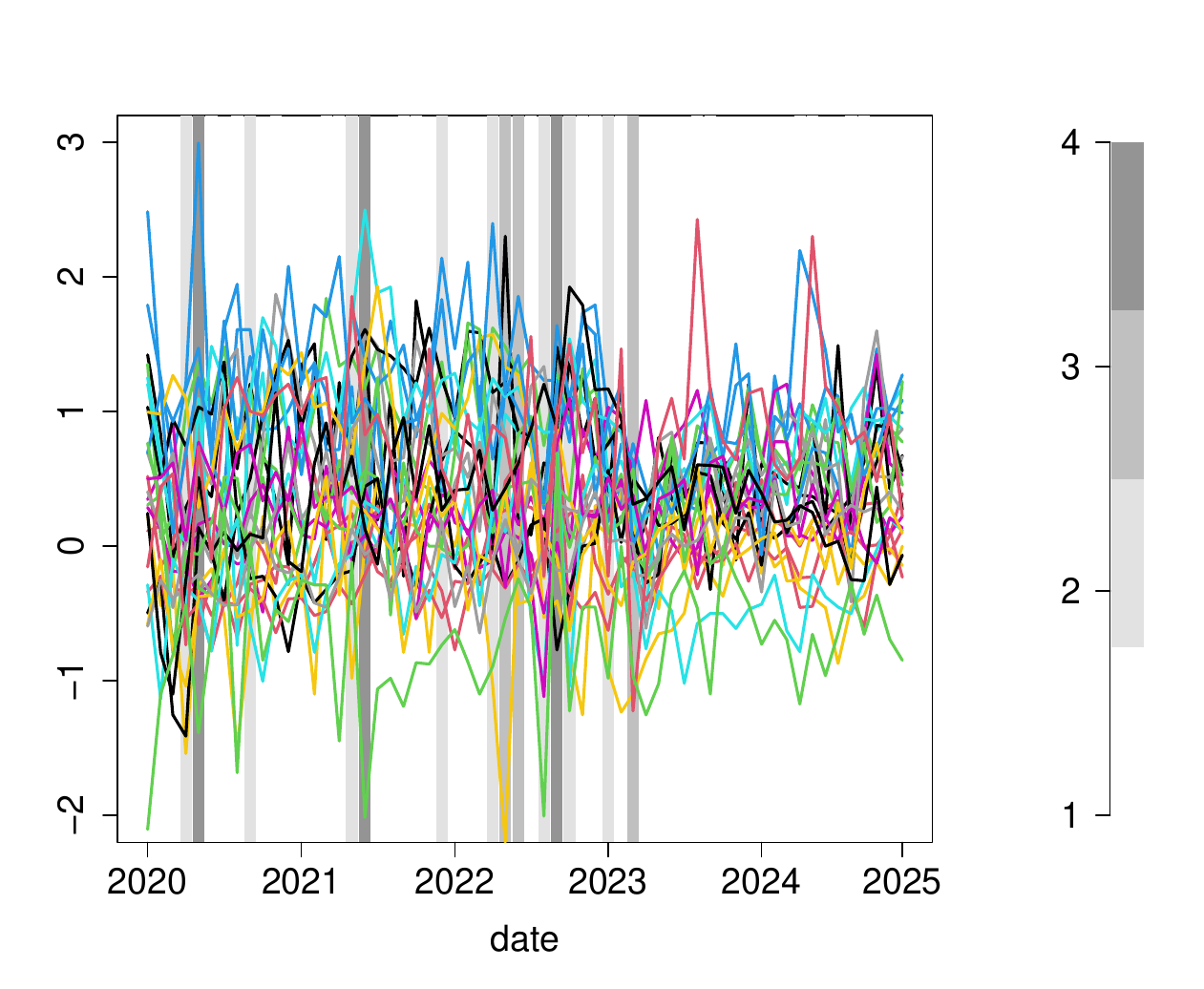}
\end{center}
\caption{AI unsmoothed data. Summary of the results of the bootstrap tests on the finite differences of the positivity curves for all the countries. Gray background colors are proportional to the count of countries for which a significant difference has been found.
\label{fig:ALL_AI_finitedifferences}}
\end{figure}

Fig~\ref{fig:ALL_AI_finitedifferences} summarizes the results for all the countries. With respect to the results obtained on the smoothed curves, (see Fig~\ref{fig:ALL_AI_derivative}), we obtain more isolated signals. Many signals detected are concentrated in the same period identified by the analysis on smooth data (i.e. between the end of 2022 and the beginning of 2023), but other periods (such as mid-2020 and mid-2021) are identified as well.

\begin{figure}[ht]
\begin{minipage}[c]{0.04\textwidth}
\centering
\textbf{ } 
\end{minipage}
\hspace{3mm} \hfill
\begin{minipage}[c]{0.45\textwidth}
\centering\textbf{Finite differences} 
\end{minipage}
\hfill
\begin{minipage}[c]{0.45\textwidth}
\centering\textbf{Original curves (unsmoothed)} 
\end{minipage}
\\[3mm]
\begin{minipage}[c]{0.04\textwidth}
\centering
\rotatebox{90}{%
\textbf{Spain} 
}
\end{minipage}
\hspace{3mm} \hfill
\begin{minipage}[c]{0.95\textwidth}
\includegraphics[width=\textwidth,trim=0 12mm 0 28mm,clip]{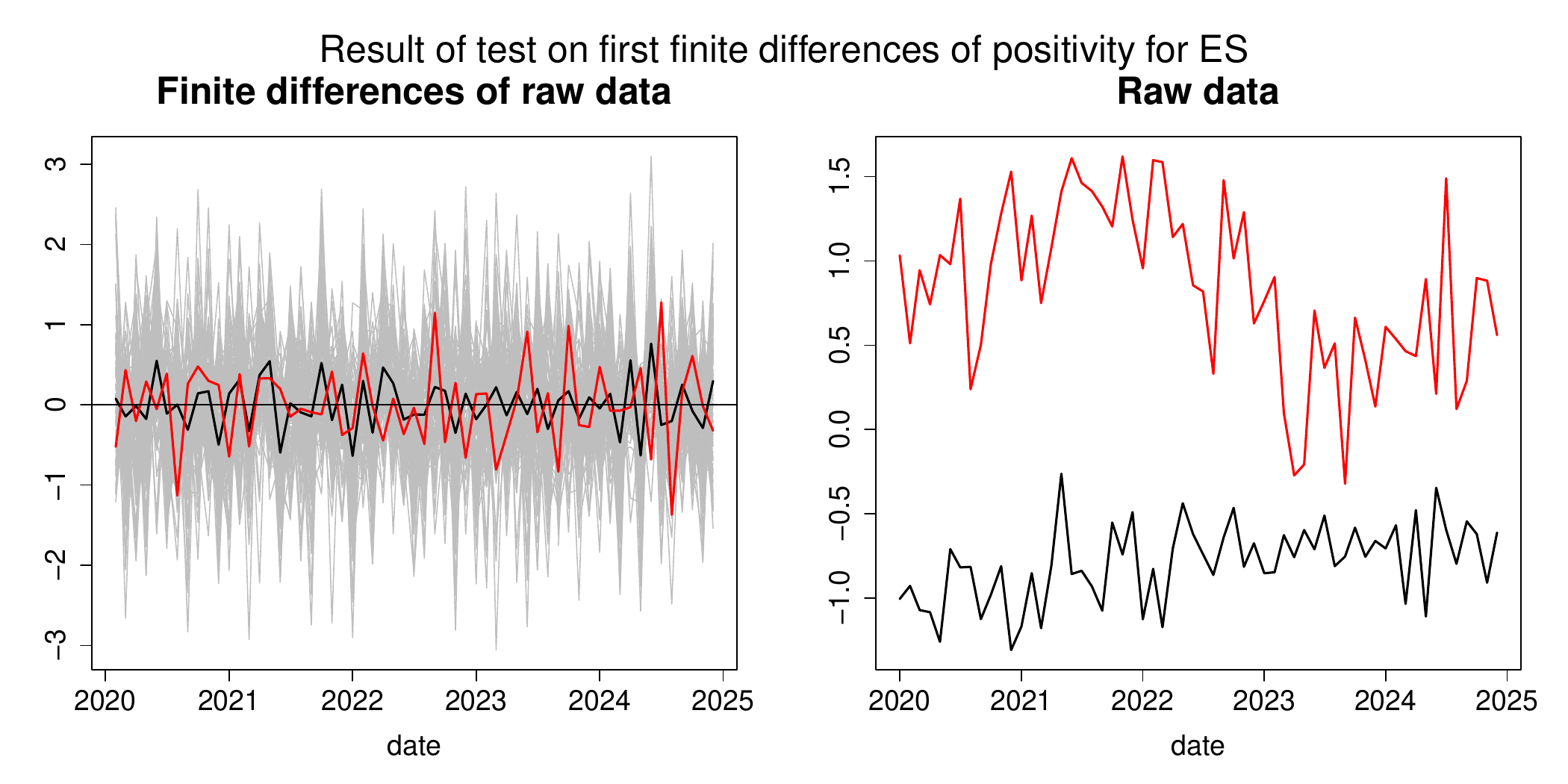}
\end{minipage}
\\[2mm]
\begin{minipage}[c]{0.04\textwidth}
\centering
\rotatebox{90}{%
\textbf{Austria} 
}
\end{minipage}
\hspace{3mm} \hfill
\begin{minipage}[c]{0.95\textwidth}
\includegraphics[width=\textwidth,trim=0 12mm 0 28mm,clip]{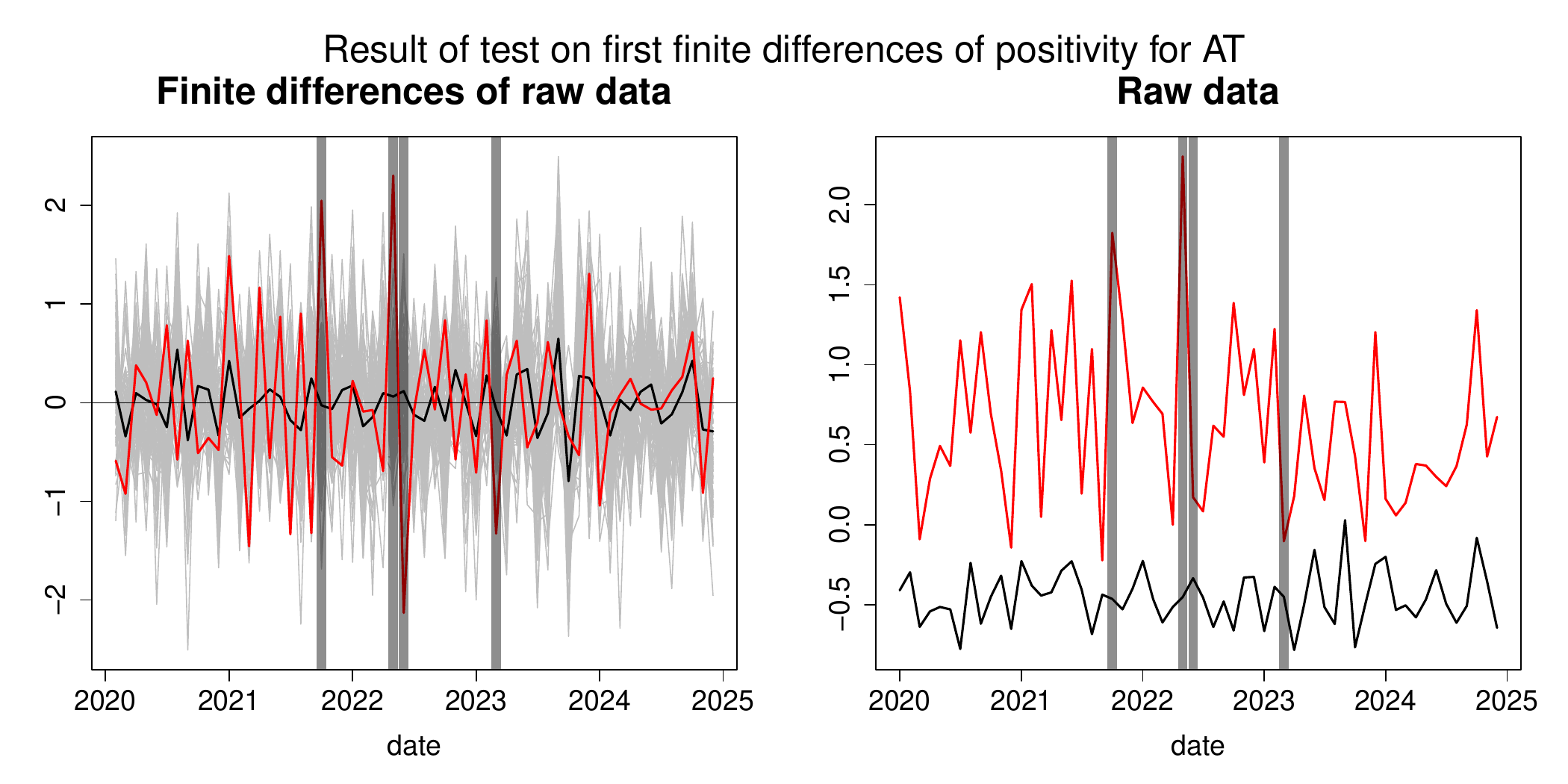}
\end{minipage}
\\[3mm]
\caption{Unsmoothed AI data. Results of the bootstrap tests on the finite differences of the positivity curves for Spain and Austria. The red curves correspond to AI data, the black curves correspond to baseline data and the gray curves correspond to resampled baseline data.
\label{fig:AT_ES_AI_finitedifferences}}
\end{figure}

Fig~\ref{fig:AT_ES_AI_finitedifferences} shows the results of the bootstrap tests applied to the finite differences for two example countries (Spain and Austria). In the fist case (Spain, top panels) no significant difference is found, while the analysis performed on the first derivative of the smoothed data identified a significant difference at the end of 2022 and beginning of 2023 (see top panels of Fig.~\ref{fig:ES_GR_IE_AI_derivative}). In the second case (Austria, bottom panels), significant differences are found in four isolated months, while the corresponding analysis on the smoothed version of the data did not detect any significant difference.  

\end{document}